\documentclass[letterpaper]{article} 
\usepackage{aaai25}  
\usepackage{times}  
\usepackage{helvet}  
\usepackage{courier}  
\usepackage[hyphens]{url}  
\usepackage{graphicx} 
\usepackage{natbib}  
\usepackage{caption} 
\usepackage{algorithm}
\usepackage{algorithmic}

\usepackage{booktabs}
\usepackage{multirow}
\usepackage{amsfonts}
\usepackage{amsmath}
\usepackage{mathtools}
\usepackage{subcaption}
\usepackage[table]{xcolor}
\usepackage{bm}

\usepackage{newfloat}
\usepackage{listings}
\DeclareCaptionStyle{ruled}{labelfont=normalfont,labelsep=colon,strut=off} 
\floatstyle{ruled}
\newfloat{listing}{tb}{lst}{}
\floatname{listing}{Listing}
\def\equaladvising{%
  \ifnum\value{eqfn}=0%
    \footnote{Equal advising.}%
    \setcounter{eqfn}{\value{footnote}}%
  \else%
    \footnotemark[\value{eqfn}]%
  \fi%
}%

\title{Offline Safe Reinforcement Learning Using Trajectory Classification}
\author{
    Ze Gong, Akshat Kumar\equaladvising, Pradeep Varakantham\equaladvising
}
\affiliations{

    School of Computing and Information Systems\\
    Singapore Management University\\
    \{zegong, akshatkumar, pradeepv\}@smu.edu.sg
}

\begin{document}

\maketitle

\begin{abstract}
Offline safe reinforcement learning (RL) has emerged as a promising approach for learning safe behaviors without engaging in risky online interactions with the environment. Most existing methods in offline safe RL rely on cost constraints at each time step (derived from global cost constraints) and this can result in either overly conservative policies or violation of safety constraints. In this paper, we propose to learn a policy that generates {\em desirable} trajectories and avoids {\em undesirable} trajectories.  To be specific, we first partition the pre-collected dataset of state-action trajectories into \textit{desirable} and \textit{undesirable} subsets. Intuitively, the desirable set contains high reward and safe trajectories, and undesirable set contains {unsafe trajectories and low-reward safe trajectories}. \textit{Second}, we learn a policy that generates desirable trajectories and avoids undesirable trajectories, where (un)desirability scores are provided by a classifier learnt from the dataset of {\em desirable} and {\em undesirable} trajectories. 
This approach bypasses the computational complexity and stability issues of a min-max objective that is employed in existing methods. {Theoretically, we also show our approach's strong connections to existing learning paradigms involving human feedback}. \textit{Finally}, we extensively evaluate our method using the DSRL benchmark for offline safe RL. Empirically, our method outperforms competitive baselines, achieving higher rewards and better constraint satisfaction across a wide variety of benchmark tasks.
\end{abstract}

%

\section{Introduction}

Reinforcement learning (RL) has achieved remarkable success in autonomous decision-making tasks \cite{mnih2015human,ibarz2021train,kiran2021deep}. However, ensuring safety during policy training and deployment remains a significant challenge, especially for safety-critical tasks where unsafe behavior can lead to unexpected consequences. To address the problem, safe RL has been extensively studied to develop policies that satisfy safety constraints, typically formulated within the framework of constrained Markov Decision Processes \cite{altman2021constrained,gu2022review}. Safe RL agents generally interact with the environment to gather information on rewards and costs, and learn policies through constrained optimization. Nonetheless, online safe RL may inevitably violate safety constraints during training and deployment due to the need for environmental interaction. To mitigate this issue, offline safe RL \cite{liu2023datasets} has emerged as a promising learning paradigm, enabling policy learning using pre-collected offline datasets, thus avoiding direct interactions with the environment.

Existing offline safe RL research has explored several directions~\cite{liu2023datasets}, such as enforcing constrained optimization into offline RL~\cite{xu2022constraints}, using distribution correction-based methods~\cite{leecoptidice}, and employing sequence modeling frameworks~\cite{liu2023constrained}. Though promising, several limitations remain. First, most existing methods rely on local cost constraints at each time step, which can lead to loss of sequential information related to global cost constraints at the trajectory level~\cite{xu2022constraints,leecoptidice}. This often results in either overly conservative policies with low rewards or the failure to generate a safe policy. Additionally, most existing works employ min-max optimization within the conventional RL framework~\cite{xu2022constraints}, resulting in an intertwined training process involving reward maximization and safety compliance resulting in stability issues. Lastly, some approaches require auxiliary models to be learned alongside the policy~\cite{leecoptidice}, or depend on complex model architectures \cite{liu2023constrained,zheng2024safe}, which contribute to high computational complexity.

To address such challenges, we propose a novel approach that solves the offline safe RL problem by utilizing a binary trajectory classifier. This classifier allows us to directly optimize a policy by leveraging both \textit{desirable} and \textit{undesirable} samples at the trajectory level, without relying on conventional min-max optimization or traditional RL techniques. We apply a two-phase algorithm. In the first phase, we divide the offline dataset into two subsets, one containing desirable trajectories and the other containing undesirable ones. {The desirable set includes \textit{safe trajectories} that satisfy the constraints, with each trajectory weighted according to its cumulative reward to reflect its level of desirability. The undesirable set comprises (a) all unsafe trajectories, and (b) selection of safe trajectories with low rewards}. {We also develop a concrete empirical methodology to construct such subsets}. In the second phase, we optimize the policy within a classifier designed to distinguish between desirable and undesirable trajectories. The classifier is specifically designed to provide a high score ({derived using the principle of maximum entropy RL}) to desirable trajectories, and low score to undesirable ones. By solving this trajectory classification problem, our algorithm learns the policy directly, bypassing the need for min-max optimization and traditional RL methods.
Additionally, we identify close connections with the Reinforcement Learning from Human Feedback (RLHF) paradigm~\cite{hejnacontrastive}, providing theoretical support for our method and a promising direction for future research.

For evaluation, we adopt the well-established DSRL benchmark \cite{liu2023datasets}, designed specifically for offline safe RL approaches. We conduct extensive experiments by comparing our method with several state-of-the-art baselines in terms of normalized reward and normalized cost across 38 continuous control tasks in three popular domains. The results suggest that our method successfully learns policies that achieve high rewards while satisfying safety constraints in most tasks, outperforming state-of-the-art baselines. 

The main contributions of the work are three-fold:
\begin{itemize}
    \item We introduce a novel approach that solves offline safe RL by utilizing a trajectory classifier, that enables direct policy learning within a standard classification framework and eliminating the need for challenging min-max optimization.
    \item Furthermore, to ensure stability and handling trajectory level safety, we provide a contrastive trajectory classifier that employs a novel score function built from regret based preference models. 
    \item Extensive experiments demonstrate that our method surpasses SOTA baselines in both reward maximization and constraint satisfaction within a well-established benchmark dataset.
\end{itemize}

\section{Related Work}

\subsection{Safe RL}

Ensuring safety remains a critical challenge in RL during both policy training and deployment phases \cite{garcia2015comprehensive,achiam2017constrained,gu2022review,ji2023omnisafe,hoang2024imitate}. Safe RL aims to learn a policy that maximizes the cumulative reward while satisfying the safety constraints through interactions with the environment. Various techniques have been explored to address safe RL, such as Lagrangian-based optimization \cite{chow2018risk,tessler2018reward,stooke2020responsive,chen2021primal,ding2020natural}, and correction-based approaches \cite{zhao2021model,luo2021learning}. However, ensuring zero constraint violations during training and deployment remains a significant challenge.



\subsection{Offline Safe RL}

Offline safe RL has emerged as a new paradigm for learning safe policies using pre-collected offline datasets \cite{le2019batch,guan2024voce}. Existing methods include behavior cloning (BC) \cite{liu2023constrained}, distribution correction-based approach \cite{leecoptidice}, and incorporating constraints into existing offline RL techniques \cite{xu2022constraints}. Recent research has also explored decision transformer-based approaches \cite{liu2023constrained} and diffusion-based methods \cite{zhengsafe}. Despite these advancements, several limitations persist, such as performance degradation, unstable training due to min-max optimization, and computational {challenges} associated with complex model architectures.

\section{Preliminaries}

Safe RL problems are often formulated using Constrained Markov Decision Processes (CMDPs) \cite{altman2021constrained}, defined by a tuple $\mathcal{M}=(\mathcal{S}, \mathcal{A}, T, r, c, \mu_0, \gamma)$ where $\mathcal{S}$ is the state space and $\mathcal{A}$ is the actions space; $T: \mathcal{S}\times\mathcal{A} \rightarrow \Delta(\mathcal{S})$ describes the transition dynamics; $r: \mathcal{S}\times\mathcal{A}\rightarrow\mathbb{R}$ is the reward function; $c: \mathcal{S}\times\mathcal{A}\rightarrow [0, C_\text{max}]$ is cost function where $C_\text{max}$ is the maximum cost; $\mu_0: \mathcal{S} \rightarrow [0,1]$ represents the initial state distribution; $\gamma$ is the discount factor. {Let $\pi(a|s)$ denote the probability of taking action $a$ in state $s$.} The state-value function for a policy $\pi$ is defined as $V_r^\pi (s)=\mathbb{E}_\pi [ \sum_{t=0}^\infty \gamma^t r(s_t, a_t) | s_0=s]$ with discounted cumulative rewards. Similarly, the cost state-value function is defined as $V_c^\pi (s)=\mathbb{E}_\pi [ \sum_{t=0}^\infty \gamma^t c(s_t, a_t) | s_0=s]$ based on discounted cumulative costs. 

In offline safe RL, we assume access to a pre-collected offline dataset $D=\{\tau_1, \tau_2, \ldots \}$ of trajectories. {Each trajectory $\tau$ contains a sequence of tuples $(s_t,a_t,s_{t+1},r_t,c_t )$ from start till episode end, where $r_t$ and $c_t$ represent the reward and cost at time step $t$}.
The goal in offline safe RL is to maximize the expected cumulative rewards $V_r^\pi (s)$ while satisfying safety constraints determined by the cumulative costs $V_c^\pi$ and a specified cost threshold. The formal constrained objective function is written as follows: 
\begin{equation}
    \max_\pi \mathbb{E}_s \left[ V_r^\pi (s) \right] \text{, s.t., } \mathbb{E}_s \left[ V_c^\pi (s) \right] \leq l \text{; } \mathcal{D}( \pi \| \pi_\beta ) \leq \epsilon
    \label{eq:srl_obj}
\end{equation}
where $l\geq 0$ is the cost threshold and $\pi_\beta$ is the underlying behavioral policy used to collect the dataset. $\mathcal{D}( \pi \| \pi_\beta )$ denotes the KL divergence between the policy $\pi$ and the behavioral policy $\pi_\beta$ which is used to address distributional drift in the offline setting.

\section{Our Approach}

\begin{figure}[t]
    \centering
    \begin{subfigure}[t]{0.35\textwidth}
        \centering
        \includegraphics[width=\textwidth]{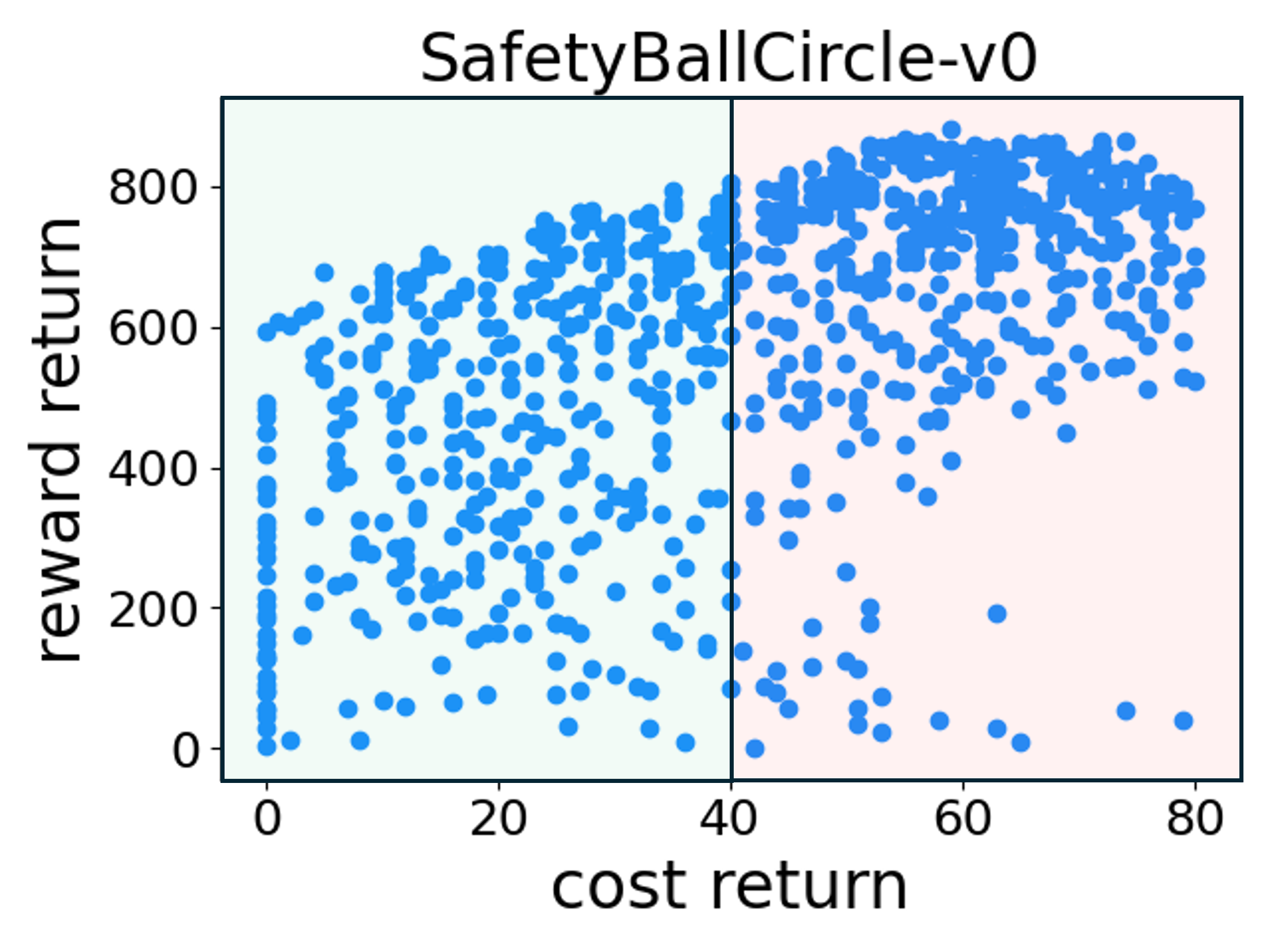}
    \end{subfigure}%
    \caption{Visualization of trajectories in the reward-cost return space with a cost threshold of $40$. The x-axis is the total cost and the y-axis is the total reward. Each blue dot in the figure corresponds to a trajectory in the offline dataset. Based on the cost threshold, trajectories are categorized into safe (in the green area) and unsafe (in the pink area).}
    \label{fig:cost}
\end{figure}

The constraint term in Equation~\ref{eq:srl_obj} serves as a criterion for distinguishing between safe and unsafe trajectories. As shown in Figure~\ref{fig:cost}, given a cost threshold, the trajectories in the offline datasets can be divided into two categories: safe and unsafe. According to the objective function in Equation~\ref{eq:srl_obj}, all unsafe trajectories are considered undesirable, while all safe trajectories are viewed as \textit{potential} candidates for desirable samples. Additionally, the maximization term in Equation~\ref{eq:srl_obj} reflects the level of desirability of trajectories, with those yielding higher cumulative rewards deemed more desirable by the objective function. Building on this insight, we aim to learn a policy capable of generating desirable trajectories with high probability. We propose a two-phase approach -- (a) we first construct two contrastive datasets (desirable, undesirable), and (b) then we transform the offline safe RL problem into a direct policy learning within a trajectory classifier.

\subsection{Contrastive Dataset Construction}
\label{sec:data_construction}

Given the pre-collected offline dataset $D$, we create two new subdatasets at the trajectory level: one containing desirable trajectories and the other containing undesirable ones. \textit{Desirable} refers to trajectories with high cumulative rewards and being safe, while \textit{undesirable} refers to those with low cumulative rewards or unsafe.

Using the predefined cost threshold {$l$}, we first split the dataset into two categories based on the cumulative cost, i.e., safe and unsafe.
Within the safe trajectories, we further rank them according to cumulative rewards. The top $x\%$ of these safe trajectories are selected as desirable. {Moreover, we identify the bottom $y\%$ of the safe trajectories, along with all unsafe trajectories as undesirable ($x, y$ are hyperparameters that we show how to set empirically)}. To reflect the level of desirability, we assign normalized weights $w$ ($w \in [\delta, 1]$, and $\delta=0.5$) to the trajectories in the desirable datasets, based on their cumulative rewards:
\begin{equation}
    w_\tau=\frac{v(\tau)-v_\text{min}}{v_\text{max}-v_\text{min}} \cdot (1-\delta) + \delta
    \label{eq:w}
\end{equation}
where $v(\tau)$ denotes the return of the trajectory $\tau=(s_1, a_1, s_2, a_2, \cdots, s_k, a_k)$ with length $k$. The terms $v_\text{max}$ and $v_\text{min}$ represent the empirical maximum and minimum values of trajectory returns over all samples in the original dataset. Equation \ref{eq:w} is used to normalize the original returns to the range of $[\delta, 1]$. Furthermore, we assign weights to the unsafe trajectories to be $1$, since we aim to avoid unsafe outcomes irrespective of their return values, treating all unsafe trajectories equally. Moreover, undesirable safe trajectories within the undesirable set are weighted reversely according to their returns, meaning that safe trajectories with lower returns receive higher weights during training:
\begin{equation}
    w_\tau=\left(1 - \frac{v(\tau)-v_\text{min}}{v_\text{max}-v_\text{min}}\right) \cdot (1-\delta) + \delta
    \label{eq:w-reverse}
\end{equation}

\subsection{Contrastive Trajectory Classification (\texttt{TraC})}

With the desirable and undesirable datasets, our goal is to learn a policy that generates desirable behavior with high probability and undesirable behavior with very low probability. Intuitively, we can formulate the following objective function \footnote{In Equation \ref{eq:bc}, our objective emphasizes training a policy to favor desirable trajectories and avoid undesirable ones, initially ignoring the KL term. Later, in Equation \ref{eq:score}, the score function integrates the reference (behavioral) policy through the KL-regularized RL framework, addressing the distribution drift in offline RL.}:
\begin{equation}
    \max_{\pi_\theta} \lambda_d \mathbb{E}_{\tau \sim D^d} [w_\tau f(\tau; \pi_\theta)] - \lambda_u \mathbb{E}_{\tau \sim D^u} [w_\tau f(\tau; \pi_\theta)]
    \label{eq:bc}
\end{equation}
where $D^d$ represents the dataset containing all desirable trajectories, and $D^u$ represents the dataset containing all undesirable trajectories. The function $f(\tau; \pi_\theta)$ intuitively should return high value for desirable $\tau$ under $\pi_\theta$, and low value for undesirable trajectories. The  $\lambda_d$ and $\lambda_u$ are the factors that represent the relative importance of desirable and undesirable data during training, respectively. The influence of these factors may vary depending on the dataset ratios and specific tasks involved in addressing dataset imbalance.

Specifically, the function $f(\tau; \pi)$ could be modeled using Behavior Cloning (BC) as follows:
\begin{equation}
    f(\tau; \pi) = \sum_t\log\pi(a_t | s_t) \nonumber
\end{equation}
However, there are several limitations of BC approach. First, it loses sequential information by optimizing the policy for each $(s_t, a_t)$ pair individually. Additionally, using the BC formulation, minimizing the logarithmic terms in the second negative term of Equation~\ref{eq:bc} can be unstable because the logarithm has no lower bound as $\pi(a_t | s_t) \in [0,1]$.

\paragraph{Our method} To address above issues, we make two key contributions:
\begin{itemize}
    \item Instead of direct log likelihood maximization, we consider a Contrastive Trajectory Classification (\texttt{TraC}) problem that avoids the stability issues, and also design $f(\cdot)$ that depends on entire trajectory and not just on individual $(s_t, a_t)$ pairs.
    \item More importantly, to enable imitation of desirable trajectories and avoidance of undesirable trajectories, we propose $f(\cdot)$ to be sigmoid over a score function that assigns higher value to a {\em desirable} trajectory and lower values to {\em undesirable} trajectories. 
\end{itemize}

\noindent First, the \texttt{TraC} problem formulation is defined as:
\begin{equation}
    \max_{\pi_\theta} \lambda_d \mathbb{E}_{\tau \sim D^d} [w_\tau f(y_\tau, \tau; \pi_\theta)] + \lambda_u \mathbb{E}_{\tau \sim D^u} [w_\tau f(y_\tau, \tau; \pi_\theta)] \nonumber
\end{equation}
where $y\in\{0,+1\}$ is the binary desirability label, where $+1$ is for desired trajectories, and $0$ is for undesired trajectories. The parameters $\lambda_d$ and $\lambda_u$ are determined under the constraints $\lambda_d+\lambda_u=1$ and $\frac{\lambda_d N_d}{\lambda_u N_u} = \eta$, where $N_d$ and $N_u$ are the numbers of desirable and undesirable trajectories in our constructed datasets. The parameter $\eta$ is predefined to control the relative values of $\lambda_d$ and $\lambda_u$\footnote{The formulation of the constraints for defining $\lambda_d$ and $\lambda_u$ is motivated by the work of Kahneman-Tversky Optimization (KTO) \cite{ethayarajh2024kto}, and we adapt it to suit our setting.}. We define $f(y_\tau, \tau; \pi_\theta)$ as the probability of predicting the binary desirability label $y_\tau$ for the corresponding trajectory $\tau$ under policy $\pi_\theta$:
\begin{equation}
    f(y_\tau, \tau; \pi_\theta) = p(Y=y_\tau | \tau, \pi_\theta) \coloneqq \sigma(\psi(\tau, \pi_\theta))
    \label{eq:f}
\end{equation} 
where $\sigma$ represents the sigmoid function that maps the input to the range of $[0,1]$. The function $\psi(\tau, \pi_\theta)$ computes a score for the trajectory $\tau$ under policy $\pi_\theta$ which is used to calculate the probability.

\paragraph{Trajectory score function $\psi$}Next, we define the score function $\psi(\tau, \pi_\theta)$, and establish some of its properties. Since, we have to ensure $\psi$ captures a preference for desirable trajectories over undesirable trajectories, we build on key insights from {regret-based preference model~\cite{knox2024models}} that is expressed as:
{\small
\begin{align}
    P[\tau^+ \!\succ\! \tau^-] \!=\! 
    \frac{\exp \sum_{\tau^+} \gamma^t A_r^*(s_t^+, a_t^+)}{\exp \sum_{\tau^+} \!\! \gamma^t A_r^*(s_t^+, a_t^+) \!+\! \exp \sum_{\tau^-} \!\! \gamma^t A_r^*(s_t^-, a_t^-)} \nonumber
\end{align}}
where preference between two trajectories is denoted as $\tau^+ \succ \tau^-$; $\tau^+$ indicates the preferred trajectory, and $\tau^-$ represents the less preferred trajectory. The optimal advantage function $A_r^*(s, a) \coloneqq Q_r^*(s, a) - V_r^*(s)$ measures how much worse taking action $a$ in state $s$ is than acting according to optimal policy $\pi^*$, where $Q_r^*(s, a)$ denotes the optimal state-action value function and $V_r^*(s)$ denotes the optimal state value function under optimal policy $\pi^*$.

A reason for choosing this regret based preference model is that as discussed in CPL \cite{hejnacontrastive}, this preference learning framework shares similarities with contrastive learning approaches. The discounted sum of advantage function (i.e., $\sum_{\tau} \gamma^t A_r^*(s_t, a_t)$) can be considered as a trajectory’s score from the Noise Contrastive Estimation (NCE) \cite{gutmann2010noise} perspective. Furthermore, according to the principle of maximum entropy \cite{ziebart2008maximum,ziebart2010modeling,hejnacontrastive}, the optimal advantage function $A_r^*(s, a)$ can be equivalently represented:
\begin{equation}
    A_r^*(s, a) = \alpha \log \frac{\pi^*(a|s)}{\pi_\text{ref}(a|s)} \label{eq:adv}
\end{equation}
where {$\alpha(>0)$} is a temperature parameter. The reference policy $\pi_\text{ref}$ is introduced to regularize $\pi^*$, a common practice in KL-regularized RL \cite{galashov2019information}. Typically, standard behavior cloning on the offline dataset is used as $\pi_\text{ref}$. In our context, it can also be seen as a behavioral policy that mitigates the distribution drift issue in offline settings. We define the score for a trajectory in our setting as:
\begin{equation}
    \psi(\tau, \pi^*) \!=\! \sum_{t=0}^T \gamma^t A_r^*(s_t, a_t) \!=\! \sum_{t=0}^T \gamma^t \alpha \log \frac{\pi^*(a_t|s_t)}{\pi_\text{ref}(a_t|s_t)}  \label{eq:score}
\end{equation}

\paragraph{Score function justification} We next justify why having a high score for a trajectory $\tau$ also implies it is more likely under $\pi^*$. For simplicity, we ignore $\alpha$. From Equation~\ref{eq:adv}:
\begin{align}
    \log \pi^*(a|s) &= A_r^*(s, a) +  \log \pi_\text{ref}(a|s) \\
    \log p(\tau; \pi^*) &\propto \sum_{t=0}^T \log \pi^*(a_t|s_t) \label{eq:trjprob1}\\
    &= \sum_{t=0}^T [A_r^*(s_t, a_t) +  \log \pi_\text{ref}(a_t|s_t) ]\label{eq:trjprob2}
\end{align}
where $p(\tau; \pi^*)$ is the probability of trajectory under $\pi^*$. The ignored terms in Equation~\ref{eq:trjprob1} are the log probabilities of the transition function, which are independent of the policy. Thus, the log probability of observing $\tau$ under $\pi^*$ is directly proportional to its score function $\sum_{t=0}^T A_r^*(s_t, a_t)$ (in practice, to reduce variance, we use $\gamma^t$ to discount $A_r^*(s_t, a_t)$). This implies that assigning higher score (in Equation~\ref{eq:score}) to a trajectory makes it more likely (which should be the case for desirable trajectories).

\paragraph{Overall loss function} Using the score function in Equation~\ref{eq:score}, we formulate a new loss function that allows us to directly optimize the parameterized policy $\pi_\theta$ to align with the classification objective as follows:
\begin{align}
    L(\pi_\theta, D) &= - \mathbb{E}_{\tau \sim D} \Big[\lambda_d w_\tau \cdot y_\tau \log \left( \sigma \left( \psi(\tau, \pi_\theta) \right) \right)  \nonumber \\
    & + \lambda_u w_\tau \cdot (1 - y_\tau) \log \left( 1 - \sigma \left( \psi(\tau, \pi_\theta) \right) \right) \Big]
    \label{eq:loss}
\end{align}


In the first term of Equation \ref{eq:loss}, the objective is to learn a policy that assigns high values to desirable trajectories. This means the learning policy should be able to closely align with safe and high-reward trajectories, where desirability is determined by the discounted cumulative rewards of each safe trajectory. The second term focuses on ensuring the policy assigns low values to undesirable trajectories, enabling it to effectively avoid unsafe behaviors. Additionally, incorporating $\pi_\text{ref}$ in each term helps mitigate the distribution drift problem in offline setting and can be pretrained using Behavior Cloning (BC) across the entire dataset. Compared to Equation~\ref{eq:srl_obj}, our loss function implicitly captures the objective, the constraints, and behavior regularization, which are essential components in the conventional offline safe RL objective. {Our objective also does not require challenging min-max optimization, and only a single neural net to parameterize policy $\pi$ is used.}

\begin{figure*}
    \centering
    \begin{minipage}{0.02\textwidth}
        \centering
        \rotatebox[origin=center]{90}{BulletGym}
    \end{minipage}
    \begin{minipage}{0.97\textwidth}
        \centering
        \begin{subfigure}[t]{0.19\textwidth}
            \centering
            \includegraphics[width=\textwidth]{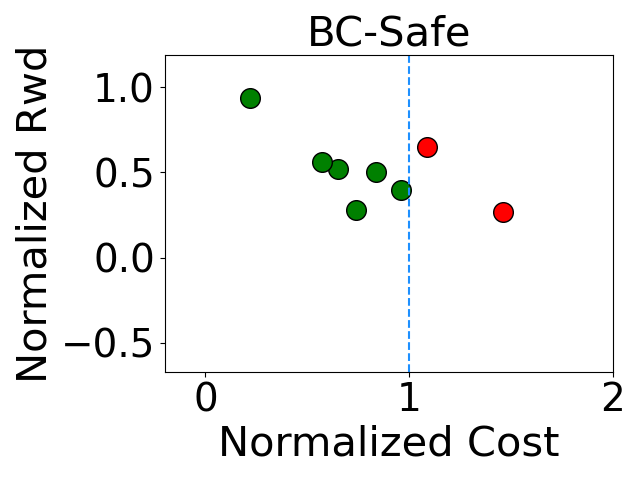}
        \end{subfigure}
        \begin{subfigure}[t]{0.19\textwidth}
            \centering
            \includegraphics[width=\textwidth]{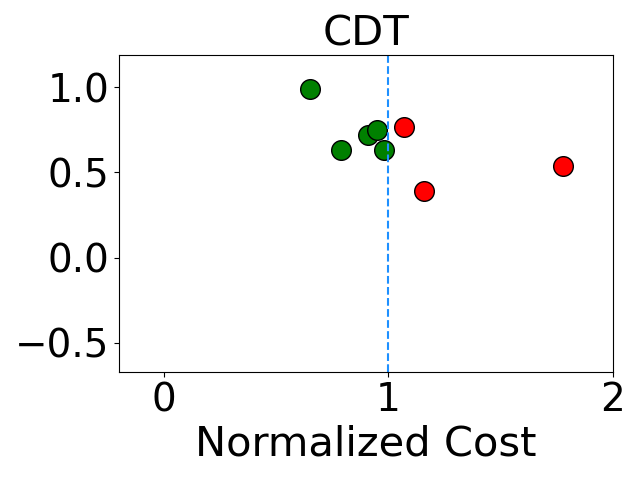}
        \end{subfigure}
        \begin{subfigure}[t]{0.19\textwidth}
            \centering
            \includegraphics[width=\textwidth]{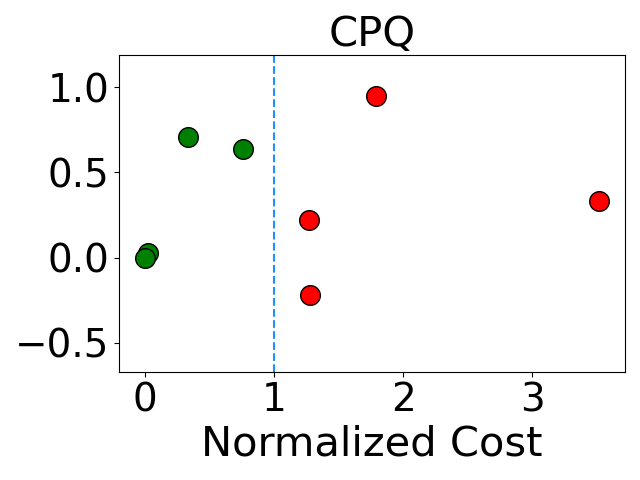}
        \end{subfigure}
        \begin{subfigure}[t]{0.19\textwidth}
            \centering
            \includegraphics[width=\textwidth]{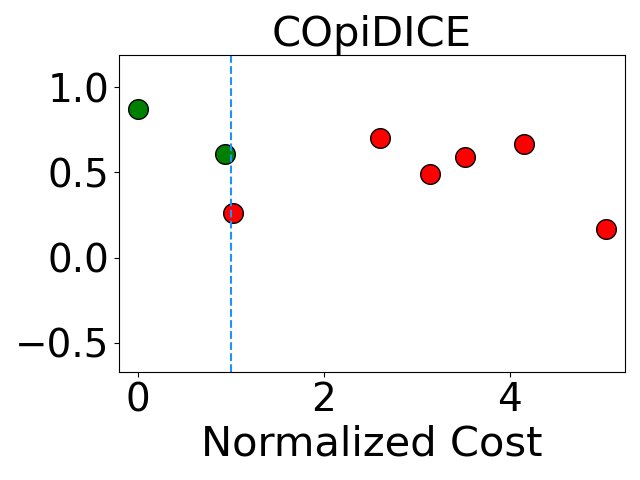}
        \end{subfigure}
        ~
        \begin{subfigure}[t]{0.19\textwidth}
            \centering
            \includegraphics[width=\textwidth]{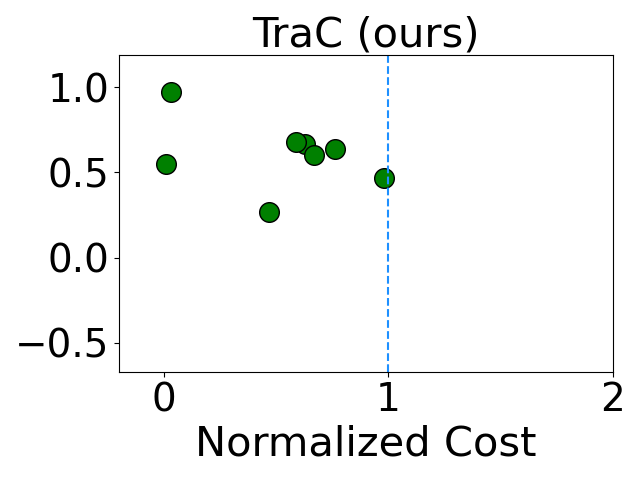}
        \end{subfigure}
    \end{minipage}
    \begin{minipage}{0.02\textwidth}
        \centering
        \rotatebox[origin=center]{90}{SafetGym}
    \end{minipage}
    \begin{minipage}{0.97\textwidth}
        \centering
        \begin{subfigure}[t]{0.19\textwidth}
            \centering
            \includegraphics[width=\textwidth]{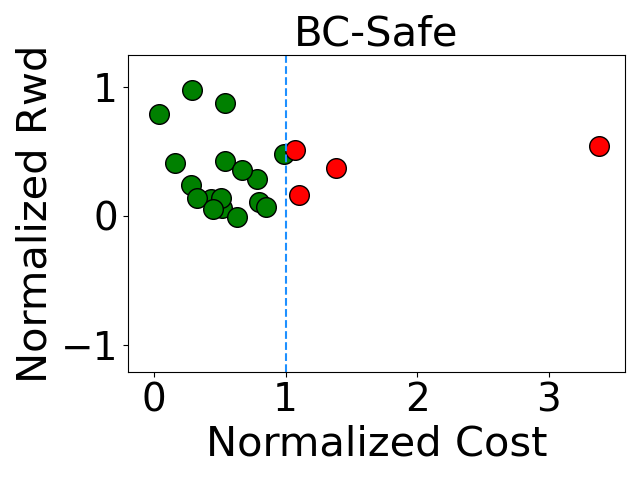}
        \end{subfigure}
        \begin{subfigure}[t]{0.19\textwidth}
            \centering
            \includegraphics[width=\textwidth]{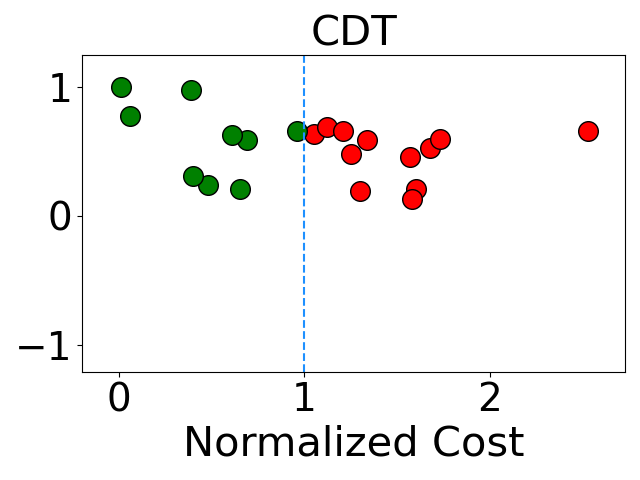}
        \end{subfigure}
        \begin{subfigure}[t]{0.19\textwidth}
            \centering
            \includegraphics[width=\textwidth]{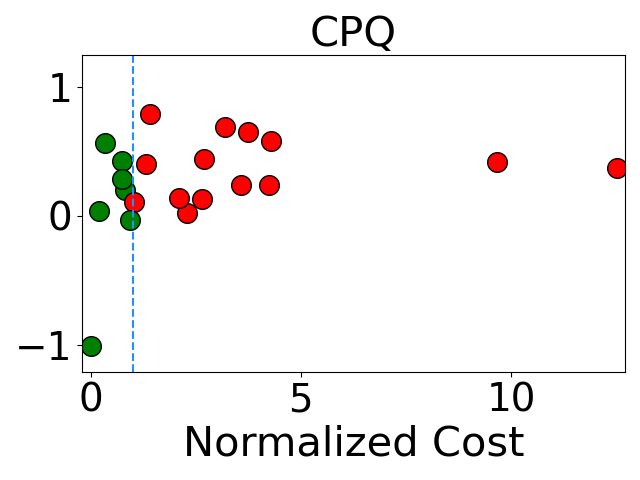}
        \end{subfigure}
        \begin{subfigure}[t]{0.19\textwidth}
            \centering
            \includegraphics[width=\textwidth]{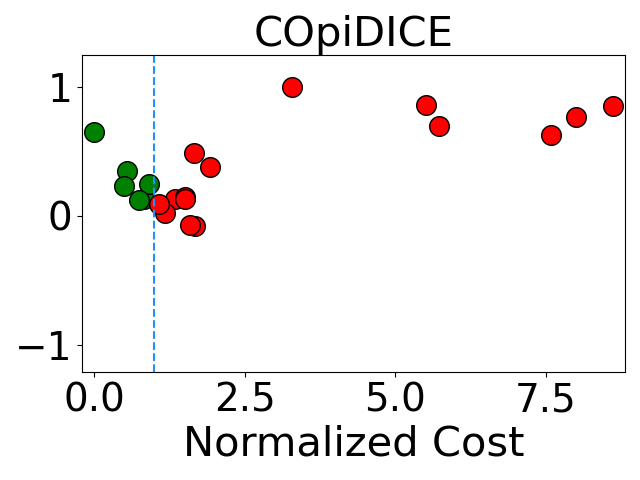}
        \end{subfigure}
        ~ 
        \begin{subfigure}[t]{0.19\textwidth}
            \centering
            \includegraphics[width=\textwidth]{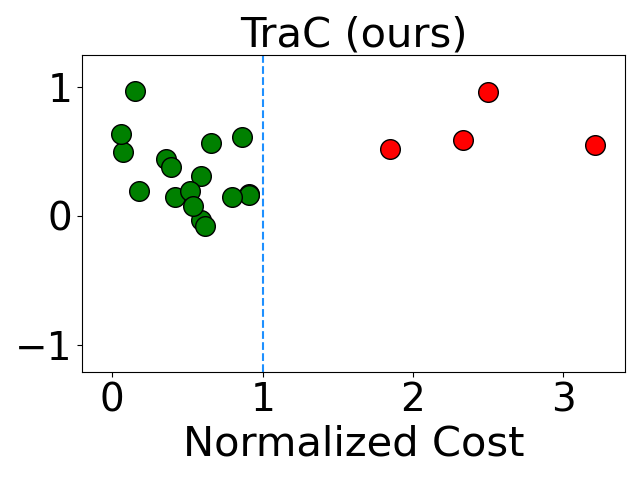}
        \end{subfigure}
    \end{minipage}
    \begin{minipage}{0.02\textwidth}
        \centering
        \rotatebox[origin=center]{90}{MetaDrive}
    \end{minipage}
    \begin{minipage}{0.97\textwidth}
        \centering
        \begin{subfigure}[t]{0.19\textwidth}
            \centering
            \includegraphics[width=\textwidth]{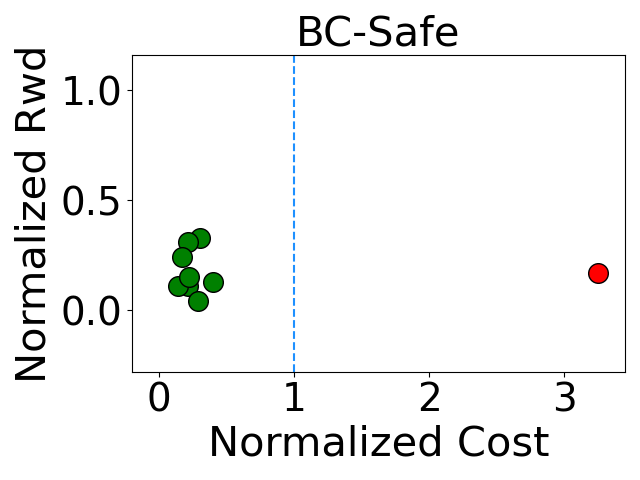}
        \end{subfigure}
        \begin{subfigure}[t]{0.19\textwidth}
            \centering
            \includegraphics[width=\textwidth]{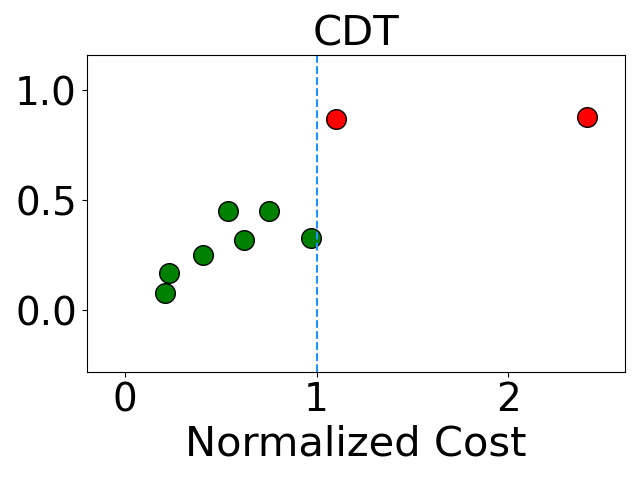}
        \end{subfigure}
        \begin{subfigure}[t]{0.19\textwidth}
            \centering
            \includegraphics[width=\textwidth]{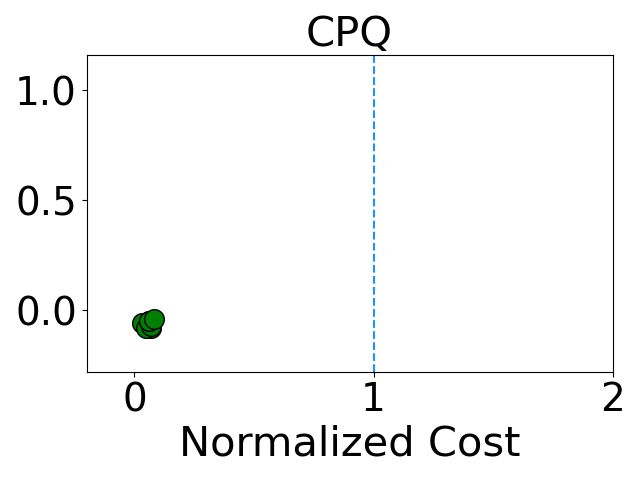}
        \end{subfigure}
        \begin{subfigure}[t]{0.19\textwidth}
            \centering
            \includegraphics[width=\textwidth]{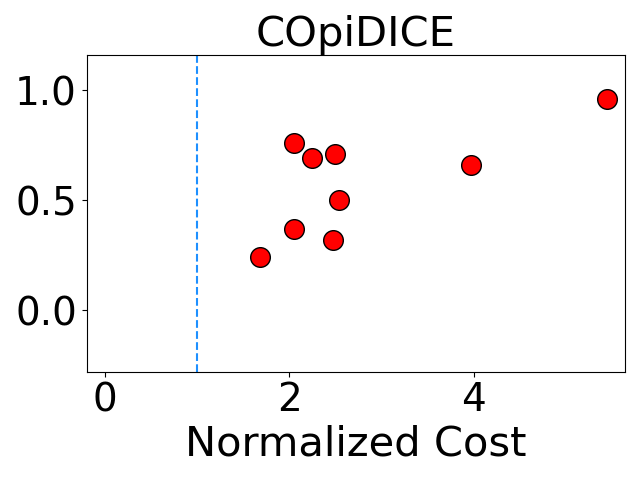}
        \end{subfigure}
        ~ 
        \begin{subfigure}[t]{0.19\textwidth}
            \centering
            \includegraphics[width=\textwidth]{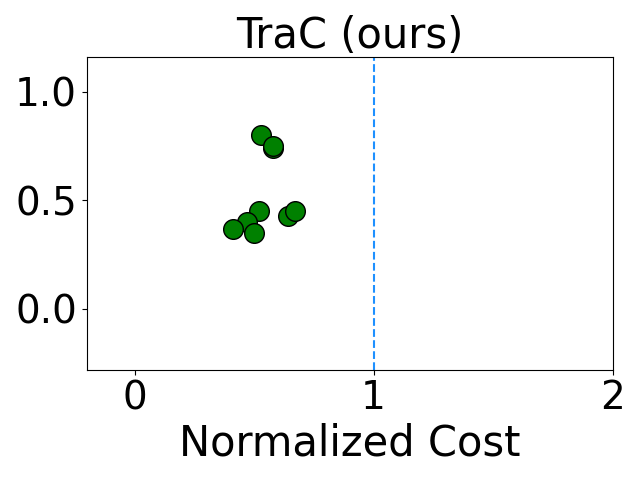}
        \end{subfigure}
    \end{minipage}
    \caption{Visualization of normalized reward and cost for each task. The dotted blue vertical lines represent the cost threshold of 1. Each round dot represents a task, with green indicating safety and red indicating constraint violation.}
    \label{fig:reward_cost_scatter}
\end{figure*}

\section{Evaluation}

In this section, we present experiments designed to evaluate the performance of $\texttt{TraC}$ in maximizing the cumulative rewards while satisfying safety constraints. We specifically address the following questions:
\begin{itemize}
    \item Can \texttt{TraC} effectively improve policy performance and strengthen safety constraints compared to state-of-the-art offline safe RL approaches?
    \item How do different compositions of desirable and undesirable datasets affect the performance of \texttt{TraC}?
    \item What ingredients of \texttt{TraC} are important for achieving high performance while adhering to safety constraints?
\end{itemize}

\subsection{Experimental Setup}

We conduct experiments using the well-established DSRL benchmark \cite{liu2023datasets}, specifically designed to evaluate offline safety RL approaches. The offline datasets were precollected for 38 tasks across three widely recognized environments, SafetyGymnasium \cite{ray2019benchmarking,ji2023omnisafe}, BulletSafetyGym \cite{gronauer2022bullet} and MetaDrive \cite{li2022metadrive}.  To assess performance, we adopt the constraint variation evaluation \cite{liu2023datasets} which is introduced in DSRL to evaluate algorithm versatility. Each algorithm is tested on each dataset using three distinct cost thresholds and three random seeds to ensure a fair comparison. We use normalized reward and normalized cost as evaluation metrics \cite{liu2023datasets,fu2020d4rl}, where a normalized cost
below 1 indicates safety. For the practical implementation of \texttt{TraC}, we first pretrain the policy using behavior cloning (BC) with the offline dataset, which we then maintain as the reference policy $\pi_\text{ref}$. We subsequently learn a policy using our method on the newly constructed desirable and undesirable datasets.

\subsection{Baselines}

We compare \texttt{TraC} with several state-of-the-art offline safe RL baselines to demonstrate the effectiveness of our approach: 
1) BC-All: Behavior cloning trained on the entire datasets. 2) BC-Safe: Behavior cloning trained exclusively on safe trajectories that satisfy the safety constraints. 3) CDT \cite{liu2023constrained}: A sequence modeling approach that incorporates safety constraints into Decision Transformer architecture. 4) BCQ-Lag: A Lagrangian-based method that incorporates safety constraints into BCQ \cite{fujimoto2019off}. 
5) BEAR-Lag: A Lagrangian-based method that incorporates safety constraints into BEAR \cite{kumar2019stabilizing}. 6) CPQ \cite{xu2022constraints}: a modified Q-learning approach that penalizes unsafe actions by treating them as out-of-distribution actions. 7) COptiDICE \cite{leecoptidice}: A DICE (distribution correction estimation) based safe offline RL method built on OptiDICE \cite{lee2021optidice}. 

\begin{table*}
  \centering
  \resizebox{\textwidth}{!}{
  \begin{tabular}{ccccccccccccccccc}
    \toprule
    \multirow{2}[2]{*}{Benchmark} & \multicolumn{2}{c}{BC-All} & \multicolumn{2}{c}{BC-Safe} & \multicolumn{2}{c}{CDT} & \multicolumn{2}{c}{BCQ-Lag} & \multicolumn{2}{c}{BEAR-Lag} & \multicolumn{2}{c}{CPQ} & \multicolumn{2}{c}{COptiDICE} & \multicolumn{2}{c}{TraC (ours)} \\
    \cmidrule(lr){2-3}
    \cmidrule(lr){4-5}
    \cmidrule(lr){6-7}
    \cmidrule(lr){8-9}
    \cmidrule(lr){10-11}
    \cmidrule(lr){12-13}
    \cmidrule(lr){14-15}
    \cmidrule(lr){16-17}
      &   reward$\uparrow$    &   cost$\downarrow$ &   reward$\uparrow$    &   cost$\downarrow$  &   reward$\uparrow$    &   cost$\downarrow$ &   reward$\uparrow$    &   cost$\downarrow$  &   reward$\uparrow$    &   cost$\downarrow$    &   reward$\uparrow$    &   cost$\downarrow$    &   reward$\uparrow$    &   cost$\downarrow$    &   reward$\uparrow$    &   cost$\downarrow$    \\
    \midrule
    SafetyGym   &   \textcolor{gray}{0.46}    &   \textcolor{gray}{3.03}    &   \textbf{0.34}    &   \textbf{0.75}    &   \textcolor{gray}{0.54}    &   \textcolor{gray}{1.06}    &   \textcolor{gray}{0.5} &   \textcolor{gray}{3.29}    &   \textcolor{gray}{0.39}    &   \textcolor{gray}{2.7} &   \textcolor{gray}{0.27}    &   \textcolor{gray}{2.79}    &   \textcolor{gray}{0.37}    &   \textcolor{gray}{2.65}    &   \textcolor{blue}{\textbf{0.4}} &   \textcolor{blue}{\textbf{0.92}}    \\
    BulletGym   &   \textcolor{gray}{0.64}    &   \textcolor{gray}{3.36}    &   \textbf{0.52}    &   \textbf{0.82}    &   \textcolor{gray}{0.68}    &   \textcolor{gray}{1.04}    &   \textcolor{gray}{0.74}    &   \textcolor{gray}{3.11}    &   \textcolor{gray}{0.48}    &   \textcolor{gray}{3.8} &   \textcolor{gray}{0.33}    &   \textcolor{gray}{1.12}    &   \textcolor{gray}{0.55}    &   \textcolor{gray}{2.55}    &   \textcolor{blue}{\textbf{0.61}}    &   \textcolor{blue}{\textbf{0.52}}    \\
    MetaDrive  &  \textcolor{gray}{0.42}    &   \textcolor{gray}{2.06}    &   \textbf{0.18}    &   \textbf{0.58}    &   \textbf{0.42}    &   \textbf{0.8} &   \textcolor{gray}{0.54}    &   \textcolor{gray}{2.05}    &   \textbf{0.02}    &   \textbf{0.39}    &   \textbf{-0.06}   &   \textbf{0.06}    &   \textcolor{gray}{0.58}    &   \textcolor{gray}{2.77}    &   \textcolor{blue}{\textbf{0.53}}    &   \textcolor{blue}{\textbf{0.55}}   \\
    \bottomrule
  \end{tabular}
  }
  \caption{Results of normalized reward and cost averaged over tasks in each environment. $\uparrow$ means the higher the better. $\downarrow$ means the lower the better. Each value is averaged over 3 distinct cost thresholds, 20 evaluation episodes, and 3 random seeds. \textbf{Bold}: Safe agents whose normalized cost is smaller than 1. \textcolor{gray}{Gray}: Unsafe agents. \textcolor{blue}{\textbf{Blue}}: Safe agents with the highest reward. The comprehensive results are provided in Table \ref{tab:result} in the Appendix.}
  \label{tab:average_result}
\end{table*}

\begin{figure}
    \centering
    \includegraphics[width=0.95\columnwidth]{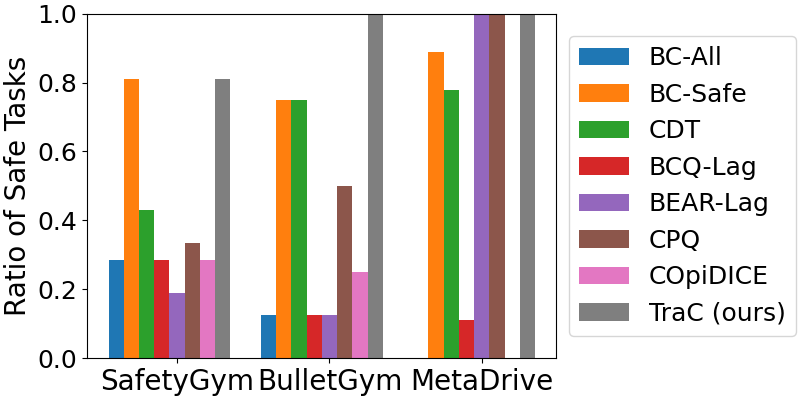}
    \caption{Ratio of tasks solved regarding safety.}
    \label{fig:safety_num}
\end{figure}

\subsection{Results}

\subsubsection{How does \texttt{TraC} perform against SOTA offline safe RL baselines?}

We conducted extensive experiments using DSRL and compared \texttt{TraC} against baselines. The main results are presented in Table \ref{tab:average_result}
, which shows the average performance of all approaches in terms of normalized reward and normalized cost across 38 tasks in three environments. \texttt{TraC} demonstrates satisfactory safety performance and high rewards in all environments, achieving the highest rewards in SafetyGym and BulletGym. In contrast, other baselines either suffer from low rewards or fail to meet safety constraints. On average, BC-All, BCQ-Lag, COptiDICE fail to learn safe policies in any of the three environments. While CDT, BEAR-Lag, and CPQ satisfy safety constraints in MetaDrive, they fail in the other two environments. BC-Safe also learns safe policies across all environments but results in conservative policies with low rewards. Additionally, although CDT can achieve high rewards across all three environments, it does so at the cost of violating safety constraints, further confirmed in Figure \ref{fig:safety_num} also.


Regarding safety constraint satisfaction, we present the proportion of tasks solved {safely} by each approach in Figure \ref{fig:safety_num}. \texttt{TraC} successfully learned the most safe policies for tasks across all three environments, especially achieving safety in all tasks within BulletGym and MetaDrive. While some baselines, such as BC-Safe and CDT, demonstrate relatively low average normalized costs (as noted in Table~\ref{tab:average_result}), but they still fail to learn policies that meet the required cost thresholds for several tasks as shown in Figure~\ref{fig:safety_num}.

Additionally, we illustrate the performance of each approach by visualizing their results in the normalized reward-cost space for each task in Figure \ref{fig:reward_cost_scatter}. For some baselines, such as CDT, CPQ, and COptiDICE, high average rewards, as shown in Table \ref{tab:average_result}, do not necessarily indicate good performance. This is because these rewards often stem from unsafe policies across various tasks. In particular, while CDT achieves the highest average reward in MetaDrive, this high average is due to two unsafe tasks that produce high rewards. 

\begin{figure}
    \centering
    \begin{subfigure}[t]{0.48\columnwidth}
        \centering
        \includegraphics[width=\textwidth]{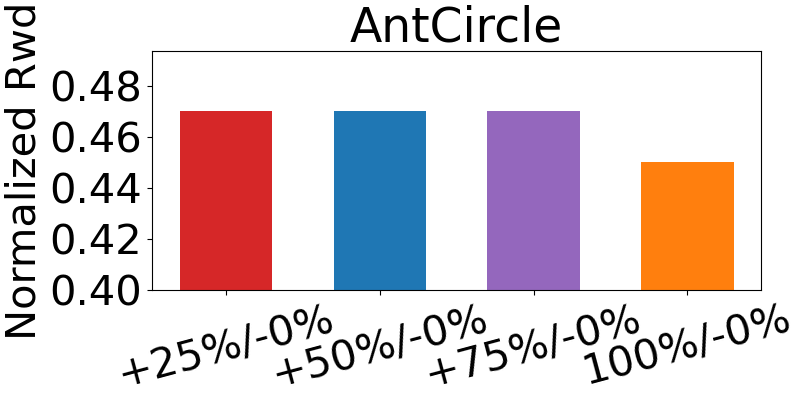}
    \end{subfigure}
    ~
    \begin{subfigure}[t]{0.48\columnwidth}
        \centering
        \includegraphics[width=\textwidth]{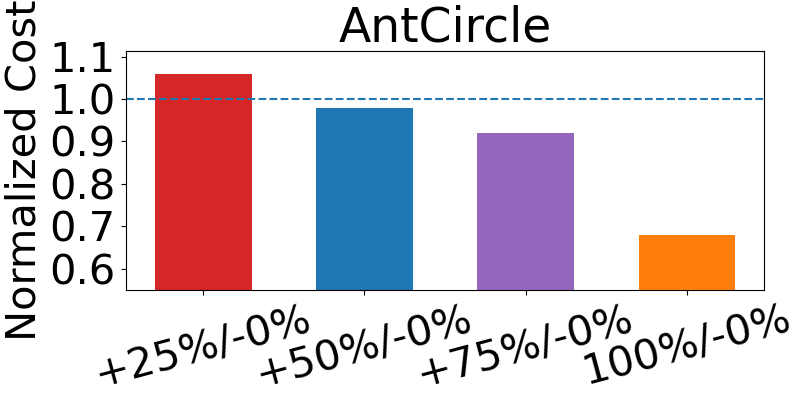}
    \end{subfigure}
    
    
    \begin{subfigure}[t]{0.48\columnwidth}
        \centering
        \includegraphics[width=\textwidth]{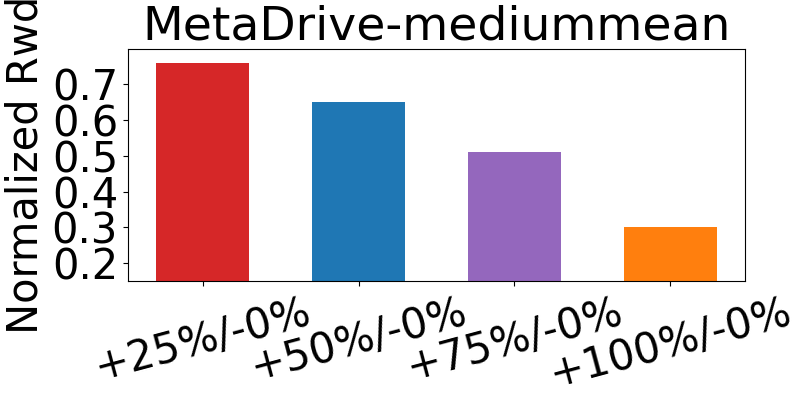}
    \end{subfigure}
    ~
    \begin{subfigure}[t]{0.48\columnwidth}
        \centering
        \includegraphics[width=\textwidth]{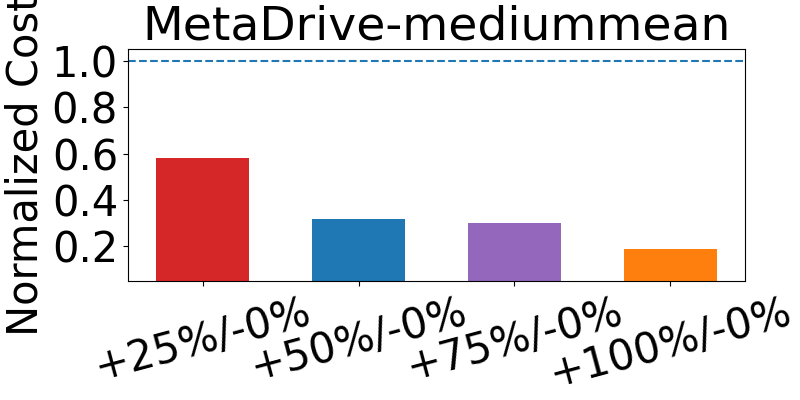}
    \end{subfigure}
    \caption{Results of normalized reward and normalized cost with various $x\%$ in two task.}
    \label{fig:x-percentage}
\end{figure}

\begin{figure}
    \centering
    \begin{subfigure}[t]{0.48\columnwidth}
        \centering
        \includegraphics[width=\textwidth]{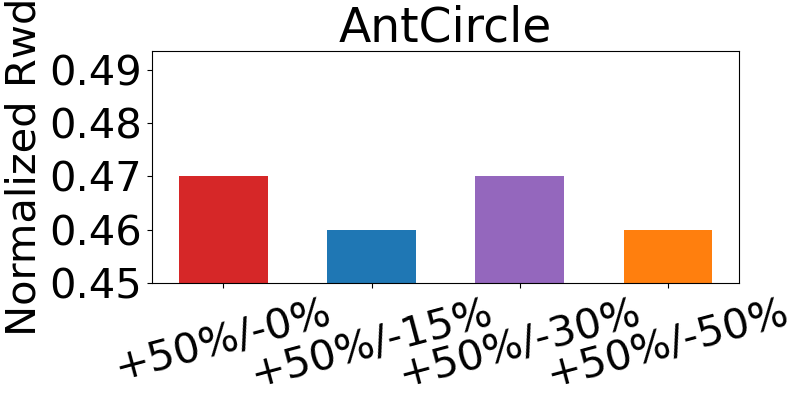}
    \end{subfigure}
    ~
    \begin{subfigure}[t]{0.48\columnwidth}
        \centering
        \includegraphics[width=\textwidth]{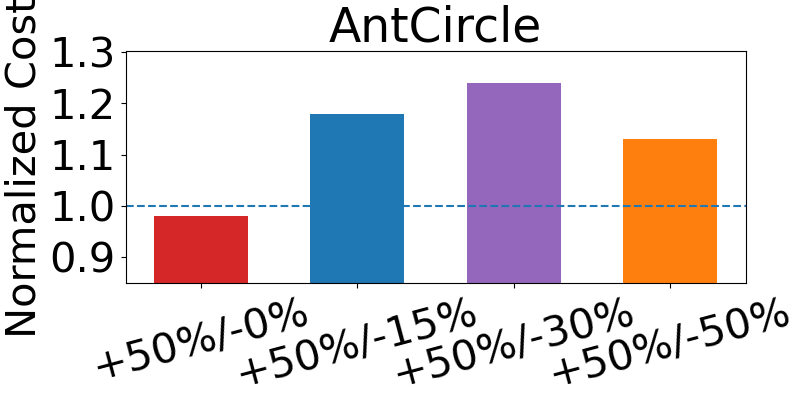}
    \end{subfigure}
    
    
    \begin{subfigure}[t]{0.48\columnwidth}
        \centering
        \includegraphics[width=\textwidth]{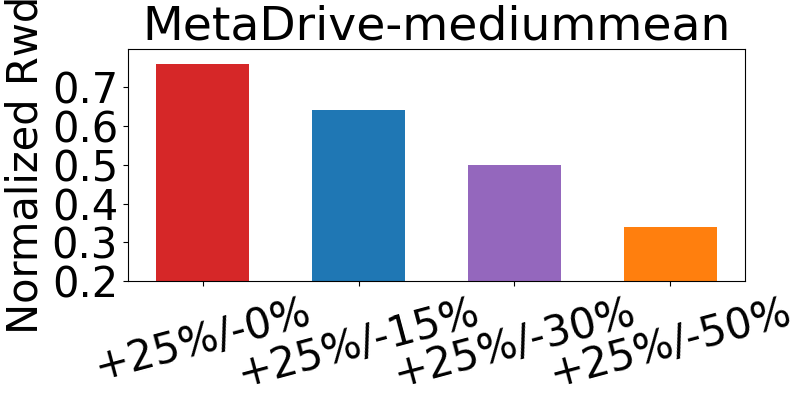}
    \end{subfigure}
    ~
    \begin{subfigure}[t]{0.48\columnwidth}
        \centering
        \includegraphics[width=\textwidth]{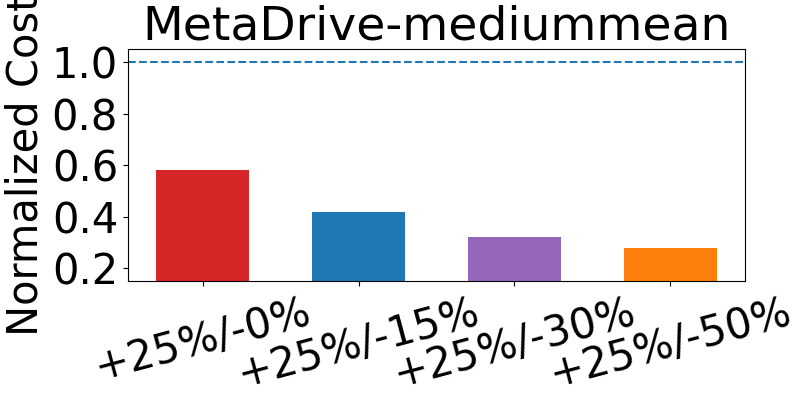}
    \end{subfigure}
    \caption{Results of normalized reward and normalized cost with various $y\%$ in two tasks.}
    \label{fig:y-percentage}
\end{figure}

\subsubsection{How do different compositions of desirable and undesirable datasets affect the performance of \texttt{TraC}?}

As a key component of \texttt{TraC}, the trajectory construction phase is expected to yield desirable and undesirable datasets that enable the learning of policies with high rewards while satisfying safety constraints. As outlined in Section \ref{sec:data_construction}, we select the top $x\%$ of safe trajectories as desirable, and the bottom $y\%$ as undesirable, based on their rewards. We conducted experiments with various selections of $x\%$ and $y\%$ to examine how different compositions influence the performance of $\texttt{TraC}$.

\begin{table*}
  \centering
  \resizebox{\textwidth}{!}{
  \begin{tabular}{ccccccccc|cccccccc}
    \toprule
    \multirow{2}[2]{*}{Task} & \multicolumn{2}{c}{$\delta=0$} & \multicolumn{2}{c}{$\delta=0.4$} & \multicolumn{2}{c}{$\delta=0.7$} & \multicolumn{2}{c}{$\delta=1.0$} & \multicolumn{2}{c}{$\eta=0.25$} & \multicolumn{2}{c}{$\eta=0.5$} & \multicolumn{2}{c}{$\eta=1.0$} & \multicolumn{2}{c}{$\eta=2.0$} \\
    \cmidrule(lr){2-3}
    \cmidrule(lr){4-5}
    \cmidrule(lr){6-7}
    \cmidrule(lr){8-9}
    \cmidrule(lr){10-11}
    \cmidrule(lr){12-13}
    \cmidrule(lr){14-15}
    \cmidrule(lr){16-17}
      &   reward$\uparrow$    &   cost$\downarrow$ &   reward$\uparrow$    &   cost$\downarrow$  &   reward$\uparrow$    &   cost$\downarrow$ &   reward$\uparrow$    &   cost$\downarrow$  &   reward$\uparrow$    &   cost$\downarrow$ &   reward$\uparrow$    &   cost$\downarrow$  &   reward$\uparrow$    &   cost$\downarrow$ &   reward$\uparrow$    &   cost$\downarrow$     \\
    \midrule
    AntCircle   &   0.46    &   0.87    &   0.47    &   0.95    &   0.47    &   0.98    &   0.46 &   0.9    &   0.47    &   0.98    &   0.47    &   0.98    &   0.46    &   1.18    &   0.46 &   0.92    \\
    CarPush1   &   0.18    &   0.18    &   0.2    &   0.14    &   0.2    &   0.22    &   0.14    &   0.2    &   0.21    &   0.33    &   0.2    &   0.22    &   0.16    &   0.29    &   0.14    &   0.17    \\
    mediummean  &  0.68    &   0.53    &   0.71    &   0.56    &   0.74    &   0.58 &   0.65    &   0.4   &  0.74    &   0.58    &   0.69    &   0.55    &   0.68    &   0.52 &   0.66    &   0.5   \\
    \bottomrule
  \end{tabular}
  }
  \caption{Ablation study of varying values of $\delta$ and $\eta$ in three tasks.}
  \label{tab:delta}
\end{table*}

In Figure \ref{fig:x-percentage}, we present the results of varying $x\%$ while keeping $y\%$ fixed in two tasks. On the x-axis, the notation $+x\%-y\%$ represents the experimental setting. In AntCircle task, the normalized reward remains consistent across difference $x\%$ selections, while the normalized cost decreases as $x\%$ increases. 
We also conducted experiments with varying $y\%$ while keeping $x\%$ fixed, as shown in Figure \ref{fig:y-percentage}. Altering $y\%$ results in minor changes in normalized reward, with the lowest normalized cost obtained when $y\%$ is set to $0\%$. While in MetaDrive-mediummean task, both the normalized reward and cost decrease as either $x\%$ or $y\%$ increases. These findings suggest that \texttt{TraC} is generally robust to varying selections of $x\%$ and $y\%$ though further fine-tuning for specific tasks may yield improved performance.

\begin{figure}
    \centering
    \begin{subfigure}[t]{0.85\columnwidth}
        \centering
        \includegraphics[width=\textwidth]{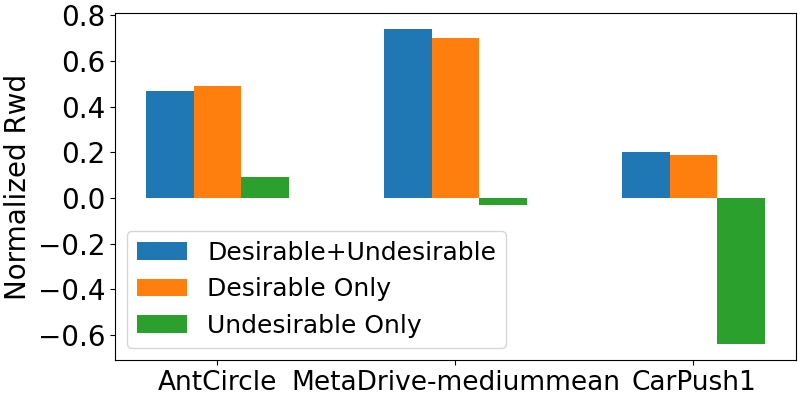}
    \end{subfigure}
    \smallskip
    \begin{subfigure}[t]{0.85\columnwidth}
        \centering
        \includegraphics[width=\textwidth]{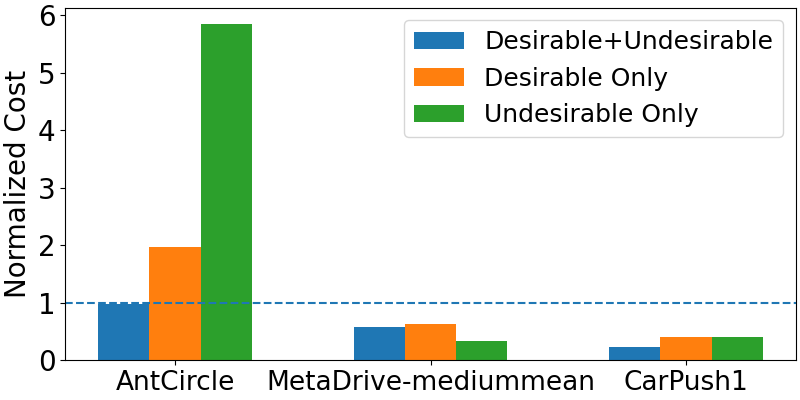}
    \end{subfigure}
    \caption{Ablation on training with only desirable or only undesirable trajectories.}
    \label{fig:only}
\end{figure}

As we partition the original offline dataset into two subsets, desirable and undesirable, and train a policy using both, the goal is to encourage behavior that aligns more closely with desirable trajectories while avoiding undesirable behaviors. To verify the effectiveness of using both the subsets for training, we also conducted experiments using only desirable or only undesirable trajectories. Figure \ref{fig:only} presents the ablation results across three tasks. Training with only undesirable trajectories typically results in a policy with very low rewards and, in some cases, very high costs, as seen in the AntCircle task. This outcome occurs because, with only undesirable trajectories, \texttt{TraC} attempts to learn a policy that avoids undesirable behavior, but lacks information about what desirable behavior looks like. Conversely, when training with only desirable trajectories, our algorithm learns a policy with high rewards and relatively low costs, as the agent is expected to imitate the desirable behaviors in those trajectories. However, solely learning from desirable trajectories limits the agent from imitating behaviors in the desirable dataset, potentially overlooking how to avoid undesirable actions. As shown in Figure \ref{fig:only}, training with both desirable and undesirable trajectories yields the best performance, as it allows the agent to learn both how to behave desirably and how to avoid undesirable behaviors.

\begin{table}[t]
  \centering
  \begin{tabular}{ccccc}
    \toprule
    \multirow{2}[2]{*}{Task} & \multicolumn{2}{c}{w/o $\pi_\text{ref}$ pretraining} & \multicolumn{2}{c}{\texttt{TraC}} \\
    \cmidrule(lr){2-3}
    \cmidrule(lr){4-5}
      &   reward$\uparrow$    &   cost$\downarrow$ &   reward$\uparrow$    &   cost$\downarrow$     \\
    \midrule
    AntCircle   &   0.0    &   \cellcolor{blue!10}0.0    &   \cellcolor{blue!10}0.47    &   0.98    \\
    AntRun   &   0.11    &   \cellcolor{blue!10}0.1    &   \cellcolor{blue!10}0.67    &   \cellcolor{blue!10}0.63    \\
    BallCircle  &  0.02    &   6.68    &   \cellcolor{blue!10}0.68    &   \cellcolor{blue!10}0.59      \\
    BallRun   &   0.26    &   \cellcolor{blue!10}0.15    &   \cellcolor{blue!10}0.27    &   0.47    \\
    CarCircle   &   -0.01    &   3.81    &   \cellcolor{blue!10}0.64    &   \cellcolor{blue!10}0.76    \\
    CarRun  &  0.66    &   2.62    &   \cellcolor{blue!10}0.97    &   \cellcolor{blue!10}0.03      \\
    DroneCircle   &   -0.27    &   1.79    &   \cellcolor{blue!10}0.6    &   \cellcolor{blue!10}0.67    \\
    DroneRun  &  0.16    &   6.53    &   \cellcolor{blue!10}0.55    &   \cellcolor{blue!10}0.01      \\
    \midrule
    Average  &  0.12    &   2.71    &   \cellcolor{blue!10}0.61    &   \cellcolor{blue!10}0.52      \\
    \bottomrule
  \end{tabular}
  \caption{Ablation on $\pi_\text{ref}$ pretraining for all task in BulletGym.}
  \label{tab:pretraining}
\end{table}

\subsection{Ablation Study}

We evaluate the sensitivity of hyperparameter selections of $\delta$ and $\eta$ in \texttt{TraC} to demonstrate its robustness and effectiveness. We tested four different values for each hyperparameter, and the results are shown in Table \ref{tab:delta}. The results indicate that \texttt{TraC} exhibits little sensitivity to the choice of $\delta$, maintaining robust performance across different selections. For $\eta$, we observe that the normalized reward decreases as $\eta$ increases while the normalized cost remains stable. That is because, a higher $\eta$, reduces the weights on desirable trajectories during training, potentially leading to a policy with lower rewards that does not strictly imitate desirable behaviors but instead explores how to avoid undesirable ones.

Furthermore, in the loss function of \texttt{TraC}, the reference policy $\pi_\text{ref}$ which is pretrained on the entire dataset {using behavior cloning}, serves as an approximation of the underlying behavioral policy of the offline dataset. Incorporating $\pi_\text{ref}$ helps mitigate the distributional drift issues in offline settings. To demonstrate the effectiveness of reference policy $\pi_\text{ref}$, we conduct an ablation study across all tasks in BulletGym as presented in Table \ref{tab:pretraining}. The results indicate a significant decrease in performance, both in normalized reward and normalized cost, when $\pi_\text{ref}$ pretraining is omitted. This demonstrates that a pretrained reference policy is crucial for learning a more effective policy.

\section{Conclusion}

In this paper, we propose a novel approach that reformulates the offline safe RL problem into a trajectory classification setting. Our goal is to develop policies that generate "desirable" trajectories and avoid "undesirable" ones. Specifically, the approach involves partitioning a pre-collected dataset of trajectories into desirable and undesirable subsets. The desirable subset contains high-reward, safe trajectories, while the undesirable subset includes low-reward safe and unsafe trajectories. We then optimize a policy using a classifier that evaluates each trajectory based on the learning policy. This approach bypasses the computational and stability challenges of traditional min-max objectives. Empirical results from the DSRL benchmark show that our method surpasses competitive baselines in terms of reward and constraint satisfaction across a range of tasks. Additional ablation studies further confirm the robustness and effectiveness of our approach.


\section*{Acknowledgement}
This research/project is supported by the National Research Foundation Singapore and DSO National Laboratories under the AI Singapore Programme (Award Number: AISG2-RP-2020-016).

{\small
\bibliography{aaai25}
}

\newpage
\appendix
\section{Experimental Details}

This section outlines the experimental details necessary for reproducing the experiments and the results reported in our paper.

\subsection{Task Description}

We conducted experiments using the well-established DSRL benchmark \cite{liu2023datasets}, which offers a comprehensive collection of datasets specifically designed to advance offline safe RL research. This benchmark includes 38 datasets spanning various safe RL environments and difficulty levels in SafetyGymnasium \cite{ray2019benchmarking,ji2023omnisafe}, BulletSafetyGym \cite{gronauer2022bullet}, and MetaDrive \cite{li2022metadrive}, all developed with safety considerations.

\begin{figure}
    \centering
    \includegraphics[width=0.8\columnwidth]{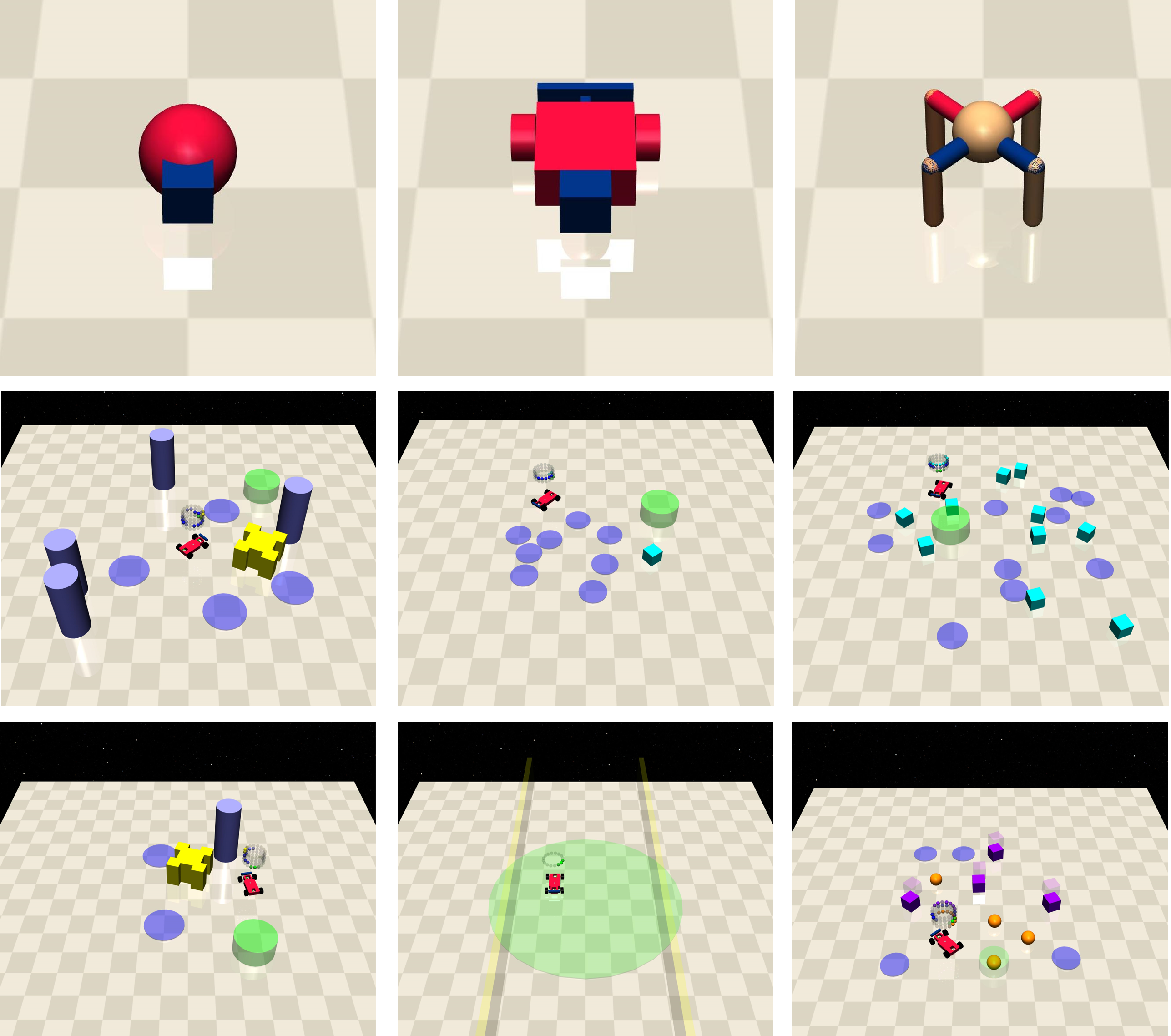}
    \caption{Visualization of agents and tasks in SafetyGymnasium.}
    \label{fig:safetgym}
\end{figure}

\begin{figure*}[h]
    \centering
    \subfloat[BulletSafetyGym]{
        \includegraphics[height=4.1cm]{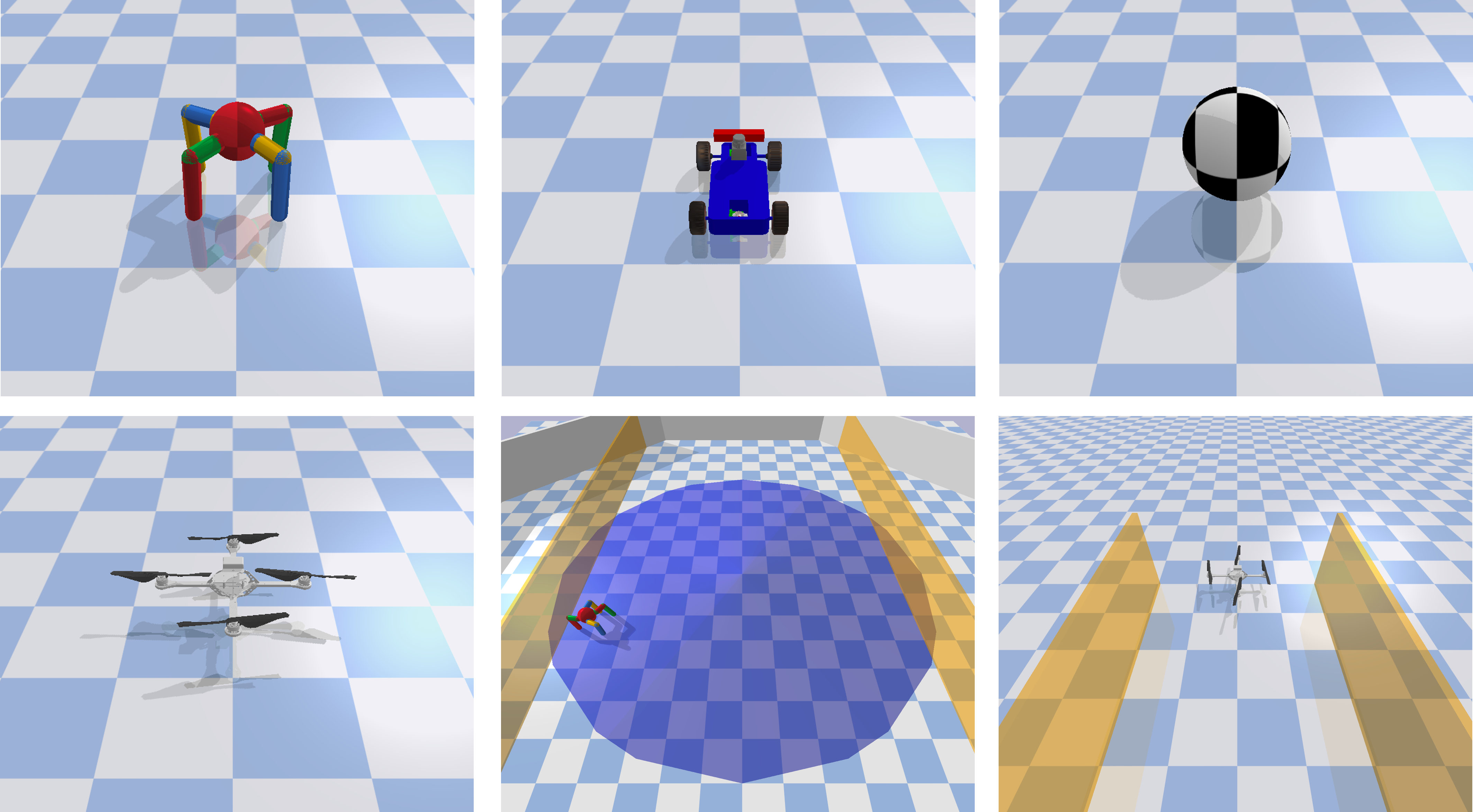}%
        \label{fig:bulletgym}
    }
    \quad
    \subfloat[MetaDrive]{
        \includegraphics[height=4.1cm]{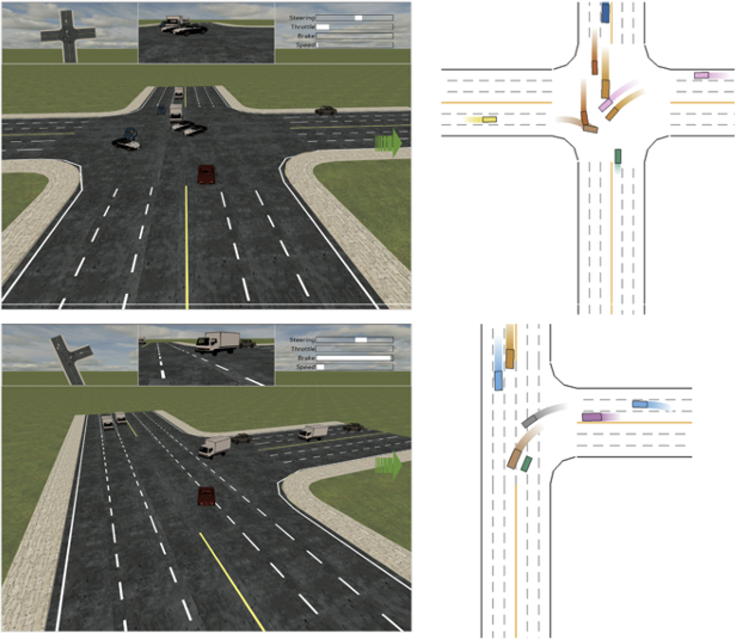}
        \label{fig:metadrive}
    }
    \caption{Visualization of agents and tasks in BulletSafetyGym and MetaDrive.}
\end{figure*}

\begin{itemize}
    \item \textbf{SafetyGymnasium} is a collection of environments built on the Mujoco physics simulator, offering a diverse range of tasks. It includes two types of agents, \texttt{Car} and \texttt{Point}, each with four tasks: \texttt{Button}, \texttt{Circle}, \texttt{Goal}, and \texttt{Push}. The tasks are further categorized by difficulty, indicated by the numbers \texttt{1} and \texttt{2}. In these tasks, agents must reach a goal while avoiding hazards, with task names formatted as \{\texttt{Agent}\}\{\texttt{Task}\}\{\texttt{Difficulty}\}. Additionally, SafetyGymnasium includes five velocity-constrained tasks for the agents, \texttt{Ant}, \texttt{HalfCheetah}, \texttt{Hopper}, \texttt{Walker2d}, and \texttt{Swimmer}. Figure \ref{fig:safetgym} provides a visualization of these tasks in SafetyGymnasium.
    \item \textbf{BulletSafetyGym} is a suite of environments developed using the PyBullet physics simulator, similar to SafetyGymnasium but featuring shorter horizons and a greater variety of agents. It includes four types of agents: \texttt{Ball}, \texttt{Car}, \texttt{Drone}, and \texttt{Ant}, each with two task types: \texttt{Circle} and \texttt{Run}. The tasks are named in the format \{\texttt{Agent}\}\{\texttt{Task}\}. Figure \ref{fig:bulletgym} illustrates the agents and tasks in BulletSafetyGym.
    \item \textbf{MetaDrive} is built on the Panda3D game engine \cite{goslin2004panda3d}, offering complex road conditions and dynamic scenarios that closely emulate real-world driving situations, making it ideal for evaluating safe RL algorithms. in high-stakes, realistic environments. The environments have three types of roads: \texttt{easy}, \texttt{medium}, and \texttt{hard}, each with varying levels of surrounding traffic vehicles: \texttt{sparse}, \texttt{mean}, and \texttt{dense}. The tasks are named in the format \{\texttt{Road}\}\{\texttt{Vehicle}\}. Figure \ref{fig:metadrive} provides a visualization of the tasks in MetaDrive.
\end{itemize}

\subsection{Evaluation Metrics}

To evaluate the algorithm's performance, we follow the approach used in DSRL \cite{liu2023datasets} and assess the algorithms using normalized reward and normalized cost. The normalized reward is calculated by
\begin{equation}
    R_\text{normalized} = \frac{R_\pi - r_\text{min}}{r_\text{max} - r_\text{min}} \nonumber
\end{equation}
where $R_\pi$ is the cumulative reward under policy $\pi$, and $r_\text{max}$ and $r_\text{min}$ are the empirical maximum and minimum reward returns, respectively. The normalized cost is written as:
\begin{equation}
    C_\text{normalized} = \frac{C_\pi + \epsilon}{\kappa + \epsilon} \nonumber
\end{equation}
where $C_\pi$ is the cumulative cost under policy $\pi$. The cost threshold is given by $\kappa$, and $\epsilon$ is a small positive number to ensure numerical stability if $\kappa=0$. According to the DSRL benchmark, the task is considered safe when $C_\text{normalized}$ does not exceed 1.

\begin{table*}
  \centering
  \begin{tabular}{cccc}
    \toprule
    Hyperparameters &   BulletGym   & SafetGym  &   MetaDrive     \\
    \midrule
    Training steps   &   100k    &   100k    &   200k    \\
    Pretraining steps   &   30k    &   30k    &   60k    \\
    Batch size  &  96    &   96    &   96      \\
    Policy network architecture   &   [256, 256] MLP    &   [256, 256] MLP    &   [256, 256] MLP  \\
    Policy network dropout   &   0.25    &   0.25    &   0.25    \\
    Optimizer   &   Adam    &   Adam    &   Adam    \\
    Learning rate   &   0.0001    &   0.0001    &   0.0001    \\
    Trajectory segment length ratio  &  1.0    &   0.75    &   0.25      \\
    Temperature $\alpha$  &  0.2    &   0.1    &   0.2      \\
    Discount factor $\gamma$  &  0.99    &   0.99    &   1.0      \\
    Trajectory weight lower bound $\delta$   &   0.7    &   0.7    &   0.7    \\
    Balancing factor $\eta$  &  0.25    &   0.5    &   0.25      \\
    Dataset construction factor $x\%$ and $y\%$   &   $50\%$ \& $0\%$  &   $50\%$ \& $50\%$ &   $25\%$ \& $0\%$   \\
    \bottomrule
  \end{tabular}
  \caption{Hyperparameters of \texttt{TraC} for tasks in three environments.}
  \label{tab:hyperparameters}
\end{table*}

\subsection{Training Details and Hyperparameters}

We adopt a two-step training procedure. First, we pretrain the policy using behavior cloning (BC) on the entire offline dataset, which serves as the reference policy $\pi_\text{ref}$. Next, we refine the policy by applying \texttt{TraC} on the newly created desirable and undesirable datasets.
The hyperparameters used in the experiments are summarized in Table \ref{tab:hyperparameters}. 
We assumed all policies to be Gaussian with fixed variance. Thus, we simply predicted actions using a standard MLP and computed the log probability $\log\pi(a|s)$ as $-\| \pi(s) - a \|^2_2$. The policy networks consist of two hidden layers with ReLU activation functions, each with 256 hidden units, and dropout is applied. We used the same batch size, optimizer, and learning rates for all tasks. Additionally, for different environments, we employ different values for the hyperparameters in \texttt{TraC}. 
In practical implementations of \texttt{TraC}, we may use trajectory segments instead of the full trajectory, as the latter can be too long and a shorter segment may yield better performance. For different environments, segments are sampled uniformly, and various length ratios are applied.

\section{Extended Results}

In this section, we present the comprehensive experimental results, including full evaluation results with other baselines in a table and illustrated in scatter plots, learning curves of \texttt{TraC} for all 38 tasks, and additional ablation studies.

\begin{figure*}
    \centering
    \begin{subfigure}[t]{0.23\textwidth}
        \centering
        \includegraphics[width=\textwidth]{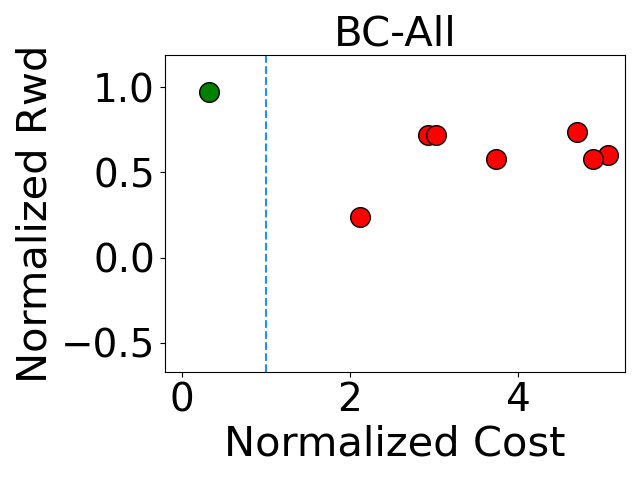}
    \end{subfigure}%
    ~ 
    \begin{subfigure}[t]{0.23\textwidth}
        \centering
        \includegraphics[width=\textwidth]{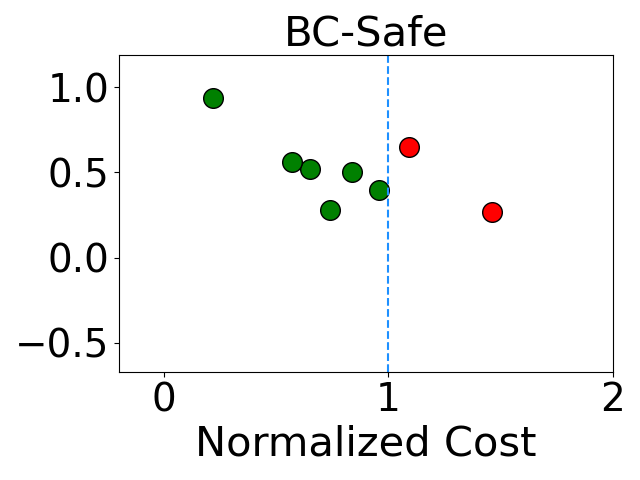}
    \end{subfigure}
    ~ 
    \begin{subfigure}[t]{0.23\textwidth}
        \centering
        \includegraphics[width=\textwidth]{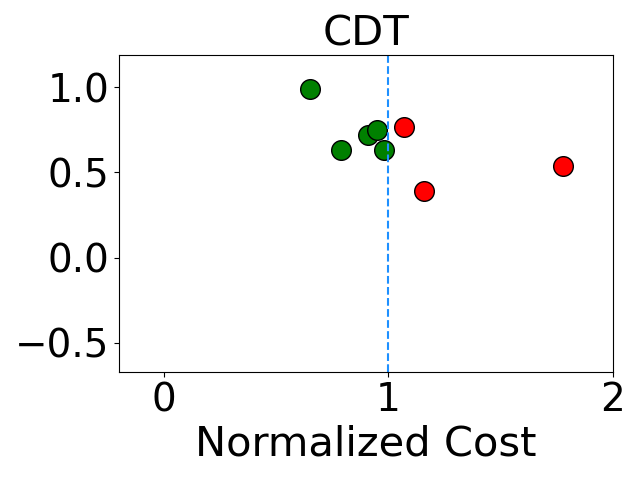}
    \end{subfigure}
    ~ 
    \begin{subfigure}[t]{0.23\textwidth}
        \centering
        \includegraphics[width=\textwidth]{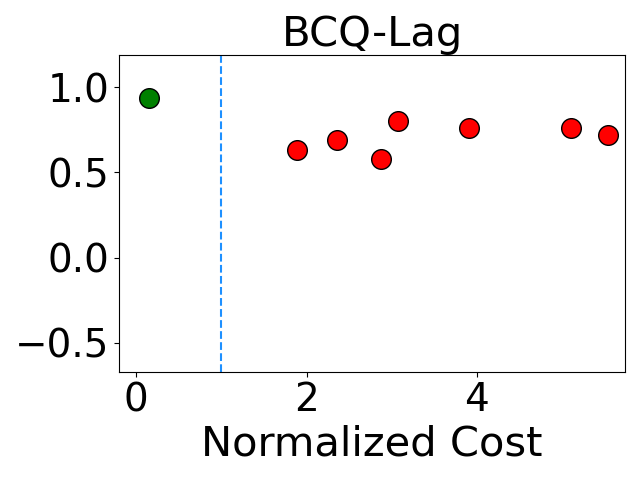}
    \end{subfigure}
    
    \begin{subfigure}[t]{0.23\textwidth}
        \centering
        \includegraphics[width=\textwidth]{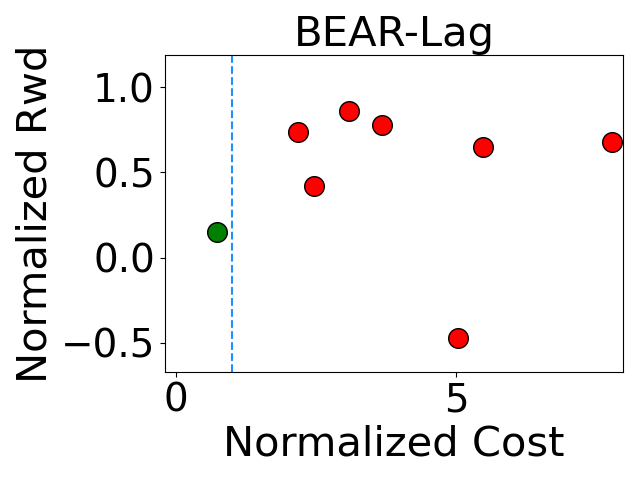}
    \end{subfigure}
    ~ 
    \begin{subfigure}[t]{0.23\textwidth}
        \centering
        \includegraphics[width=\textwidth]{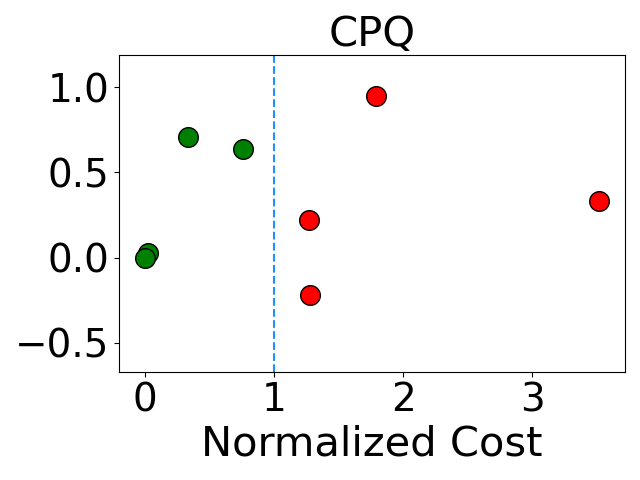}
    \end{subfigure}
    ~ 
    \begin{subfigure}[t]{0.23\textwidth}
        \centering
        \includegraphics[width=\textwidth]{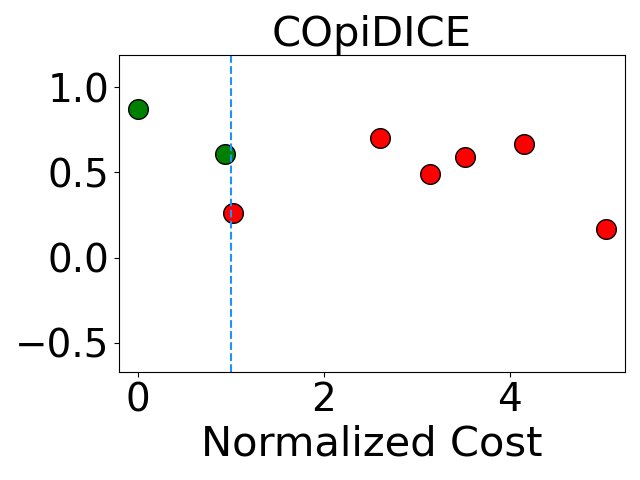}
    \end{subfigure}
    ~ 
    \begin{subfigure}[t]{0.23\textwidth}
        \centering
        \includegraphics[width=\textwidth]{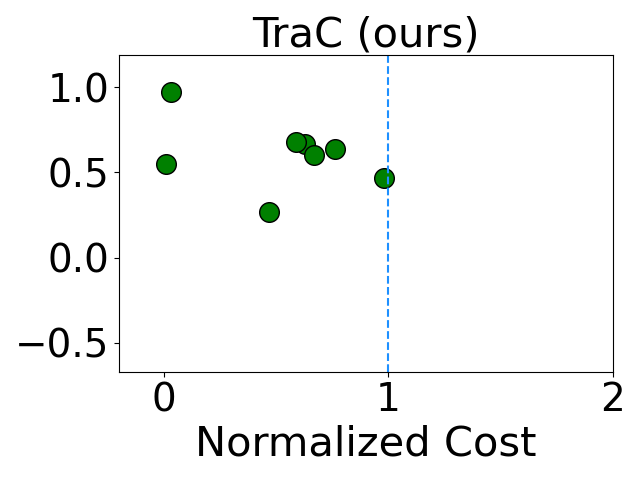}
    \end{subfigure}
    \caption{Results of normalized reward and cost for each task in BulletGym tasks. The dotted blue vertical lines represent the cost threshold of 1. Each round dot represents a task, with green indicating safety and red indicating constraint violation.}
    \label{fig:bullet_scatter}
\end{figure*}

\begin{figure*}
    \centering
    \begin{subfigure}[t]{0.23\textwidth}
        \centering
        \includegraphics[width=\textwidth]{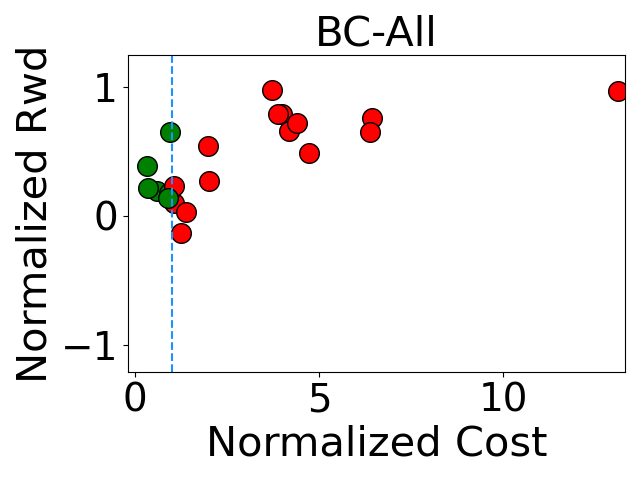}
    \end{subfigure}%
    ~ 
    \begin{subfigure}[t]{0.23\textwidth}
        \centering
        \includegraphics[width=\textwidth]{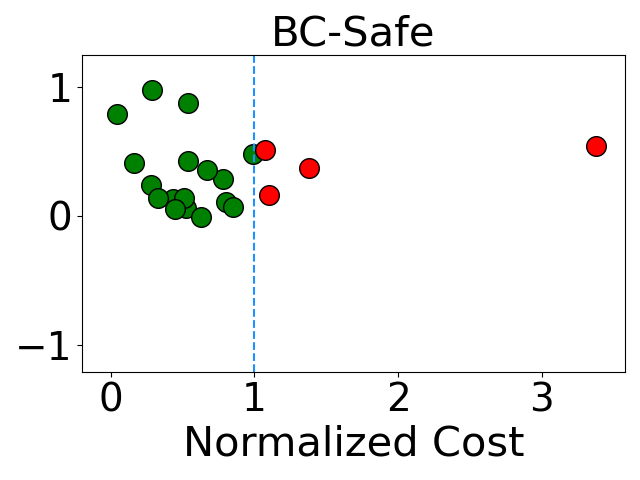}
    \end{subfigure}
    ~ 
    \begin{subfigure}[t]{0.23\textwidth}
        \centering
        \includegraphics[width=\textwidth]{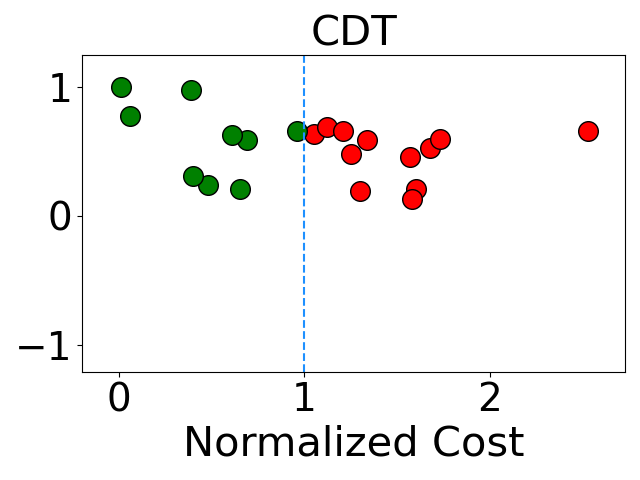}
    \end{subfigure}
    ~ 
    \begin{subfigure}[t]{0.23\textwidth}
        \centering
        \includegraphics[width=\textwidth]{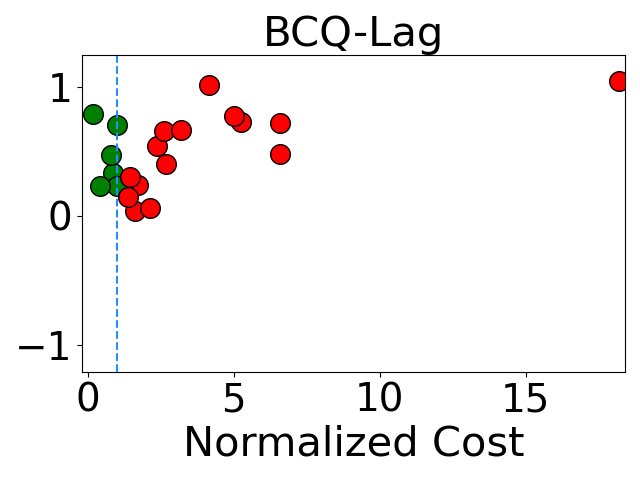}
    \end{subfigure}
    
    \begin{subfigure}[t]{0.23\textwidth}
        \centering
        \includegraphics[width=\textwidth]{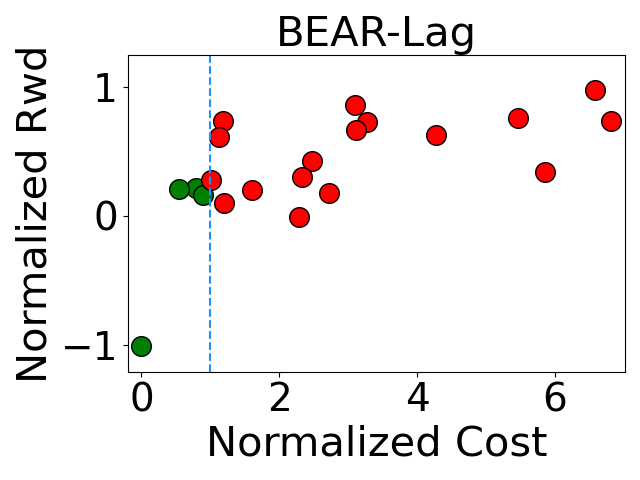}
    \end{subfigure}
    ~ 
    \begin{subfigure}[t]{0.23\textwidth}
        \centering
        \includegraphics[width=\textwidth]{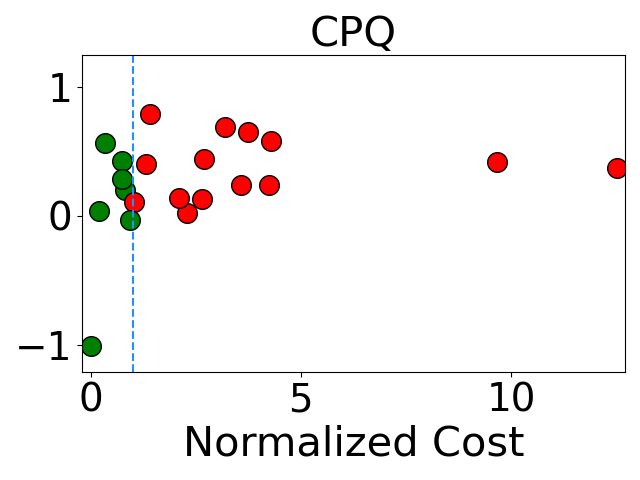}
    \end{subfigure}
    ~ 
    \begin{subfigure}[t]{0.23\textwidth}
        \centering
        \includegraphics[width=\textwidth]{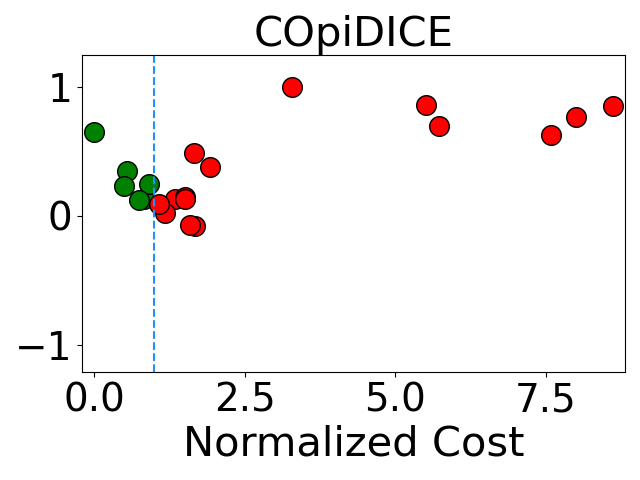}
    \end{subfigure}
    ~ 
    \begin{subfigure}[t]{0.23\textwidth}
        \centering
        \includegraphics[width=\textwidth]{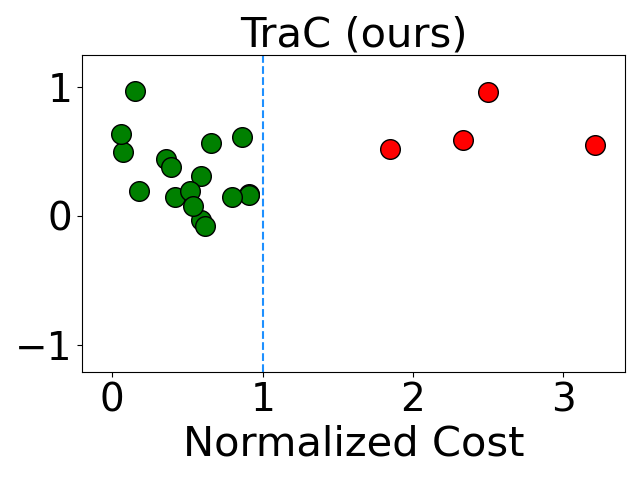}
    \end{subfigure}
    \caption{Results of normalized reward and cost for each task in SafetyGym tasks. The dotted blue vertical lines represent the cost threshold of 1. Each round dot represents a task, with green indicating safety and red indicating constraint violation.}
    \label{fig:safety_scatter}
\end{figure*}

\begin{figure*}
    \centering
    \begin{subfigure}[t]{0.23\textwidth}
        \centering
        \includegraphics[width=\textwidth]{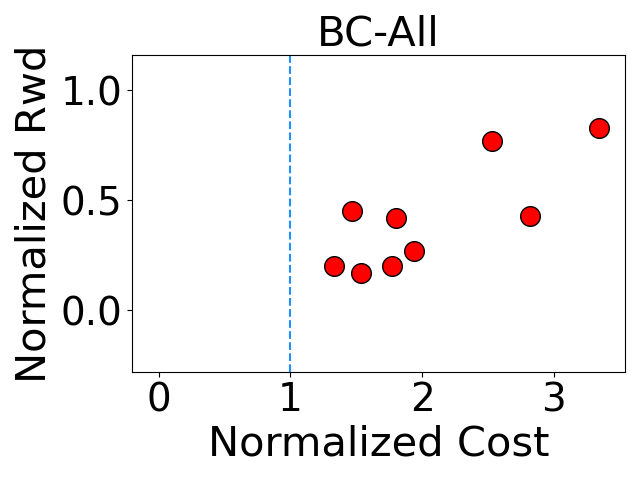}
    \end{subfigure}%
    ~ 
    \begin{subfigure}[t]{0.23\textwidth}
        \centering
        \includegraphics[width=\textwidth]{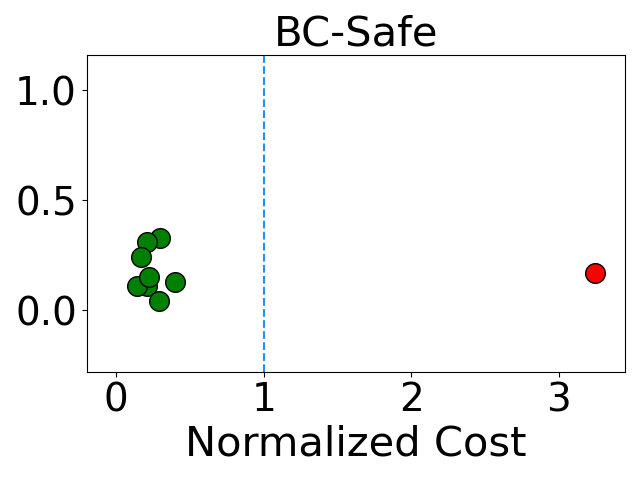}
    \end{subfigure}
    ~ 
    \begin{subfigure}[t]{0.23\textwidth}
        \centering
        \includegraphics[width=\textwidth]{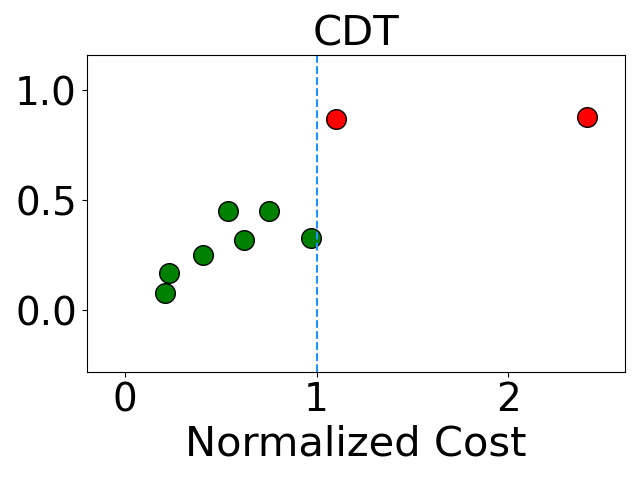}
    \end{subfigure}
    ~ 
    \begin{subfigure}[t]{0.23\textwidth}
        \centering
        \includegraphics[width=\textwidth]{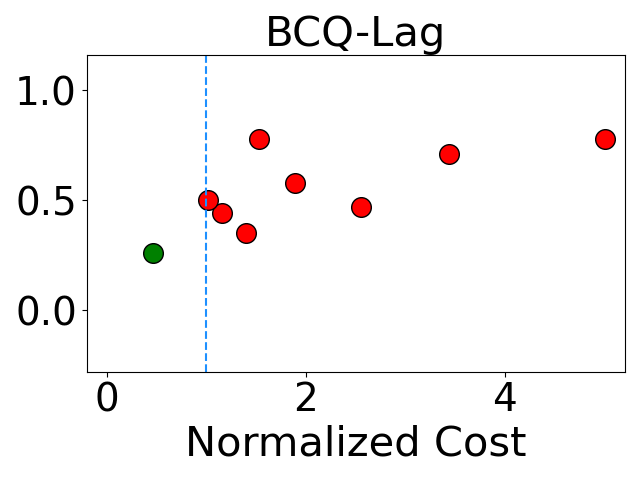}
    \end{subfigure}
    
    \begin{subfigure}[t]{0.23\textwidth}
        \centering
        \includegraphics[width=\textwidth]{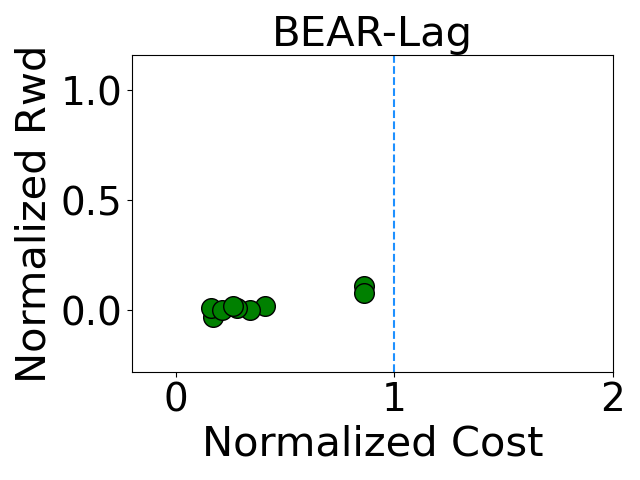}
    \end{subfigure}
    ~ 
    \begin{subfigure}[t]{0.23\textwidth}
        \centering
        \includegraphics[width=\textwidth]{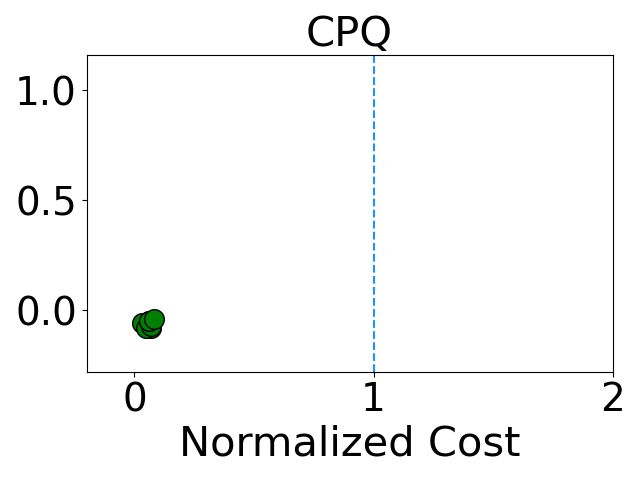}
    \end{subfigure}
    ~ 
    \begin{subfigure}[t]{0.23\textwidth}
        \centering
        \includegraphics[width=\textwidth]{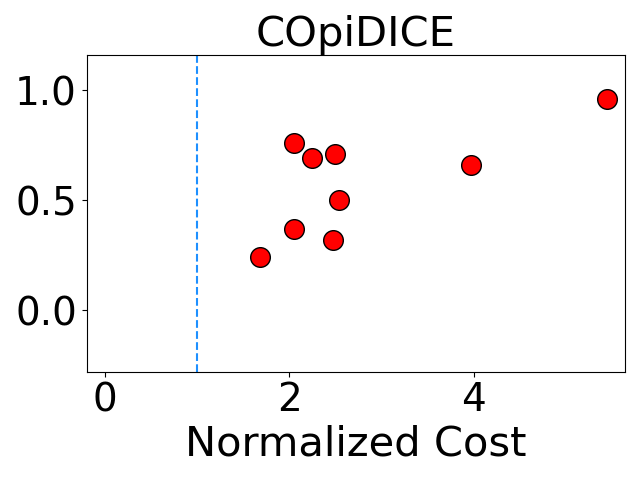}
    \end{subfigure}
    ~ 
    \begin{subfigure}[t]{0.23\textwidth}
        \centering
        \includegraphics[width=\textwidth]{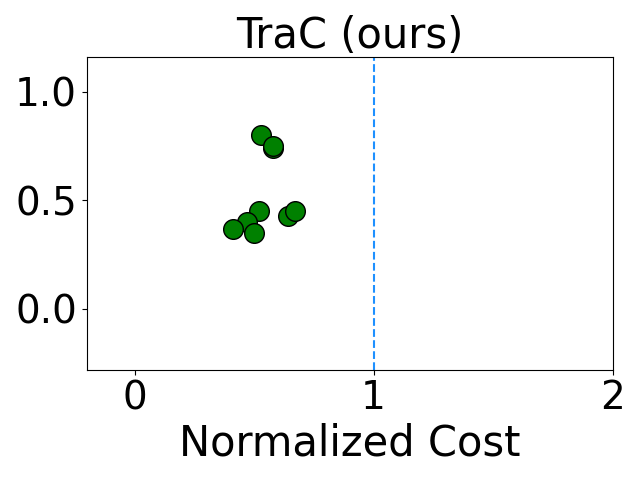}
    \end{subfigure}
    \caption{Results of normalized reward and cost for each task in MetaDrive. The dotted blue vertical lines represent the cost threshold of 1. Each round dot represents a task, with green indicating safety and red indicating constraint violation.}
    \label{fig:meta_scatter}
\end{figure*}

The full evaluation results are presented in Table \ref{tab:result}, where all approaches were evaluated using 3 distinct cost thresholds and 3 random seeds. BC-All generally learns policies that achieve high rewards but fails to ensure safety. BC-Safe, which applies behavior cloning specifically on safe trajectories, successfully learns safe policies but results in relatively low rewards across all three environments. CDT demonstrates balanced performance, learning policies that achieve high rewards while satisfying safety constraints for several tasks. However, it performs poorly in SafetyGym, with fewer safe policies being learned. BCQ-Lag achieves high rewards while maintaining safety for some tasks in SafetyGym, but fails to learn safe policies for most tasks across all three environments. Both BEAR-Lag and CPQ can learn highly safe policies for MetaDrive tasks, but at the cost of extremely low rewards, and they fail to learn safe policies for most tasks in SafetyGym and BulletGym, which may limit their applicability to safety-critical tasks. Similarly, COptiDICE struggles to learn policies that stay within the cost threshold for all tasks. Our approach, \texttt{TraC}, outperforms other baselines, learning policies that meet safety constraints for all tasks in BulletGym and MetaDrive, and for most tasks in SafetyGym, while achieving high rewards across all environments. Moreover, we note that CDT achieves the highest average rewards in MetaDrive tasks mainly due to the high rewards of two unsafe tasks, mediumsparse and mediumdense. The average reward drops to 0.29 when considering only the safe policies. We further illustrate the performance of each approach by visualizing their results in the normalized reward-cost space for each task in Figure \ref{fig:bullet_scatter}, \ref{fig:safety_scatter}, \ref{fig:meta_scatter}.

\begin{table*}
  \resizebox{\textwidth}{!}{
  \begin{tabular}{ccccccccccccccccc}
    \toprule
    \multirow{2}[2]{*}{Task} & \multicolumn{2}{c}{BC-All} & \multicolumn{2}{c}{BC-Safe} & \multicolumn{2}{c}{CDT} & \multicolumn{2}{c}{BCQ-Lag} & \multicolumn{2}{c}{BEAR-Lag} & \multicolumn{2}{c}{CPQ} & \multicolumn{2}{c}{COptiDICE} & \multicolumn{2}{c}{\texttt{TraC} (ours)} \\
    \cmidrule(lr){2-3}
    \cmidrule(lr){4-5}
    \cmidrule(lr){6-7}
    \cmidrule(lr){8-9}
    \cmidrule(lr){10-11}
    \cmidrule(lr){12-13}
    \cmidrule(lr){14-15}
    \cmidrule(lr){16-17}
      &   reward$\uparrow$    &   cost$\downarrow$ &   reward$\uparrow$    &   cost$\downarrow$  &   reward$\uparrow$    &   cost$\downarrow$ &   reward$\uparrow$    &   cost$\downarrow$  &   reward$\uparrow$    &   cost$\downarrow$    &   reward$\uparrow$    &   cost$\downarrow$    &   reward$\uparrow$    &   cost$\downarrow$    &   reward$\uparrow$    &   cost$\downarrow$    \\
    \midrule
    PointButton1    &   $\textcolor{gray}{0.1 \scriptstyle{\pm 0.06}}$   &   $\textcolor{gray}{1.05 \scriptstyle{\pm 0.39}}$   &   $\bm{0.06 \scriptstyle{\pm 0.04}}$   &   $\bm{0.52 \scriptstyle{\pm 0.21}}$   &   $\textcolor{gray}{0.53 \scriptstyle{\pm 0.01}}$   &   $\textcolor{gray}{1.68 \scriptstyle{\pm 0.13}}$   &   $\textcolor{gray}{0.24 \scriptstyle{\pm 0.04}}$   &   $\textcolor{gray}{1.73 \scriptstyle{\pm 1.11}}$    &   $\textcolor{gray}{0.2 \scriptstyle{\pm 0.04}}$ &   $\textcolor{gray}{1.6 \scriptstyle{\pm 0.99}}$ &   $\textcolor{gray}{0.69 \scriptstyle{\pm 0.05}}$    &   $\textcolor{gray}{3.2 \scriptstyle{\pm 1.57}}$ &   $\textcolor{gray}{0.13 \scriptstyle{\pm 0.02}}$    &   $\textcolor{gray}{1.35 \scriptstyle{\pm 0.91}}$  &   $\textcolor{blue}{\bm{0.17 \scriptstyle{\pm 0.08}}}$    &   $\textcolor{blue}{\bm{0.91 \scriptstyle{\pm 0.07}}}$    \\
    PointButton2    &   $\textcolor{gray}{0.27 \scriptstyle{\pm 0.08}}$    &   $\textcolor{gray}{2.02 \scriptstyle{\pm 0.38}}$    &   $\textcolor{gray}{0.16 \scriptstyle{\pm 0.04}}$    &   $\textcolor{gray}{1.1 \scriptstyle{\pm 0.84}}$ &   $\textcolor{gray}{0.46 \scriptstyle{\pm 0.01}}$    &   $\textcolor{gray}{1.57 \scriptstyle{\pm 0.1}}$    &   $\textcolor{gray}{0.4 \scriptstyle{\pm 0.03}}$ &   $\textcolor{gray}{2.66 \scriptstyle{\pm 1.47}}$    &   $\textcolor{gray}{0.43 \scriptstyle{\pm 0.05}}$    &   $\textcolor{gray}{2.47 \scriptstyle{\pm 1.17}}$    &   $\textcolor{gray}{0.58 \scriptstyle{\pm 0.07}}$    &   $\textcolor{gray}{4.3 \scriptstyle{\pm 2.35}}$ &   $\textcolor{gray}{0.15 \scriptstyle{\pm 0.03}}$    &   $\textcolor{gray}{1.51 \scriptstyle{\pm 0.96}}$    &   $\textcolor{blue}{\bm{0.16 \scriptstyle{\pm 0.05}}}$    &   $\textcolor{blue}{\bm{0.91 \scriptstyle{\pm 0.16}}}$    \\
    PointCircle1    &   $\textcolor{gray}{0.79 \scriptstyle{\pm 0.05}}$    &   $\textcolor{gray}{3.98 \scriptstyle{\pm 0.55}}$    &   $\bm{0.41 \scriptstyle{\pm 0.08}}$    &   $\bm{0.16 \scriptstyle{\pm 0.11}}$    &   $\textcolor{blue}{\bm{0.59 \scriptstyle{\pm 0.0}}}$    &   $\textcolor{blue}{\bm{0.69 \scriptstyle{\pm 0.04}}}$    &   $\textcolor{gray}{0.54 \scriptstyle{\pm 0.17}}$    &   $\textcolor{gray}{2.38 \scriptstyle{\pm 1.3}}$    &   $\textcolor{gray}{0.73 \scriptstyle{\pm 0.11}}$    &   $\textcolor{gray}{3.28 \scriptstyle{\pm 2.07}}$    &   $\bm{0.43 \scriptstyle{\pm 0.07}}$    &   $\bm{0.75 \scriptstyle{\pm 1.86}}$    &   $\textcolor{gray}{0.86 \scriptstyle{\pm 0.01}}$    &   $\textcolor{gray}{5.51 \scriptstyle{\pm 2.93}}$    &   $\bm{0.5 \scriptstyle{\pm 0.11}}$ &   $\bm{0.07 \scriptstyle{\pm 0.05}}$    \\
    PointCircle2    &   $\textcolor{gray}{0.66 \scriptstyle{\pm 0.09}}$    &   $\textcolor{gray}{4.17 \scriptstyle{\pm 0.72}}$    &   $\bm{0.48 \scriptstyle{\pm 0.08}}$    &   $\bm{0.99 \scriptstyle{\pm 0.35}}$    &   $\textcolor{gray}{0.64 \scriptstyle{\pm 0.01}}$    &   $\textcolor{gray}{1.05 \scriptstyle{\pm 0.08}}$    &   $\textcolor{gray}{0.66 \scriptstyle{\pm 0.13}}$    &   $\textcolor{gray}{2.6 \scriptstyle{\pm 0.71}}$ &   $\textcolor{gray}{0.63 \scriptstyle{\pm 0.27}}$    &   $\textcolor{gray}{4.27 \scriptstyle{\pm 1.48}}$    &   $\textcolor{gray}{0.24 \scriptstyle{\pm 0.4}}$    &   $\textcolor{gray}{3.58 \scriptstyle{\pm 3.09}}$    &   $\textcolor{gray}{0.85 \scriptstyle{\pm 0.01}}$    &   $\textcolor{gray}{8.61 \scriptstyle{\pm 4.62}}$    &   $\textcolor{blue}{\bm{0.61 \scriptstyle{\pm 0.01}}}$    &   $\textcolor{blue}{\bm{0.86 \scriptstyle{\pm 0.05}}}$    \\
    PointGoal1  &   $\bm{0.65 \scriptstyle{\pm 0.03}}$    &   $\bm{0.95 \scriptstyle{\pm 0.07}}$    &   $\bm{0.43 \scriptstyle{\pm 0.12}}$    &   $\bm{0.54 \scriptstyle{\pm 0.24}}$    &   $\textcolor{gray}{0.69 \scriptstyle{\pm 0.02}}$    &   $\textcolor{gray}{1.12 \scriptstyle{\pm 0.07}}$    &   $\textcolor{blue}{\bm{0.71 \scriptstyle{\pm 0.02}}}$    &   $\textcolor{blue}{\bm{0.98 \scriptstyle{\pm 0.46}}}$    &   $\textcolor{gray}{0.74 \scriptstyle{\pm 0.02}}$    &   $\textcolor{gray}{1.18 \scriptstyle{\pm 0.64}}$    &   $\bm{0.57 \scriptstyle{\pm 0.08}}$    &   $\bm{0.35 \scriptstyle{\pm 0.37}}$    &   $\textcolor{gray}{0.49 \scriptstyle{\pm 0.05}}$    &   $\textcolor{gray}{1.66 \scriptstyle{\pm 1.05}}$    &   $\bm{0.44 \scriptstyle{\pm 0.14}}$    &   $\bm{0.36 \scriptstyle{\pm 0.13}}$    \\
    PointGoal2  &   $\textcolor{gray}{0.54 \scriptstyle{\pm 0.03}}$    &   $\textcolor{gray}{1.97 \scriptstyle{\pm 0.24}}$    &   $\bm{0.29 \scriptstyle{\pm 0.09}}$    &   $\bm{0.78 \scriptstyle{\pm 0.27}}$    &   $\textcolor{gray}{0.59 \scriptstyle{\pm 0.03}}$    &   $\textcolor{gray}{1.34 \scriptstyle{\pm 0.05}}$    &   $\textcolor{gray}{0.67 \scriptstyle{\pm 0.06}}$    &   $\textcolor{gray}{3.18 \scriptstyle{\pm 1.79}}$    &   $\textcolor{gray}{0.67 \scriptstyle{\pm 0.03}}$    &   $\textcolor{gray}{3.11 \scriptstyle{\pm 1.76}}$    &   $\textcolor{gray}{0.4 \scriptstyle{\pm 0.15}}$ &   $\textcolor{gray}{1.31 \scriptstyle{\pm 0.71}}$    &   $\textcolor{gray}{0.38 \scriptstyle{\pm 0.03}}$    &   $\textcolor{gray}{1.92 \scriptstyle{\pm 1.15}}$    &   $\textcolor{blue}{\bm{0.31 \scriptstyle{\pm 0.06}}}$    &   $\textcolor{blue}{\bm{0.59 \scriptstyle{\pm 0.37}}}$    \\
    PointPush1  &   $\bm{0.19 \scriptstyle{\pm 0.07}}$    &   $\bm{0.61 \scriptstyle{\pm 0.05}}$    &   $\bm{0.13 \scriptstyle{\pm 0.05}}$    &   $\bm{0.43 \scriptstyle{\pm 0.29}}$    &   $\bm{0.24 \scriptstyle{\pm 0.02}}$    &   $\bm{0.48 \scriptstyle{\pm 0.05}}$    &   $\textcolor{blue}{\bm{0.33 \scriptstyle{\pm 0.04}}}$    &   $\textcolor{blue}{\bm{0.86 \scriptstyle{\pm 0.45}}}$    &   $\bm{0.22 \scriptstyle{\pm 0.04}}$    &   $\bm{0.79 \scriptstyle{\pm 0.39}}$    &   $\bm{0.2 \scriptstyle{\pm 0.08}}$ &   $\bm{0.83 \scriptstyle{\pm 0.44}}$    &   $\bm{0.13 \scriptstyle{\pm 0.02}}$    &   $\bm{0.83 \scriptstyle{\pm 0.52}}$    &   $\bm{0.15 \scriptstyle{\pm 0.01}}$    &   $\bm{0.42 \scriptstyle{\pm 0.12}}$    \\
    PointPush2  &   $\bm{0.18 \scriptstyle{\pm 0.02}}$    &   $\bm{0.91 \scriptstyle{\pm 0.1}}$    &   $\bm{0.11 \scriptstyle{\pm 0.04}}$    &   $\bm{0.8 \scriptstyle{\pm 0.59}}$ &   $\bm{0.21 \scriptstyle{\pm 0.04}}$    &   $\bm{0.65 \scriptstyle{\pm 0.03}}$    &   $\textcolor{blue}{\bm{0.23 \scriptstyle{\pm 0.03}}}$    &   $\textcolor{blue}{\bm{0.99 \scriptstyle{\pm 0.57}}}$    &   $\bm{0.16 \scriptstyle{\pm 0.05}}$    &   $\bm{0.89 \scriptstyle{\pm 0.59}}$    &   $\textcolor{gray}{0.11 \scriptstyle{\pm 0.14}}$    &   $\textcolor{gray}{1.04 \scriptstyle{\pm 0.61}}$    &   $\textcolor{gray}{0.02 \scriptstyle{\pm 0.07}}$    &   $\textcolor{gray}{1.18 \scriptstyle{\pm 0.74}}$    &   $\bm{0.15 \scriptstyle{\pm 0.07}}$    &   $\bm{0.8 \scriptstyle{\pm 0.08}}$ \\
    CarButton1  &   $\textcolor{gray}{0.03 \scriptstyle{\pm 0.1}}$    &   $\textcolor{gray}{1.38 \scriptstyle{\pm 0.41}}$    &   $\textcolor{blue}{\bm{0.07 \scriptstyle{\pm 0.03}}}$    &   $\textcolor{blue}{\bm{0.85 \scriptstyle{\pm 0.39}}}$    &   $\textcolor{gray}{0.21 \scriptstyle{\pm 0.02}}$    &   $\textcolor{gray}{1.6 \scriptstyle{\pm 0.12}}$ &   $\textcolor{gray}{0.04 \scriptstyle{\pm 0.05}}$    &   $\textcolor{gray}{1.63 \scriptstyle{\pm 0.59}}$    &   $\textcolor{gray}{0.18 \scriptstyle{\pm 0.05}}$    &   $\textcolor{gray}{2.72 \scriptstyle{\pm 2.23}}$    &   $\textcolor{gray}{0.42 \scriptstyle{\pm 0.05}}$    &   $\textcolor{gray}{9.66 \scriptstyle{\pm 5.71}}$    &   $\textcolor{gray}{-0.08 \scriptstyle{\pm 0.09}}$   &   $\textcolor{gray}{1.68 \scriptstyle{\pm 1.29}}$    &   $\bm{-0.03 \scriptstyle{\pm 0.0}}$   &   $\bm{0.59 \scriptstyle{\pm 0.18}}$    \\
    CarButton2  &   $\textcolor{gray}{-0.13 \scriptstyle{\pm 0.01}}$   &   $\textcolor{gray}{1.24 \scriptstyle{\pm 0.26}}$    &   $\textcolor{blue}{\bm{-0.01 \scriptstyle{\pm 0.02}}}$   &   $\textcolor{blue}{\bm{0.63 \scriptstyle{\pm 0.3}}}$    &   $\textcolor{gray}{0.13 \scriptstyle{\pm 0.01}}$    &   $\textcolor{gray}{1.58 \scriptstyle{\pm 0.01}}$    &   $\textcolor{gray}{0.06 \scriptstyle{\pm 0.05}}$    &   $\textcolor{gray}{2.13 \scriptstyle{\pm 1.19}}$    &   $\textcolor{gray}{-0.01 \scriptstyle{\pm 0.09}}$   &   $\textcolor{gray}{2.29 \scriptstyle{\pm 2.07}}$    &   $\textcolor{gray}{0.37 \scriptstyle{\pm 0.11}}$    &   $\textcolor{gray}{12.51 \scriptstyle{\pm 8.54}}$   &   $\textcolor{gray}{-0.07 \scriptstyle{\pm 0.06}}$   &   $\textcolor{gray}{1.59 \scriptstyle{\pm 1.1}}$    &   $\bm{-0.08 \scriptstyle{\pm 0.07}}$   &   $\bm{0.62 \scriptstyle{\pm 0.08}}$    \\
    CarCircle1  &   $\textcolor{gray}{0.72 \scriptstyle{\pm 0.01}}$    &   $\textcolor{gray}{4.39 \scriptstyle{\pm 0.1}}$    &   $\textcolor{gray}{0.37 \scriptstyle{\pm 0.1}}$    &   $\textcolor{gray}{1.38 \scriptstyle{\pm 0.44}}$    &   $\textcolor{gray}{0.6 \scriptstyle{\pm 0.01}}$ &   $\textcolor{gray}{1.73 \scriptstyle{\pm 0.04}}$    &   $\textcolor{gray}{0.73 \scriptstyle{\pm 0.02}}$    &   $\textcolor{gray}{5.25 \scriptstyle{\pm 2.76}}$    &   $\textcolor{gray}{0.76 \scriptstyle{\pm 0.04}}$    &   $\textcolor{gray}{5.46 \scriptstyle{\pm 2.62}}$    &   $\textcolor{gray}{0.02 \scriptstyle{\pm 0.15}}$    &   $\textcolor{gray}{2.29 \scriptstyle{\pm 2.13}}$    &   $\textcolor{gray}{0.7 \scriptstyle{\pm 0.02}}$ &   $\textcolor{gray}{5.72 \scriptstyle{\pm 3.04}}$    &   $\textcolor{gray}{0.52 \scriptstyle{\pm 0.04}}$    &   $\textcolor{gray}{1.85 \scriptstyle{\pm 0.38}}$    \\
    CarCircle2  &   $\textcolor{gray}{0.76 \scriptstyle{\pm 0.03}}$    &   $\textcolor{gray}{6.44 \scriptstyle{\pm 0.19}}$    &   $\textcolor{gray}{0.54 \scriptstyle{\pm 0.08}}$    &   $\textcolor{gray}{3.38 \scriptstyle{\pm 1.3}}$    &   $\textcolor{gray}{0.66 \scriptstyle{\pm 0.0}}$    &   $\textcolor{gray}{2.53 \scriptstyle{\pm 0.03}}$    &   $\textcolor{gray}{0.72 \scriptstyle{\pm 0.04}}$    &   $\textcolor{gray}{6.58 \scriptstyle{\pm 3.02}}$    &   $\textcolor{gray}{0.74 \scriptstyle{\pm 0.04}}$    &   $\textcolor{gray}{6.82 \scriptstyle{\pm 2.95}}$    &   $\textcolor{gray}{0.44 \scriptstyle{\pm 0.1}}$    &   $\textcolor{gray}{2.69 \scriptstyle{\pm 2.61}}$    &   $\textcolor{gray}{0.77 \scriptstyle{\pm 0.03}}$    &   $\textcolor{gray}{7.99 \scriptstyle{\pm 4.23}}$    &   $\textcolor{gray}{0.59 \scriptstyle{\pm 0.02}}$    &   $\textcolor{gray}{2.33 \scriptstyle{\pm 0.89}}$    \\
    CarGoal1    &   $\bm{0.39 \scriptstyle{\pm 0.04}}$    &   $\bm{0.33 \scriptstyle{\pm 0.12}}$    &   $\bm{0.24 \scriptstyle{\pm 0.08}}$    &   $\bm{0.28 \scriptstyle{\pm 0.11}}$    &   $\textcolor{gray}{0.66 \scriptstyle{\pm 0.01}}$    &   $\textcolor{gray}{1.21 \scriptstyle{\pm 0.17}}$    &   $\textcolor{blue}{\bm{0.47 \scriptstyle{\pm 0.05}}}$    &   $\textcolor{blue}{\bm{0.78 \scriptstyle{\pm 0.5}}}$    &   $\textcolor{gray}{0.61 \scriptstyle{\pm 0.04}}$    &   $\textcolor{gray}{1.13 \scriptstyle{\pm 0.61}}$    &   $\textcolor{gray}{0.79 \scriptstyle{\pm 0.07}}$    &   $\textcolor{gray}{1.42 \scriptstyle{\pm 0.81}}$    &   $\bm{0.35 \scriptstyle{\pm 0.06}}$    &   $\bm{0.54 \scriptstyle{\pm 0.33}}$    &   $\bm{0.38 \scriptstyle{\pm 0.17}}$    &   $\bm{0.39 \scriptstyle{\pm 0.19}}$    \\
    CarGoal2    &   $\textcolor{gray}{0.23 \scriptstyle{\pm 0.02}}$    &   $\textcolor{gray}{1.05 \scriptstyle{\pm 0.07}}$    &   $\bm{0.14 \scriptstyle{\pm 0.05}}$    &   $\bm{0.51 \scriptstyle{\pm 0.26}}$    &   $\textcolor{gray}{0.48 \scriptstyle{\pm 0.01}}$    &   $\textcolor{gray}{1.25 \scriptstyle{\pm 0.14}}$    &   $\textcolor{gray}{0.3 \scriptstyle{\pm 0.05}}$ &   $\textcolor{gray}{1.44 \scriptstyle{\pm 0.99}}$    &   $\textcolor{gray}{0.28 \scriptstyle{\pm 0.04}}$    &   $\textcolor{gray}{1.01 \scriptstyle{\pm 0.62}}$    &   $\textcolor{gray}{0.65 \scriptstyle{\pm 0.2}}$    &   $\textcolor{gray}{3.75 \scriptstyle{\pm 2.0}}$    &   $\textcolor{blue}{\bm{0.25 \scriptstyle{\pm 0.04}}}$    &   $\textcolor{blue}{\bm{0.91 \scriptstyle{\pm 0.41}}}$    &   $\bm{0.19 \scriptstyle{\pm 0.08}}$    &   $\bm{0.52 \scriptstyle{\pm 0.12}}$    \\
    CarPush1    &   $\bm{0.22 \scriptstyle{\pm 0.04}}$    &   $\bm{0.36 \scriptstyle{\pm 0.12}}$    &   $\bm{0.14 \scriptstyle{\pm 0.03}}$    &   $\bm{0.33 \scriptstyle{\pm 0.23}}$    &   $\textcolor{blue}{\bm{0.31 \scriptstyle{\pm 0.01}}}$    &   $\textcolor{blue}{\bm{0.4 \scriptstyle{\pm 0.1}}}$ &   $\bm{0.23 \scriptstyle{\pm 0.03}}$    &   $\bm{0.43 \scriptstyle{\pm 0.19}}$    &   $\bm{0.21 \scriptstyle{\pm 0.02}}$    &   $\bm{0.54 \scriptstyle{\pm 0.28}}$    &   $\bm{-0.03 \scriptstyle{\pm 0.24}}$   &   $\bm{0.95 \scriptstyle{\pm 0.53}}$    &   $\bm{0.23 \scriptstyle{\pm 0.04}}$    &   $\bm{0.5 \scriptstyle{\pm 0.4}}$ &   $\bm{0.19 \scriptstyle{\pm 0.03}}$    &   $\bm{0.18 \scriptstyle{\pm 0.07}}$    \\
    CarPush2    &   $\textcolor{blue}{\bm{0.14 \scriptstyle{\pm 0.03}}}$    &   $\textcolor{blue}{\bm{0.9 \scriptstyle{\pm 0.08}}}$ &   $\bm{0.05 \scriptstyle{\pm 0.02}}$    &   $\bm{0.45 \scriptstyle{\pm 0.19}}$    &   $\textcolor{gray}{0.19 \scriptstyle{\pm 0.01}}$    &   $\textcolor{gray}{1.3 \scriptstyle{\pm 0.16}}$ &   $\textcolor{gray}{0.15 \scriptstyle{\pm 0.02}}$    &   $\textcolor{gray}{1.38 \scriptstyle{\pm 0.68}}$    &   $\textcolor{gray}{0.1 \scriptstyle{\pm 0.02}}$ &   $\textcolor{gray}{1.2 \scriptstyle{\pm 0.98}}$ &   $\textcolor{gray}{0.24 \scriptstyle{\pm 0.06}}$    &   $\textcolor{gray}{4.25 \scriptstyle{\pm 2.44}}$    &   $\textcolor{gray}{0.09 \scriptstyle{\pm 0.02}}$    &   $\textcolor{gray}{1.07 \scriptstyle{\pm 0.69}}$    &   $\bm{0.08 \scriptstyle{\pm 0.04}}$    &   $\bm{0.54 \scriptstyle{\pm 0.13}}$    \\
    SwimmerVelocity &   $\textcolor{gray}{0.49 \scriptstyle{\pm 0.27}}$    &   $\textcolor{gray}{4.72 \scriptstyle{\pm 4.01}}$    &   $\textcolor{gray}{0.51 \scriptstyle{\pm 0.2}}$    &   $\textcolor{gray}{1.07 \scriptstyle{\pm 0.07}}$    &   $\textcolor{blue}{\bm{0.66 \scriptstyle{\pm 0.01}}}$    &   $\textcolor{blue}{\bm{0.96 \scriptstyle{\pm 0.08}}}$    &   $\textcolor{gray}{0.48 \scriptstyle{\pm 0.33}}$    &   $\textcolor{gray}{6.58 \scriptstyle{\pm 3.95}}$    &   $\textcolor{gray}{0.3 \scriptstyle{\pm 0.01}}$ &   $\textcolor{gray}{2.33 \scriptstyle{\pm 0.04}}$    &   $\textcolor{gray}{0.13 \scriptstyle{\pm 0.06}}$    &   $\textcolor{gray}{2.66 \scriptstyle{\pm 0.96}}$    &   $\textcolor{gray}{0.63 \scriptstyle{\pm 0.06}}$    &   $\textcolor{gray}{7.58 \scriptstyle{\pm 1.77}}$    &   $\textcolor{gray}{0.55 \scriptstyle{\pm 0.02}}$    &   $\textcolor{gray}{3.21 \scriptstyle{\pm 1.92}}$    \\
    HopperVelocity  &  $\textcolor{gray}{0.65 \scriptstyle{\pm 0.01}}$    &   $\textcolor{gray}{6.39 \scriptstyle{\pm 0.88}}$    &   $\bm{0.36 \scriptstyle{\pm 0.13}}$    &   $\bm{0.67 \scriptstyle{\pm 0.27}}$    &   $\textcolor{blue}{\bm{0.63 \scriptstyle{\pm 0.06}}}$    &   $\textcolor{blue}{\bm{0.61 \scriptstyle{\pm 0.08}}}$    &   $\textcolor{gray}{0.78 \scriptstyle{\pm 0.09}}$    &   $\textcolor{gray}{5.02 \scriptstyle{\pm 3.43}}$    &   $\textcolor{gray}{0.34 \scriptstyle{\pm 0.06}}$    &   $\textcolor{gray}{5.86 \scriptstyle{\pm 4.05}}$    &   $\textcolor{gray}{0.14 \scriptstyle{\pm 0.09}}$    &   $\textcolor{gray}{2.11 \scriptstyle{\pm 2.29}}$    &   $\textcolor{gray}{0.13 \scriptstyle{\pm 0.06}}$    &   $\textcolor{gray}{1.51 \scriptstyle{\pm 1.54}}$ &   $\bm{0.57 \scriptstyle{\pm 0.18}}$    &   $\bm{0.98 \scriptstyle{\pm 0.37}}$    \\
    HalfCheetahVelocity &   $\textcolor{gray}{0.97 \scriptstyle{\pm 0.02}}$    &   $\textcolor{gray}{13.1 \scriptstyle{\pm 8.3}}$    &   $\bm{0.88 \scriptstyle{\pm 0.03}}$    &   $\bm{0.54 \scriptstyle{\pm 0.63}}$    &   $\textcolor{blue}{\bm{1.0 \scriptstyle{\pm 0.01}}}$ &   $\textcolor{blue}{\bm{0.01 \scriptstyle{\pm 0.01}}}$    &   $\textcolor{gray}{1.05 \scriptstyle{\pm 0.07}}$    &   $\textcolor{gray}{18.21 \scriptstyle{\pm 8.29}}$   &   $\textcolor{gray}{0.98 \scriptstyle{\pm 0.03}}$    &   $\textcolor{gray}{6.58 \scriptstyle{\pm 4.03}}$    &   $\bm{0.29 \scriptstyle{\pm 0.14}}$    &   $\bm{0.74 \scriptstyle{\pm 0.19}}$    &   $\bm{0.65 \scriptstyle{\pm 0.01}}$    &   $\bm{0.0 \scriptstyle{\pm 0.0}}$ &   $\textcolor{gray}{0.96 \scriptstyle{\pm 0.02}}$    &   $\textcolor{gray}{2.5 \scriptstyle{\pm 0.6}}$ \\
    Walker2dVelocity    &   $\textcolor{gray}{0.79 \scriptstyle{\pm 0.28}}$    &   $\textcolor{gray}{3.88 \scriptstyle{\pm 3.38}}$    &   $\textcolor{blue}{\bm{0.79 \scriptstyle{\pm 0.05}}}$    &   $\textcolor{blue}{\bm{0.04 \scriptstyle{\pm 0.32}}}$    &   $\bm{0.78 \scriptstyle{\pm 0.09}}$    &   $\bm{0.06 \scriptstyle{\pm 0.34}}$    &   $\bm{0.79 \scriptstyle{\pm 0.01}}$    &   $\bm{0.17 \scriptstyle{\pm 0.06}}$    &   $\textcolor{gray}{0.86 \scriptstyle{\pm 0.04}}$    &   $\textcolor{gray}{3.1 \scriptstyle{\pm 0.65}}$ &   $\bm{0.04 \scriptstyle{\pm 0.05}}$    &   $\bm{0.21 \scriptstyle{\pm 0.09}}$    &   $\bm{0.12 \scriptstyle{\pm 0.01}}$    &   $\bm{0.74 \scriptstyle{\pm 0.07}}$    &   $\bm{0.64 \scriptstyle{\pm 0.09}}$    &   $\bm{0.06 \scriptstyle{\pm 0.04}}$    \\
    AntVelocity &   $\textcolor{gray}{0.98 \scriptstyle{\pm 0.01}}$    &   $\textcolor{gray}{3.72 \scriptstyle{\pm 1.46}}$    &   $\textcolor{blue}{\bm{0.98 \scriptstyle{\pm 0.01}}}$    &   $\textcolor{blue}{\bm{0.29 \scriptstyle{\pm 0.1}}}$    &   $\bm{0.98 \scriptstyle{\pm 0.0}}$    &   $\bm{0.39 \scriptstyle{\pm 0.12}}$    &   $\textcolor{gray}{1.02 \scriptstyle{\pm 0.01}}$    &   $\textcolor{gray}{4.15 \scriptstyle{\pm 1.63}}$    &   $\bm{-1.01 \scriptstyle{\pm 0.0}}$   &   $\bm{0.0 \scriptstyle{\pm 0.0}}$ &   $\bm{-1.01 \scriptstyle{\pm 0.0}}$   &   $\bm{0.0 \scriptstyle{\pm 0.0}}$ &   $\textcolor{gray}{1.0 \scriptstyle{\pm 0.0}}$ &   $\textcolor{gray}{3.28 \scriptstyle{\pm 2.01}}$    &   $\bm{0.97 \scriptstyle{\pm 0.01}}$    &   $\bm{0.15 \scriptstyle{\pm 0.03}}$    \\
    \midrule
    \textbf{SafetyGym Average}   &   $\textcolor{gray}{0.46 \scriptstyle{\pm 0.35}}$    &   $\textcolor{gray}{3.03 \scriptstyle{\pm 5.32}}$    &   $\bm{0.34 \scriptstyle{\pm 0.31}}$    &   $\bm{0.75 \scriptstyle{\pm 0.8}}$    &   $\textcolor{gray}{0.54 \scriptstyle{\pm 0.21}}$    &   $\textcolor{gray}{1.06 \scriptstyle{\pm 0.59}}$    &   $\textcolor{gray}{0.5 \scriptstyle{\pm 0.32}}$ &   $\textcolor{gray}{3.29 \scriptstyle{\pm 4.97}}$    &   $\textcolor{gray}{0.39 \scriptstyle{\pm 0.43}}$    &   $\textcolor{gray}{2.7 \scriptstyle{\pm 3.39}}$ &   $\textcolor{gray}{0.27 \scriptstyle{\pm 0.37}}$    &   $\textcolor{gray}{2.79 \scriptstyle{\pm 3.86}}$    &   $\textcolor{gray}{0.37 \scriptstyle{\pm 0.32}}$    &   $\textcolor{gray}{2.65 \scriptstyle{\pm 3.24}}$    &   $\textcolor{blue}{\bm{0.4 \scriptstyle{\pm 0.28}}}$ &   $\textcolor{blue}{\bm{0.92 \scriptstyle{\pm 0.89}}}$    \\
    \midrule
    BallRun &   $\textcolor{gray}{0.6 \scriptstyle{\pm 0.1}}$ &   $\textcolor{gray}{5.08 \scriptstyle{\pm 0.74}}$    &   $\textcolor{gray}{0.27 \scriptstyle{\pm 0.14}}$    &   $\textcolor{gray}{1.46 \scriptstyle{\pm 0.39}}$    &   $\textcolor{gray}{0.39 \scriptstyle{\pm 0.09}}$    &   $\textcolor{gray}{1.16 \scriptstyle{\pm 0.19}}$    &   $\textcolor{gray}{0.76 \scriptstyle{\pm 0.01}}$    &   $\textcolor{gray}{3.91 \scriptstyle{\pm 0.35}}$    &   $\textcolor{gray}{-0.47 \scriptstyle{\pm 0.0}}$   &   $\textcolor{gray}{5.03 \scriptstyle{\pm 0.0}}$    &   $\textcolor{gray}{0.22 \scriptstyle{\pm 0.0}}$    &   $\textcolor{gray}{1.27 \scriptstyle{\pm 0.12}}$    &   $0.59 \scriptstyle{\pm 0.0}$    &   $\textcolor{gray}{3.52 \scriptstyle{\pm 0.0}}$    &   $\textcolor{blue}{\bm{0.27 \scriptstyle{\pm 0.02}} }$    &   $\textcolor{blue}{\bm{0.49 \scriptstyle{\pm 0.13}}}$    \\
    CarRun  &   $\bm{0.97 \scriptstyle{\pm 0.02}}$    &   $\bm{0.33 \scriptstyle{\pm 0.05}}$    &   $\bm{0.94 \scriptstyle{\pm 0.0}}$    &   $\bm{0.22 \scriptstyle{\pm 0.02}}$    &   $\textcolor{blue}{\bm{0.99 \scriptstyle{\pm 0.01}}}$    &   $\textcolor{blue}{\bm{0.65 \scriptstyle{\pm 0.31}}}$    &   $\bm{0.94 \scriptstyle{\pm 0.01}}$    &   $\bm{0.15 \scriptstyle{\pm 0.91}}$    &   $\textcolor{gray}{0.68 \scriptstyle{\pm 0.01}}$    &   $7.78 \scriptstyle{\pm 0.09}$    &   $\textcolor{gray}{0.95 \scriptstyle{\pm 0.01}}$    &   $1.79 \scriptstyle{\pm 0.18}$    &   $\bm{0.87 \scriptstyle{\pm 0.0}}$    &   $\bm{0.0 \scriptstyle{\pm 0.0}}$ &   $\bm{0.97 \scriptstyle{\pm 0.0}}$    &   $\bm{0.0 \scriptstyle{\pm 0.01}}$    \\
    DroneRun    &   $\textcolor{gray}{0.24 \scriptstyle{\pm 0.02}}$    &   $\textcolor{gray}{2.13 \scriptstyle{\pm 0.62}}$    &   $\bm{0.28 \scriptstyle{\pm 0.25}}$    &   $\bm{0.74 \scriptstyle{\pm 0.97}}$    &   $\textcolor{blue}{\bm{0.63 \scriptstyle{\pm 0.04}}}$    &   $\textcolor{blue}{\bm{0.79 \scriptstyle{\pm 0.68}}}$    &   $\textcolor{gray}{0.72 \scriptstyle{\pm 0.12}}$    &   $\textcolor{gray}{5.54 \scriptstyle{\pm 0.81}}$    &   $\textcolor{gray}{0.42 \scriptstyle{\pm 0.1}}$    &   $\textcolor{gray}{2.47 \scriptstyle{\pm 0.34}}$    &   $\textcolor{gray}{0.33 \scriptstyle{\pm 0.1}}$    &   $\textcolor{gray}{3.52 \scriptstyle{\pm 0.58}}$    &   $\textcolor{gray}{0.67 \scriptstyle{\pm 0.02}}$    &   $\textcolor{gray}{4.15 \scriptstyle{\pm 0.1}}$    &   $\bm{0.55 \scriptstyle{\pm 0.0}}$    &   $\bm{0.01 \scriptstyle{\pm 0.02}}$    \\
    AntRun  &   $\textcolor{gray}{0.7 \scriptstyle{\pm 0.06}}$   &   $\textcolor{gray}{2.93 \scriptstyle{\pm 2.4}}$    &   $\textcolor{gray}{0.65 \scriptstyle{\pm 0.15}}$    &   $\textcolor{gray}{1.09 \scriptstyle{\pm 0.84}}$    &   $\textcolor{blue}{\bm{0.72 \scriptstyle{\pm 0.04}}}$    &   $\textcolor{blue}{\bm{0.91 \scriptstyle{\pm 0.42}}}$    &   $\textcolor{gray}{0.76 \scriptstyle{\pm 0.07}}$    &   $\textcolor{gray}{5.11 \scriptstyle{\pm 2.39}}$    &   $\bm{0.15 \scriptstyle{\pm 0.02}}$    &   $\bm{0.73 \scriptstyle{\pm 0.07}}$    &   $\bm{0.03 \scriptstyle{\pm 0.02}}$    &   $\bm{0.02 \scriptstyle{\pm 0.09}}$    &   $\bm{0.61 \scriptstyle{\pm 0.01}}$    &   $\bm{0.94 \scriptstyle{\pm 0.69}}$    &   $\bm{0.69 \scriptstyle{\pm 0.01}}$    &   $\bm{0.76 \scriptstyle{\pm 0.22}}$    \\
    BallCircle  &   $\textcolor{gray}{0.74 \scriptstyle{\pm 0.15}}$    &   $\textcolor{gray}{4.71 \scriptstyle{\pm 1.79}}$    &   $\bm{0.52 \scriptstyle{\pm 0.08}}$    &   $\bm{0.65 \scriptstyle{\pm 0.17}}$    &   $\textcolor{gray}{0.77 \scriptstyle{\pm 0.06}}$    &   $\textcolor{gray}{1.07 \scriptstyle{\pm 0.27}}$    &   $\textcolor{gray}{0.69 \scriptstyle{\pm 0.11}}$    &   $\textcolor{gray}{2.36 \scriptstyle{\pm 1.04}}$    &   $\textcolor{gray}{0.86 \scriptstyle{\pm 0.18}}$    &   $\textcolor{gray}{3.09 \scriptstyle{\pm 1.53}}$    &   $\textcolor{gray}{0.64 \scriptstyle{\pm 0.01}}$    &   $\textcolor{gray}{0.76 \scriptstyle{\pm 0.0}}$    &   $\textcolor{gray}{0.7 \scriptstyle{\pm 0.04}}$ &   $\textcolor{gray}{2.61 \scriptstyle{\pm 0.79}}$    &   $\textcolor{blue}{\bm{0.69 \scriptstyle{\pm 0.06}}}$    &   $\textcolor{blue}{\bm{0.66 \scriptstyle{\pm 0.12}}}$    \\
    CarCircle   &   $\textcolor{gray}{0.58 \scriptstyle{\pm 0.25}}$    &   $\textcolor{gray}{3.74 \scriptstyle{\pm 2.2}}$    &   $\bm{0.5 \scriptstyle{\pm 0.22}}$ &   $\bm{0.84 \scriptstyle{\pm 0.67}}$    &   $\textcolor{blue}{\bm{0.75 \scriptstyle{\pm 0.06}}}$    &   $\textcolor{blue}{\bm{0.95 \scriptstyle{\pm 0.61}}}$    &   $\textcolor{gray}{0.63 \scriptstyle{\pm 0.19}}$    &   $\textcolor{gray}{1.89 \scriptstyle{\pm 1.37}}$    &   $\textcolor{gray}{0.74 \scriptstyle{\pm 0.1}}$    &   $\textcolor{gray}{2.18 \scriptstyle{\pm 1.33}}$    &   $\bm{0.71 \scriptstyle{\pm 0.02}}$    &   $\bm{0.33 \scriptstyle{\pm 0.0}}$    &   $\textcolor{gray}{0.49 \scriptstyle{\pm 0.05}}$    &   $\textcolor{gray}{3.14 \scriptstyle{\pm 2.98}}$    &   $\bm{0.61 \scriptstyle{\pm 0.06}}$    &   $\bm{0.87 \scriptstyle{\pm 0.35}}$    \\
    DroneCircle &   $\textcolor{gray}{0.72 \scriptstyle{\pm 0.04}}$    &   $\textcolor{gray}{3.03 \scriptstyle{\pm 0.29}}$    &   $\bm{0.56 \scriptstyle{\pm 0.18}}$    &   $\bm{0.57 \scriptstyle{\pm 0.27}}$    &   $\textcolor{blue}{\bm{0.63 \scriptstyle{\pm 0.07}}}$    &   $\textcolor{blue}{\bm{0.98 \scriptstyle{\pm 0.27}}}$    &   $\textcolor{gray}{0.8 \scriptstyle{\pm 0.07}}$ &   $\textcolor{gray}{3.07 \scriptstyle{\pm 0.89}}$    &   $\textcolor{gray}{0.78 \scriptstyle{\pm 0.04}}$    &   $\textcolor{gray}{3.68 \scriptstyle{\pm 0.44}}$    &   $\textcolor{gray}{-0.22 \scriptstyle{\pm 0.05}}$   &   $\textcolor{gray}{1.28 \scriptstyle{\pm 0.97}}$    &   $\textcolor{gray}{0.26 \scriptstyle{\pm 0.03}}$    &   $\textcolor{gray}{1.02 \scriptstyle{\pm 0.46}}$    &   $\bm{0.6 \scriptstyle{\pm 0.05}}$ &   $\bm{0.66 \scriptstyle{\pm 0.15}}$    \\
    AntCircle   &   $\textcolor{gray}{0.58 \scriptstyle{\pm 0.19}}$    &   $\textcolor{gray}{4.9 \scriptstyle{\pm 3.55}}$ &   $\bm{0.4 \scriptstyle{\pm 0.16}}$ &   $\bm{0.96 \scriptstyle{\pm 2.67}}$    &   $\textcolor{gray}{0.54 \scriptstyle{\pm 0.2}}$    &   $\textcolor{gray}{1.78 \scriptstyle{\pm 4.33}}$    &   $\textcolor{gray}{0.58 \scriptstyle{\pm 0.25}}$    &   $\textcolor{gray}{2.87 \scriptstyle{\pm 3.08}}$    &   $\textcolor{gray}{0.65 \scriptstyle{\pm 0.2}}$    &   $\textcolor{gray}{5.48 \scriptstyle{\pm 3.33}}$    &   $\bm{0.0 \scriptstyle{\pm 0.0}}$ &   $\bm{0.0 \scriptstyle{\pm 0.0}}$ &   $\textcolor{gray}{0.17 \scriptstyle{\pm 0.1}}$    &   $\textcolor{gray}{5.04 \scriptstyle{\pm 6.74}}$    &   $\textcolor{blue}{\bm{0.49 \scriptstyle{\pm 0.03}}}$    &   $\textcolor{blue}{\bm{0.91 \scriptstyle{\pm 0.39}}}$ \\
    \midrule
    \textbf{BulletGym Average}   &   $\textcolor{gray}{0.64 \scriptstyle{\pm 0.25}}$    &   $\textcolor{gray}{3.36 \scriptstyle{\pm 3.31}}$    &   $\bm{0.52 \scriptstyle{\pm 0.27}}$    &   $\bm{0.82 \scriptstyle{\pm 1.27}}$    &   $\textcolor{gray}{0.68 \scriptstyle{\pm 0.19}}$    &   $\textcolor{gray}{1.04 \scriptstyle{\pm 1.65}}$    &   $\textcolor{gray}{0.74 \scriptstyle{\pm 0.25}}$    &   $\textcolor{gray}{3.11 \scriptstyle{\pm 3.55}}$    &   $\textcolor{gray}{0.48 \scriptstyle{\pm 0.27}}$    &   $\textcolor{gray}{3.8 \scriptstyle{\pm 3.95}}$ &   $\textcolor{gray}{0.33 \scriptstyle{\pm 0.29}}$    &   $\textcolor{gray}{1.12 \scriptstyle{\pm 1.85}}$    &   $\textcolor{gray}{0.55 \scriptstyle{\pm 0.24}}$    &   $\textcolor{gray}{2.55 \scriptstyle{\pm 3.62}}$    &   $\textcolor{blue}{\bm{0.61 \scriptstyle{\pm 0.19}}}$    &   $\textcolor{blue}{\bm{0.55 \scriptstyle{\pm 0.4}}}$    \\
    \midrule
    easysparse  &   $\textcolor{gray}{0.17 \scriptstyle{\pm 0.05}}$    &   $\textcolor{gray}{1.54 \scriptstyle{\pm 1.38}}$    &   $\bm{0.11 \scriptstyle{\pm 0.08}}$    &   $\bm{0.21 \scriptstyle{\pm 0.02}}$    &   $\bm{0.17 \scriptstyle{\pm 0.14}}$    &   $\bm{0.23 \scriptstyle{\pm 0.32}}$    &   $\textcolor{gray}{0.78 \scriptstyle{\pm 0.0}}$    &   $\textcolor{gray}{5.01 \scriptstyle{\pm 0.06}}$    &   $\bm{0.11 \scriptstyle{\pm 0.0}}$    &   $\bm{0.86 \scriptstyle{\pm 0.01}}$    &   $\bm{-0.06 \scriptstyle{\pm 0.0}}$   &   $\bm{0.07 \scriptstyle{\pm 0.02}}$    &   $\textcolor{gray}{0.96 \scriptstyle{\pm 0.02}}$    &   $\textcolor{gray}{5.44 \scriptstyle{\pm 0.27}}$    &   $\textcolor{blue}{\bm{0.45 \scriptstyle{\pm 0.22}}}$    &   $\textcolor{blue}{\bm{0.52 \scriptstyle{\pm 0.34}}}$    \\
    easymean    &   $\textcolor{gray}{0.43 \scriptstyle{\pm 0.02}}$    &   $\textcolor{gray}{2.82 \scriptstyle{\pm 0.0}}$    &   $\bm{0.04 \scriptstyle{\pm 0.03}}$    &   $\bm{0.29 \scriptstyle{\pm 0.02}}$    &   $\textcolor{blue}{\bm{0.45 \scriptstyle{\pm 0.11}}}$    &   $\textcolor{blue}{\bm{0.54 \scriptstyle{\pm 0.55}}}$    &   $\textcolor{gray}{0.71 \scriptstyle{\pm 0.06}}$    &   $\textcolor{gray}{3.44 \scriptstyle{\pm 0.35}}$    &   $\bm{0.08 \scriptstyle{\pm 0.0}}$    &   $\bm{0.86 \scriptstyle{\pm 0.01}}$    &   $\bm{-0.07 \scriptstyle{\pm 0.0}}$   &   $\bm{0.07 \scriptstyle{\pm 0.01}}$    &   $\textcolor{gray}{0.66 \scriptstyle{\pm 0.16}}$    &   $\textcolor{gray}{3.97 \scriptstyle{\pm 1.47}}$    &   $\bm{0.4 \scriptstyle{\pm 0.21}}$    &   $\bm{0.47 \scriptstyle{\pm 0.36}}$    \\
    easydense   &   $\textcolor{gray}{0.27 \scriptstyle{\pm 0.14}}$    &   $\textcolor{gray}{1.94 \scriptstyle{\pm 1.18}}$    &   $\bm{0.11 \scriptstyle{\pm 0.07}}$    &   $\bm{0.14 \scriptstyle{\pm 0.01}}$    &   $\bm{0.32 \scriptstyle{\pm 0.18}}$    &   $\bm{0.62 \scriptstyle{\pm 0.43}}$    &   $\bm{0.26 \scriptstyle{\pm 0.0}}$    &   $\bm{0.47 \scriptstyle{\pm 0.01}}$    &   $\bm{0.02 \scriptstyle{\pm 0.05}}$    &   $\bm{0.41 \scriptstyle{\pm 0.22}}$    &   $\bm{-0.06 \scriptstyle{\pm 0.0}}$   &   $\bm{0.03 \scriptstyle{\pm 0.01}}$    &   $\textcolor{gray}{0.5 \scriptstyle{\pm 0.1}}$ &   $\textcolor{gray}{2.54 \scriptstyle{\pm 0.53}}$    &   $\textcolor{blue}{\bm{0.37 \scriptstyle{\pm 0.08}}}$    &   $\textcolor{blue}{\bm{0.41 \scriptstyle{\pm 0.21}}}$    \\
    mediumsparse    &   $\textcolor{gray}{0.83 \scriptstyle{\pm 0.13}}$    &   $\textcolor{gray}{3.34 \scriptstyle{\pm 0.58}}$    &   $\bm{0.33 \scriptstyle{\pm 0.34}}$    &   $\bm{0.3 \scriptstyle{\pm 0.32}}$ &   $\textcolor{gray}{0.87 \scriptstyle{\pm 0.11}}$    &   $\textcolor{gray}{1.1 \scriptstyle{\pm 0.26}}$ &   $\textcolor{gray}{0.44 \scriptstyle{\pm 0.0}}$    &   $\textcolor{gray}{1.16 \scriptstyle{\pm 0.02}}$    &   $\bm{-0.03 \scriptstyle{\pm 0.0}}$   &   $\bm{0.17 \scriptstyle{\pm 0.02}}$    &   $\bm{-0.08 \scriptstyle{\pm 0.02}}$   &   $\bm{0.07 \scriptstyle{\pm 0.03}}$    &   $\textcolor{gray}{0.71 \scriptstyle{\pm 0.37}}$    &   $\textcolor{gray}{2.49 \scriptstyle{\pm 1.9}}$    &   $\textcolor{blue}{\bm{0.8 \scriptstyle{\pm 0.08}}}$    &   $\textcolor{blue}{\bm{0.53 \scriptstyle{\pm 0.22}}}$    \\
    mediummean  &   $\textcolor{gray}{0.77 \scriptstyle{\pm 0.21}}$    &   $\textcolor{gray}{2.53 \scriptstyle{\pm 0.83}}$    &   $\bm{0.31 \scriptstyle{\pm 0.06}}$    &   $\bm{0.21 \scriptstyle{\pm 0.0}}$    &   $\bm{0.45 \scriptstyle{\pm 0.39}}$    &   $\bm{0.75 \scriptstyle{\pm 0.83}}$    &   $\textcolor{gray}{0.78 \scriptstyle{\pm 0.12}}$    &   $\textcolor{gray}{1.53 \scriptstyle{\pm 0.21}}$    &   $\bm{-0.0 \scriptstyle{\pm 0.0}}$    &   $\bm{0.34 \scriptstyle{\pm 0.03}}$    &   $\bm{-0.08 \scriptstyle{\pm 0.0}}$   &   $\bm{0.05 \scriptstyle{\pm 0.02}}$    &   $\textcolor{gray}{0.76 \scriptstyle{\pm 0.34}}$    &   $\textcolor{gray}{2.05 \scriptstyle{\pm 0.92}}$    &   $\textcolor{blue}{\bm{0.74 \scriptstyle{\pm 0.12}}}$    &   $\textcolor{blue}{\bm{0.58 \scriptstyle{\pm 0.14}}}$    \\
    mediumdense &   $\textcolor{gray}{0.45 \scriptstyle{\pm 0.27}}$    &   $\textcolor{gray}{1.47 \scriptstyle{\pm 1.65}}$    &   $\bm{0.24 \scriptstyle{\pm 0.0}}$    &   $\bm{0.17 \scriptstyle{\pm 0.0}}$    &   $\textcolor{gray}{0.88 \scriptstyle{\pm 0.12}}$    &   $\textcolor{gray}{2.41 \scriptstyle{\pm 0.71}}$    &   $\textcolor{gray}{0.58 \scriptstyle{\pm 0.21}}$    &   $\textcolor{gray}{1.89 \scriptstyle{\pm 1.19}}$    &   $\bm{0.01 \scriptstyle{\pm 0.02}}$    &   $\bm{0.28 \scriptstyle{\pm 0.16}}$    &   $\bm{-0.07 \scriptstyle{\pm 0.0}}$   &   $\bm{0.07 \scriptstyle{\pm 0.01}}$    &   $\textcolor{gray}{0.69 \scriptstyle{\pm 0.13}}$    &   $\textcolor{gray}{2.24 \scriptstyle{\pm 0.65}}$    &   $\textcolor{blue}{\bm{0.75 \scriptstyle{\pm 0.1}}}$    &   $\textcolor{blue}{\bm{0.58 \scriptstyle{\pm 0.15}}}$    \\
    hardsparse  &   $\textcolor{gray}{0.42 \scriptstyle{\pm 0.15}}$    &   $\textcolor{gray}{1.8 \scriptstyle{\pm 1.69}}$ &   $\textcolor{gray}{0.17 \scriptstyle{\pm 0.05}}$    &   $\textcolor{gray}{3.25 \scriptstyle{\pm 0.1}}$    &   $\bm{0.25 \scriptstyle{\pm 0.08}}$    &   $\bm{0.41 \scriptstyle{\pm 0.33}}$    &   $\textcolor{gray}{0.5 \scriptstyle{\pm 0.04}}$ &   $\textcolor{gray}{1.02 \scriptstyle{\pm 0.05}}$    &   $\bm{0.01 \scriptstyle{\pm 0.0}}$    &   $\bm{0.16 \scriptstyle{\pm 0.02}}$    &   $\bm{-0.05 \scriptstyle{\pm 0.0}}$   &   $\bm{0.06 \scriptstyle{\pm 0.01}}$    &   $\textcolor{gray}{0.37 \scriptstyle{\pm 0.1}}$    &   $\textcolor{gray}{2.05 \scriptstyle{\pm 0.27}}$    & $\textcolor{blue}{\bm{0.43 \scriptstyle{\pm 0.11}}}$    &   $\textcolor{blue}{\bm{0.64 \scriptstyle{\pm 0.18}}}$    \\
    hardmean    &   $\textcolor{gray}{0.2 \scriptstyle{\pm 0.17}}$ &   $\textcolor{gray}{1.77 \scriptstyle{\pm 1.89}}$    &   $\bm{0.13 \scriptstyle{\pm 0.0}}$    &   $\bm{0.4 \scriptstyle{\pm 0.0}}$ &   $\bm{0.33 \scriptstyle{\pm 0.21}}$    &   $\bm{0.97 \scriptstyle{\pm 0.31}}$    &   $\textcolor{gray}{0.47 \scriptstyle{\pm 0.13}}$    &   $\textcolor{gray}{2.56 \scriptstyle{\pm 0.72}}$    &   $\bm{-0.0 \scriptstyle{\pm 0.0}}$    &   $\bm{0.21 \scriptstyle{\pm 0.02}}$    &   $\bm{-0.05 \scriptstyle{\pm 0.0}}$   &   $\bm{0.06 \scriptstyle{\pm 0.02}}$    &   $\textcolor{gray}{0.32 \scriptstyle{\pm 0.19}}$    &   $\textcolor{gray}{2.47 \scriptstyle{\pm 2.0}}$    &   $\textcolor{blue}{\bm{0.45 \scriptstyle{\pm 0.08}}}$    &   $\textcolor{blue}{\bm{0.67 \scriptstyle{\pm 0.17}}}$  \\
    harddense   &   $\textcolor{gray}{0.2 \scriptstyle{\pm 0.08}}$ &   $\textcolor{gray}{1.33 \scriptstyle{\pm 0.87}}$    &   $\bm{0.15 \scriptstyle{\pm 0.06}}$    &   $\bm{0.22 \scriptstyle{\pm 0.01}}$    &   $\bm{0.08 \scriptstyle{\pm 0.15}}$    &   $\bm{0.21 \scriptstyle{\pm 0.42}}$    &   $\textcolor{gray}{0.35 \scriptstyle{\pm 0.03}}$    &   $\textcolor{gray}{1.4 \scriptstyle{\pm 0.14}}$ &   $\bm{0.02 \scriptstyle{\pm 0.0}}$    &   $\bm{0.26 \scriptstyle{\pm 0.03}}$    &   $\bm{-0.04 \scriptstyle{\pm 0.01}}$   &   $\bm{0.08 \scriptstyle{\pm 0.01}}$    &   $\textcolor{gray}{0.24 \scriptstyle{\pm 0.21}}$    &   $\textcolor{gray}{1.68 \scriptstyle{\pm 2.15}}$    &   $\textcolor{blue}{\bm{0.35 \scriptstyle{\pm 0.14}}}$    &   $\textcolor{blue}{\bm{0.5 \scriptstyle{\pm 0.21}}}$    \\
    \midrule
    \textbf{MetaDrive Average}  &  $\textcolor{gray}{0.42 \scriptstyle{\pm 0.33}}$    &   $\textcolor{gray}{2.06 \scriptstyle{\pm 1.63}}$    &   $\bm{0.18 \scriptstyle{\pm 0.27}}$    &   $\bm{0.58 \scriptstyle{\pm 0.35}}$    &   $\bm{0.42 \scriptstyle{\pm 0.31}}$    &   $\bm{0.8 \scriptstyle{\pm 0.61}}$ &   $\textcolor{gray}{0.54 \scriptstyle{\pm 0.35}}$    &   $\textcolor{gray}{2.05 \scriptstyle{\pm 2.7}}$    &   $\bm{0.02 \scriptstyle{\pm 0.09}}$    &   $\bm{0.39 \scriptstyle{\pm 0.52}}$    &   $\bm{-0.06 \scriptstyle{\pm 0.01}}$   &   $\bm{0.06 \scriptstyle{\pm 0.04}}$    &   $\textcolor{gray}{0.58 \scriptstyle{\pm 0.32}}$    &   $\textcolor{gray}{2.77 \scriptstyle{\pm 2.87}}$    &   $\textcolor{blue}{\bm{0.53 \scriptstyle{\pm 0.22}}}$    &   $\textcolor{blue}{\bm{0.55 \scriptstyle{\pm 0.24}}}$   \\
    \bottomrule
  \end{tabular}
  }
  \caption{Results of normalized reward and cost. $\uparrow$ means the higher the better. $\downarrow$ means the lower the better. Each value is averaged over 3 distinct cost thresholds, 20 evaluation episodes, and 3 random seeds. \textbf{Bold}: Safe agents whose normalized cost is smaller than 1. \textcolor{gray}{Gray}: Unsafe agents. \textcolor{blue}{\textbf{Blue}}: Safe agents with the highest reward.}
  \label{tab:result}
\end{table*}

\subsection{Additional Ablations}

\subsubsection{Tempurature $\alpha$.} 

\begin{figure}
    \centering
    \begin{subfigure}[t]{0.42\textwidth}
        \centering
        \includegraphics[width=\textwidth]{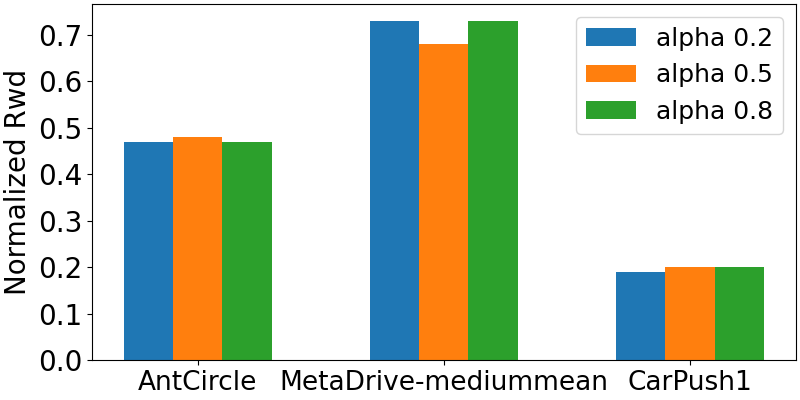}
    \end{subfigure}
    ~
    \begin{subfigure}[t]{0.42\textwidth}
        \centering
        \includegraphics[width=\textwidth]{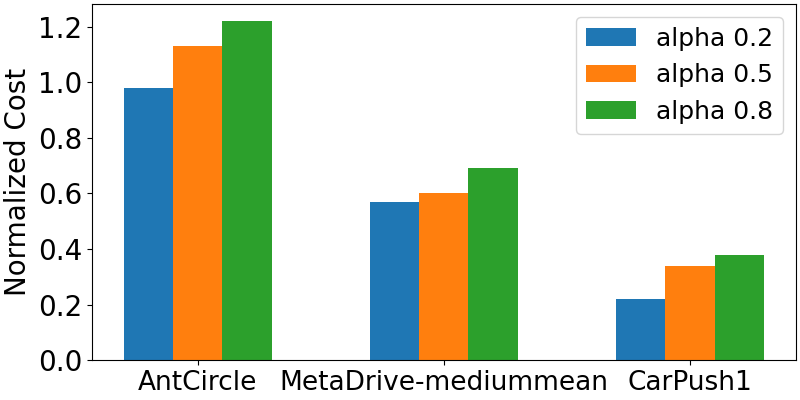}
    \end{subfigure}
    \caption{Ablation study of the temperature $\alpha$ in three tasks. The dotted red horizontal line indicates the cost threshold of 1.}
    \label{fig:alpha}
\end{figure}

We ablate the temperature $\alpha$ in three tasks: AntCircle, MetaDrive-mediummean, and CarPush1, representing each of the three environments. The results are shown in Figure \ref{fig:alpha}. While \texttt{TraC} generally performs well across all values of $\alpha$, higher performance could potentially be achieved with further hyperparameter tuning. Notably, lower values of $\alpha$ are associated with reduced cost in all three tasks.

\subsubsection{$\pi_\text{ref}$ pretraining.}

To further assess the effectiveness of using the reference policy $\pi_\text{ref}$, we conduct additional experiments for all BulletGym tasks with $\pi_\text{ref}$ set as a uniform distribution, meaning the output of $\pi_\text{ref}$ is constant for all state-action pairs. The results are presented in Table \ref{tab:pi_ref}. We observe that rewards remain unaffected by a constant $\pi_\text{ref}$ value, while the cost increases when $\pi_\text{ref}$ is a uniform distribution, particularly for AntCircle, where the cost rises significantly. This increase occurs because a uniform reference policy $\pi_\text{ref}$ lacks information about the underlying offline dataset, leading to distribution drift issues for the learning policy. This highlights the effectiveness of the $\pi_\text{ref}$ pretraining in our approach.

\subsection{Learning Curves}

\begin{figure*}
    \centering
    \begin{subfigure}[t]{0.245\textwidth}
        \centering
        \includegraphics[width=\textwidth]{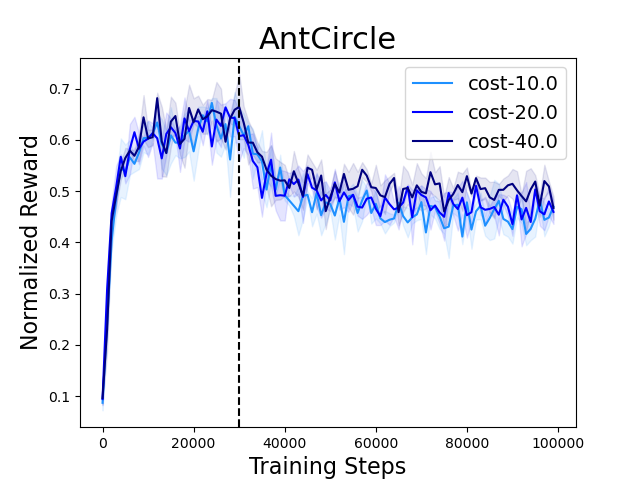}
    \end{subfigure}%
    \begin{subfigure}[t]{0.245\textwidth}
        \centering
        \includegraphics[width=\textwidth]{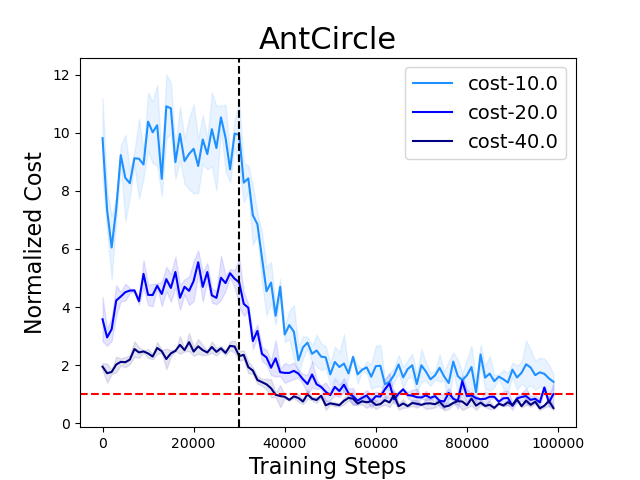}
    \end{subfigure}
    \smallskip
    \begin{subfigure}[t]{0.245\textwidth}
        \centering
        \includegraphics[width=\textwidth]{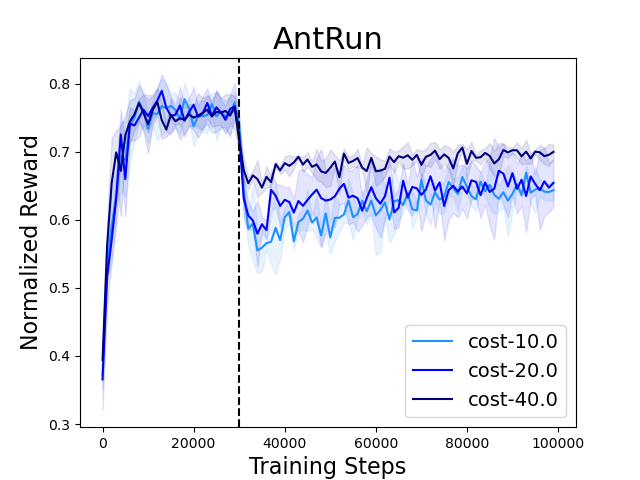}
    \end{subfigure}
    \begin{subfigure}[t]{0.245\textwidth}
        \centering
        \includegraphics[width=\textwidth]{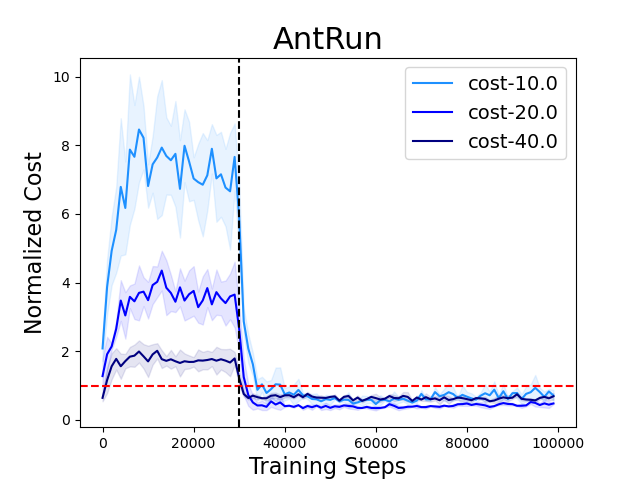}
    \end{subfigure}
    
    \begin{subfigure}[t]{0.245\textwidth}
        \centering
        \includegraphics[width=\textwidth]{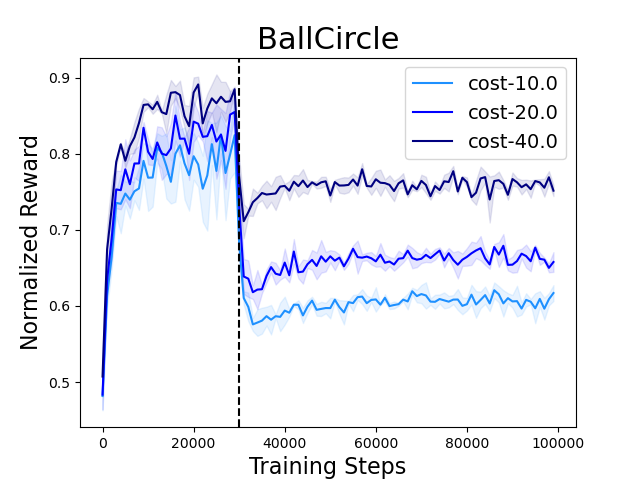}
    \end{subfigure}
    \begin{subfigure}[t]{0.245\textwidth}
        \centering
        \includegraphics[width=\textwidth]{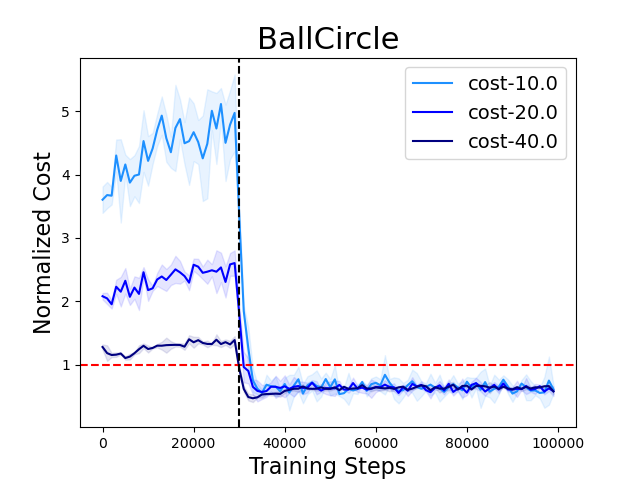}
    \end{subfigure}
    \smallskip
    \begin{subfigure}[t]{0.245\textwidth}
        \centering
        \includegraphics[width=\textwidth]{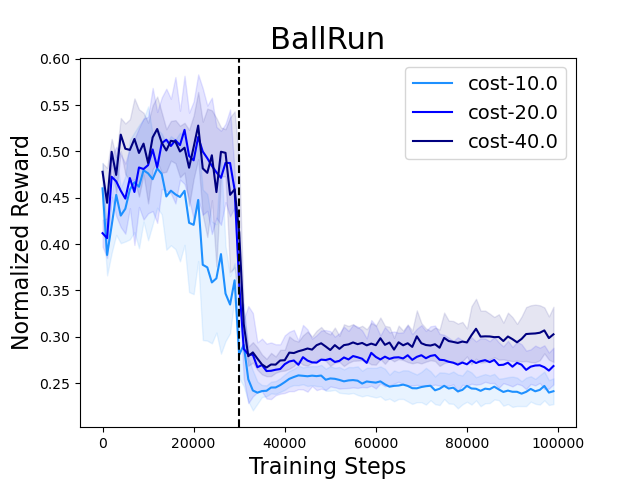}
    \end{subfigure}
    \begin{subfigure}[t]{0.245\textwidth}
        \centering
        \includegraphics[width=\textwidth]{figures/curves/bullet/BallCircle_cost_curve.png}
    \end{subfigure}

    \begin{subfigure}[t]{0.245\textwidth}
        \centering
        \includegraphics[width=\textwidth]{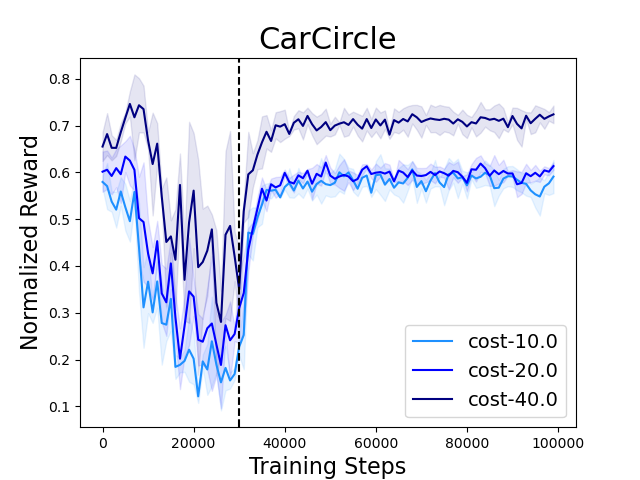}
    \end{subfigure}%
    \begin{subfigure}[t]{0.245\textwidth}
        \centering
        \includegraphics[width=\textwidth]{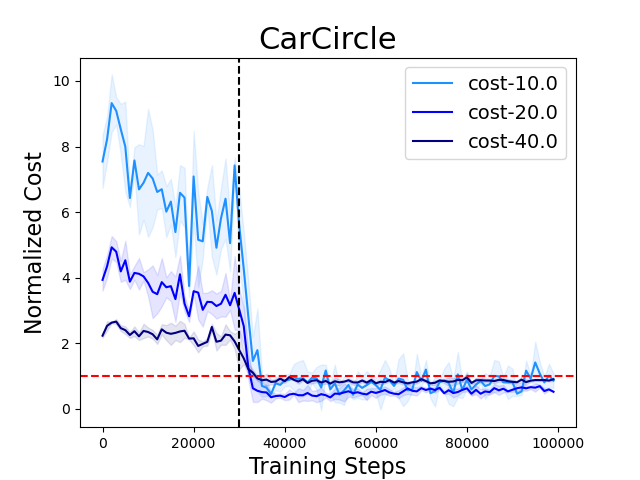}
    \end{subfigure}
    \smallskip
    \begin{subfigure}[t]{0.245\textwidth}
        \centering
        \includegraphics[width=\textwidth]{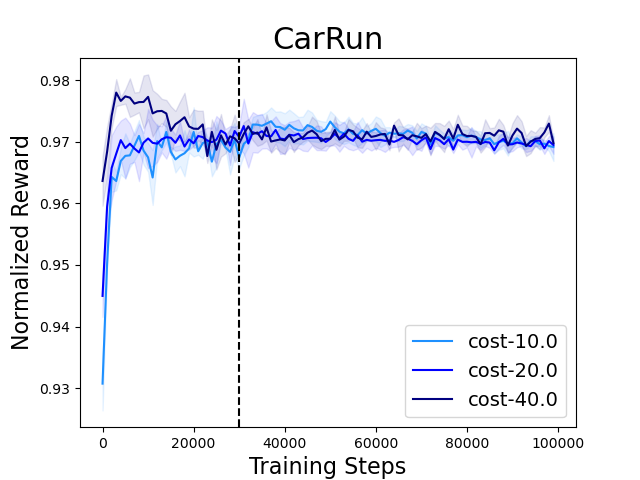}
    \end{subfigure}
    \begin{subfigure}[t]{0.245\textwidth}
        \centering
        \includegraphics[width=\textwidth]{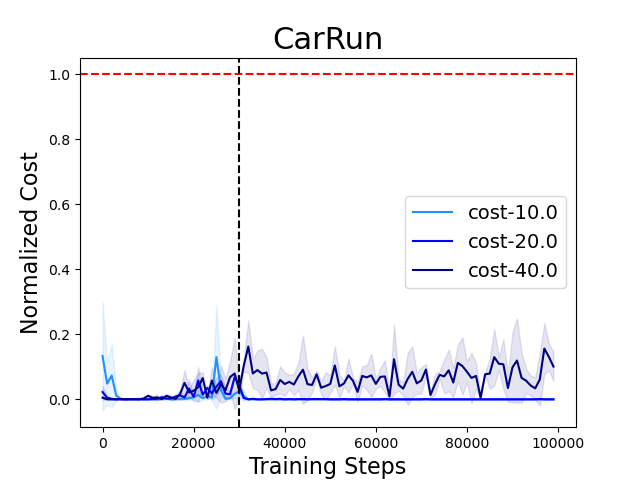}
    \end{subfigure}
    
    \begin{subfigure}[t]{0.245\textwidth}
        \centering
        \includegraphics[width=\textwidth]{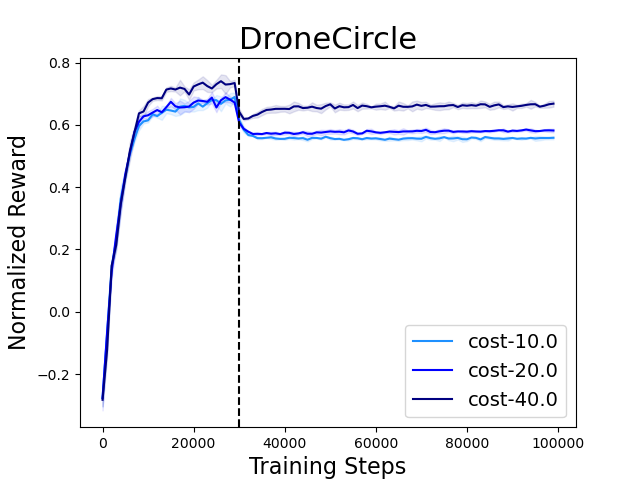}
    \end{subfigure}
    \begin{subfigure}[t]{0.245\textwidth}
        \centering
        \includegraphics[width=\textwidth]{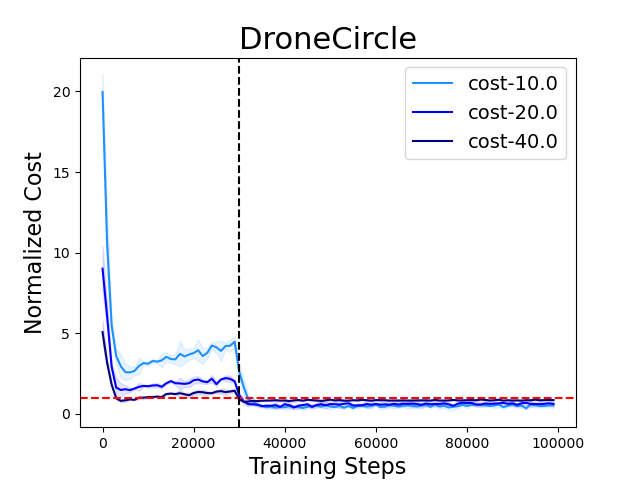}
    \end{subfigure}
    \smallskip
    \begin{subfigure}[t]{0.245\textwidth}
        \centering
        \includegraphics[width=\textwidth]{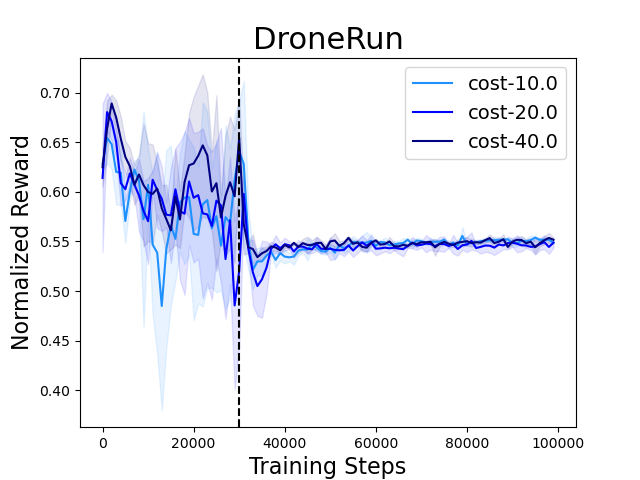}
    \end{subfigure}
    \begin{subfigure}[t]{0.245\textwidth}
        \centering
        \includegraphics[width=\textwidth]{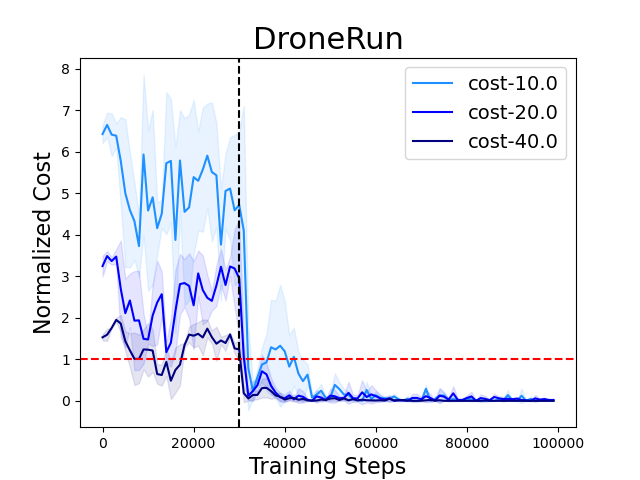}
    \end{subfigure}
    \caption{Training curves for the 8 tasks in BulletGym.}
    \label{fig:bulletgym_curve}
\end{figure*}

\begin{figure*}
    \centering
    \begin{subfigure}[t]{0.245\textwidth}
        \centering
        \includegraphics[width=\textwidth]{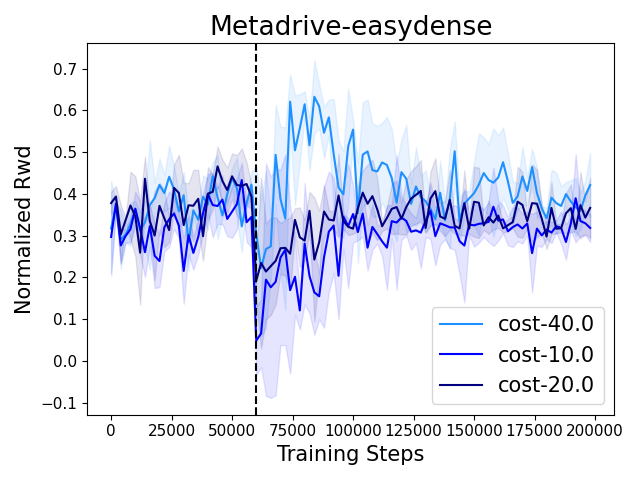}
    \end{subfigure}%
    \begin{subfigure}[t]{0.245\textwidth}
        \centering
        \includegraphics[width=\textwidth]{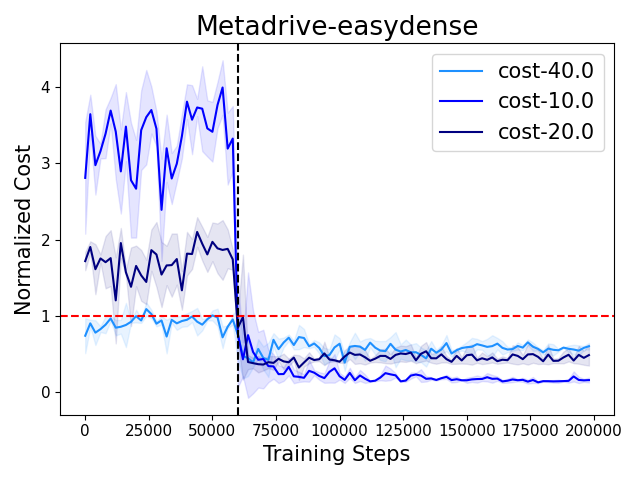}
    \end{subfigure}
    \smallskip
    \begin{subfigure}[t]{0.245\textwidth}
        \centering
        \includegraphics[width=\textwidth]{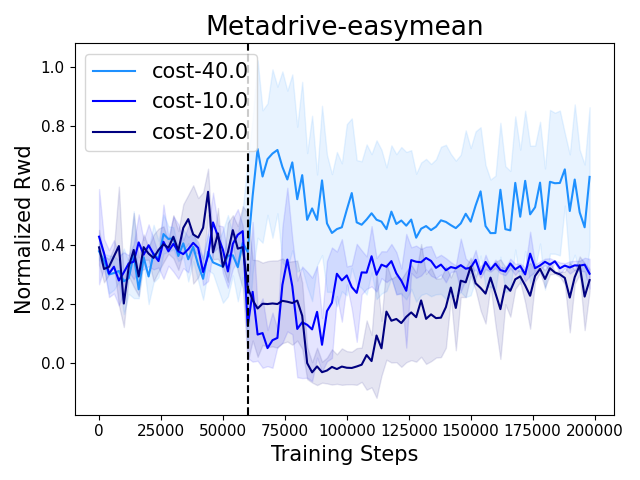}
    \end{subfigure}
    \begin{subfigure}[t]{0.245\textwidth}
        \centering
        \includegraphics[width=\textwidth]{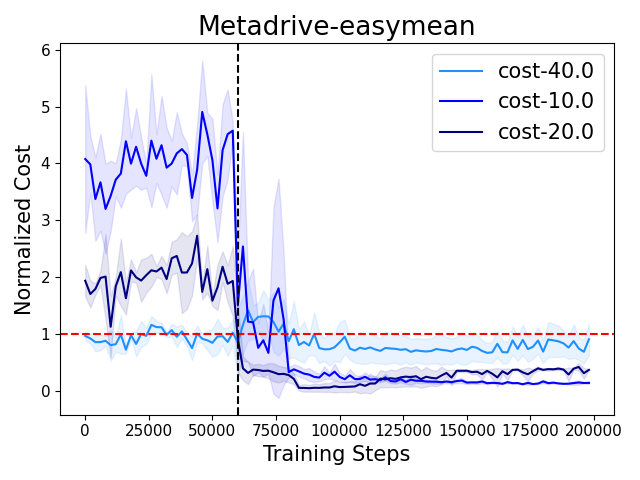}
    \end{subfigure}
    
    \begin{subfigure}[t]{0.245\textwidth}
        \centering
        \includegraphics[width=\textwidth]{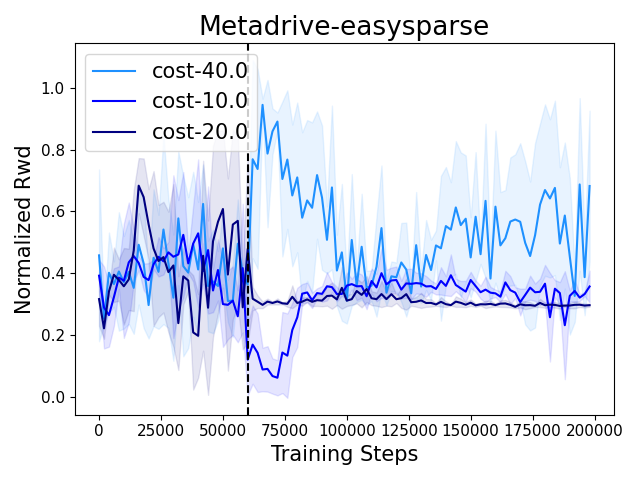}
    \end{subfigure}
    \begin{subfigure}[t]{0.245\textwidth}
        \centering
        \includegraphics[width=\textwidth]{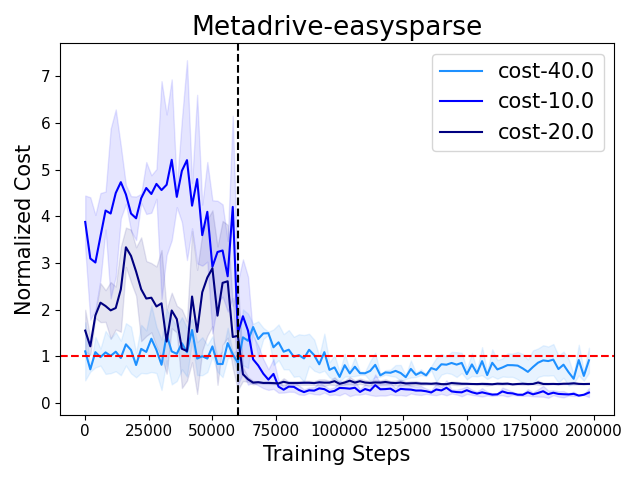}
    \end{subfigure}
    \smallskip
    \begin{subfigure}[t]{0.245\textwidth}
        \centering
        \includegraphics[width=\textwidth]{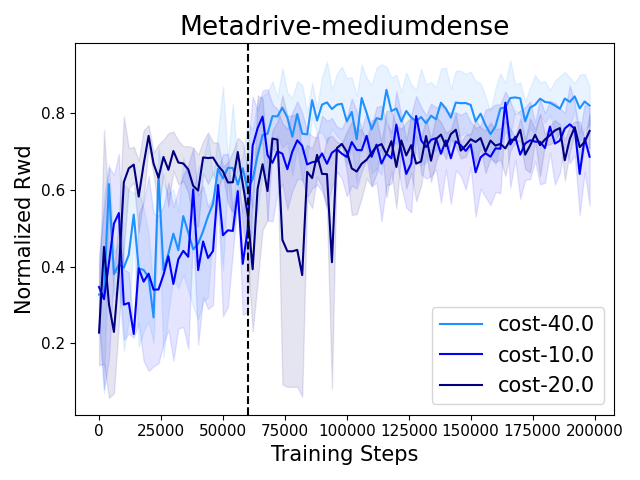}
    \end{subfigure}
    \begin{subfigure}[t]{0.245\textwidth}
        \centering
        \includegraphics[width=\textwidth]{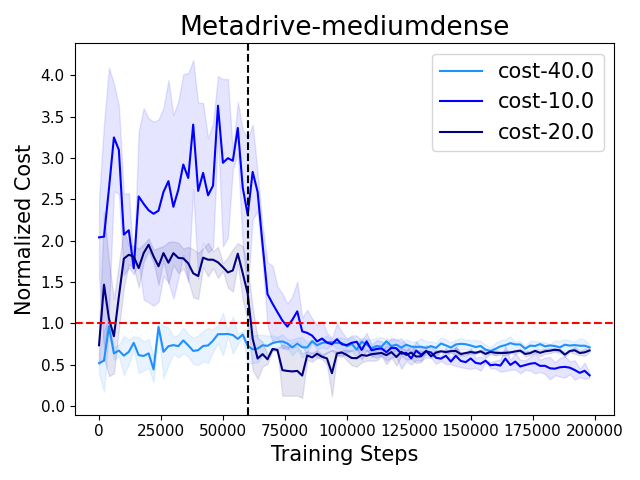}
    \end{subfigure}

    \begin{subfigure}[t]{0.245\textwidth}
        \centering
        \includegraphics[width=\textwidth]{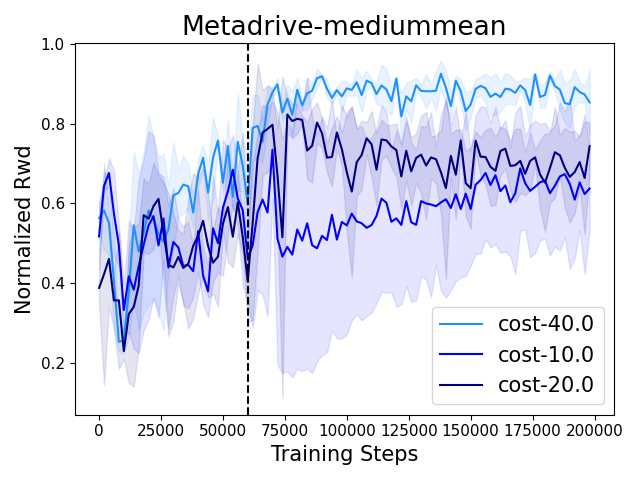}
    \end{subfigure}%
    \begin{subfigure}[t]{0.245\textwidth}
        \centering
        \includegraphics[width=\textwidth]{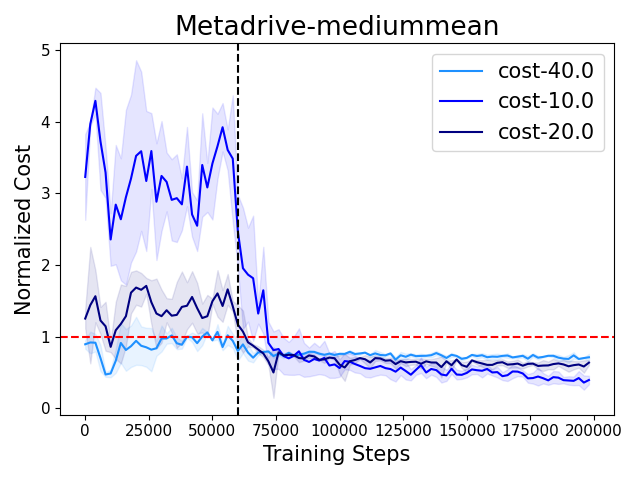}
    \end{subfigure}
    \smallskip
    \begin{subfigure}[t]{0.245\textwidth}
        \centering
        \includegraphics[width=\textwidth]{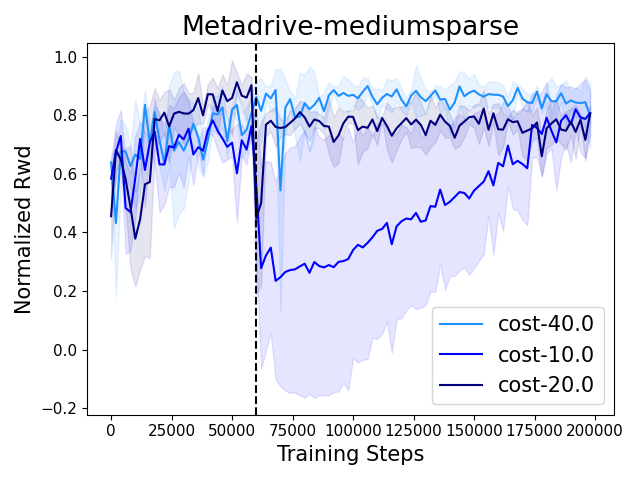}
    \end{subfigure}
    \begin{subfigure}[t]{0.245\textwidth}
        \centering
        \includegraphics[width=\textwidth]{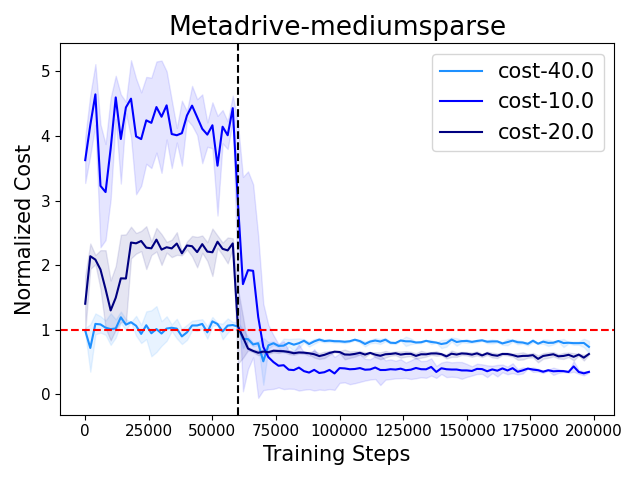}
    \end{subfigure}
    
    \begin{subfigure}[t]{0.245\textwidth}
        \centering
        \includegraphics[width=\textwidth]{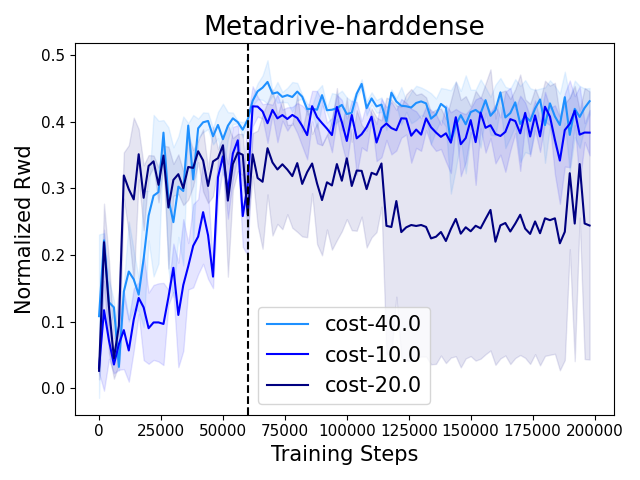}
    \end{subfigure}
    \begin{subfigure}[t]{0.245\textwidth}
        \centering
        \includegraphics[width=\textwidth]{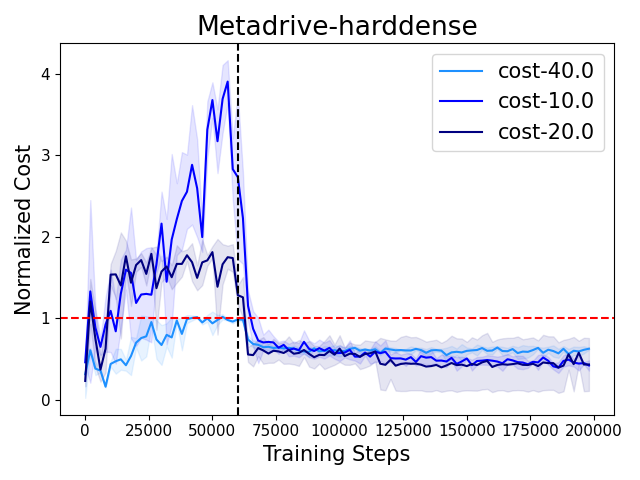}
    \end{subfigure}
    \smallskip
    \begin{subfigure}[t]{0.245\textwidth}
        \centering
        \includegraphics[width=\textwidth]{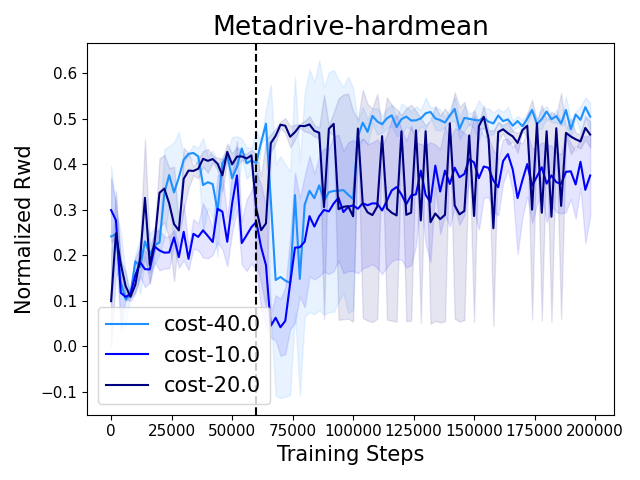}
    \end{subfigure}
    \begin{subfigure}[t]{0.245\textwidth}
        \centering
        \includegraphics[width=\textwidth]{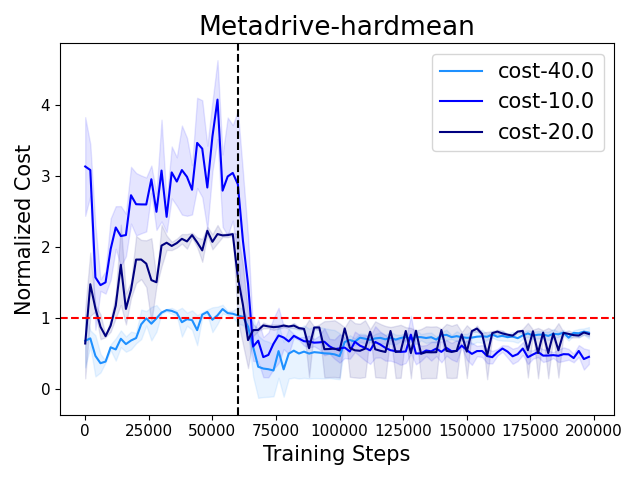}
    \end{subfigure}

    \begin{subfigure}[t]{0.245\textwidth}
        \centering
        \includegraphics[width=\textwidth]{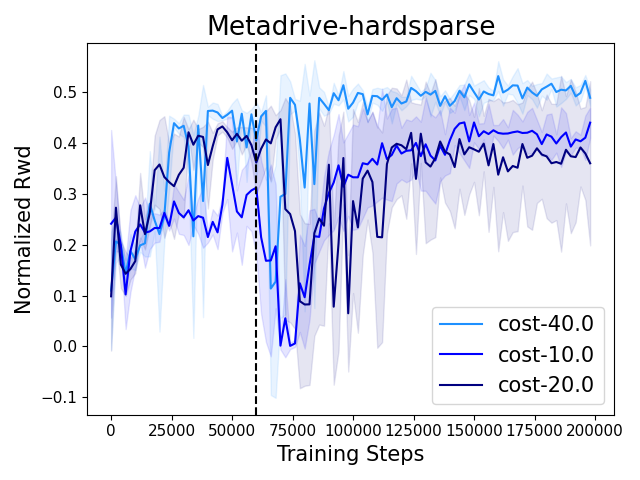}
    \end{subfigure}
    \begin{subfigure}[t]{0.245\textwidth}
        \centering
        \includegraphics[width=\textwidth]{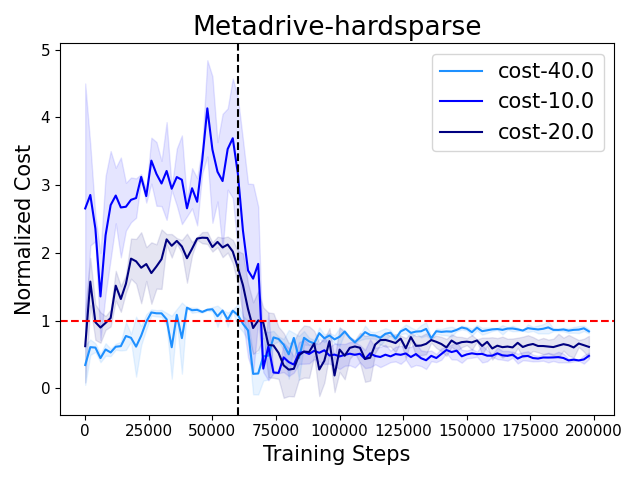}
    \end{subfigure}
    \caption{Learning curves for the 9 tasks in MetaDrive.}
    \label{fig:meta_curve}
\end{figure*}

\begin{figure*}
    \centering
    \begin{subfigure}[t]{0.245\textwidth}
        \centering
        \includegraphics[width=\textwidth]{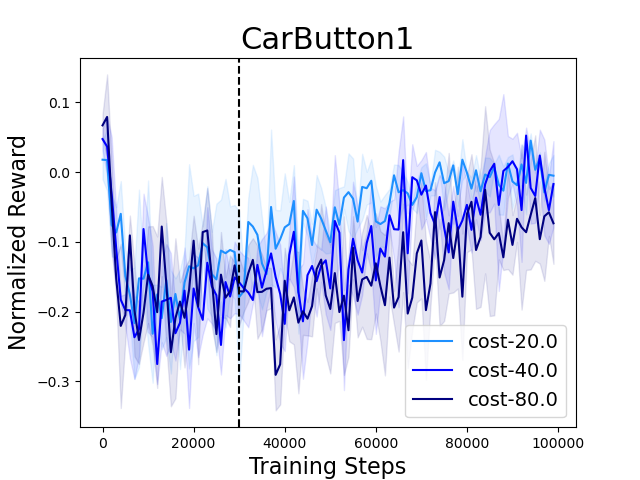}
    \end{subfigure}%
    \begin{subfigure}[t]{0.245\textwidth}
        \centering
        \includegraphics[width=\textwidth]{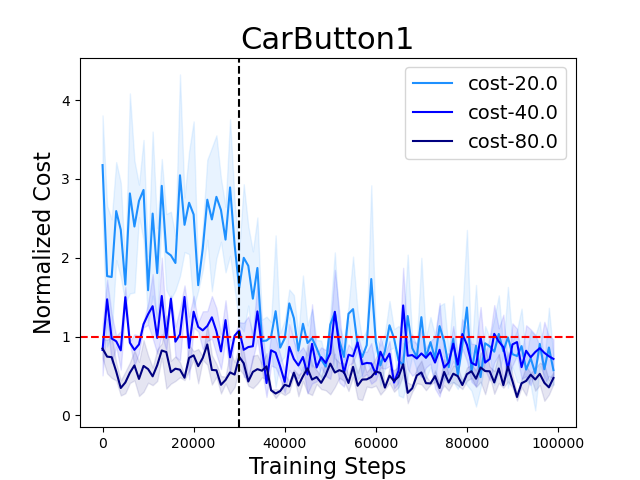}
    \end{subfigure}
    \smallskip
    \begin{subfigure}[t]{0.245\textwidth}
        \centering
        \includegraphics[width=\textwidth]{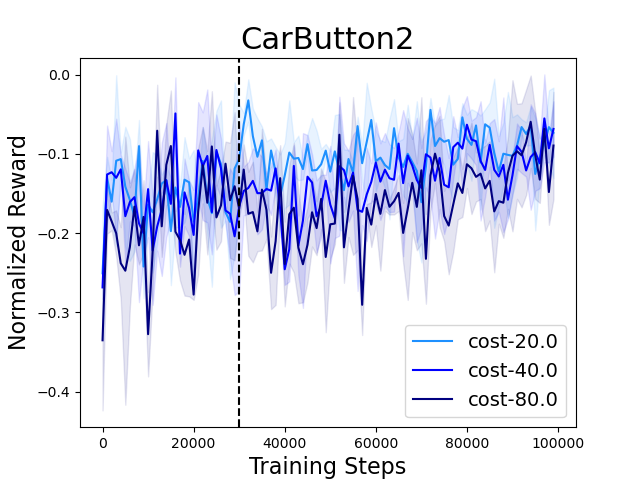}
    \end{subfigure}
    \begin{subfigure}[t]{0.245\textwidth}
        \centering
        \includegraphics[width=\textwidth]{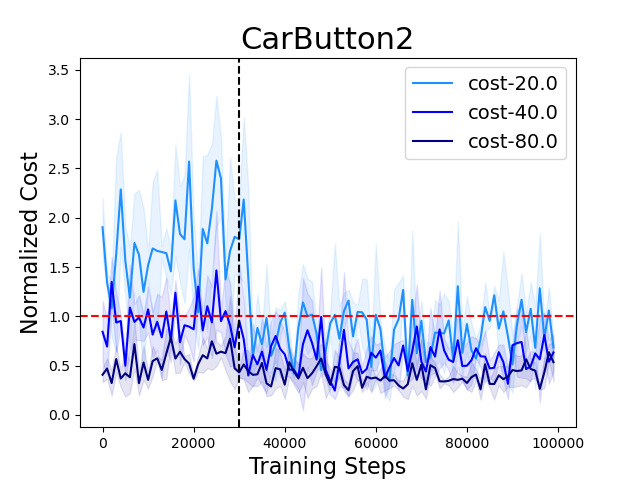}
    \end{subfigure}
    
    \begin{subfigure}[t]{0.245\textwidth}
        \centering
        \includegraphics[width=\textwidth]{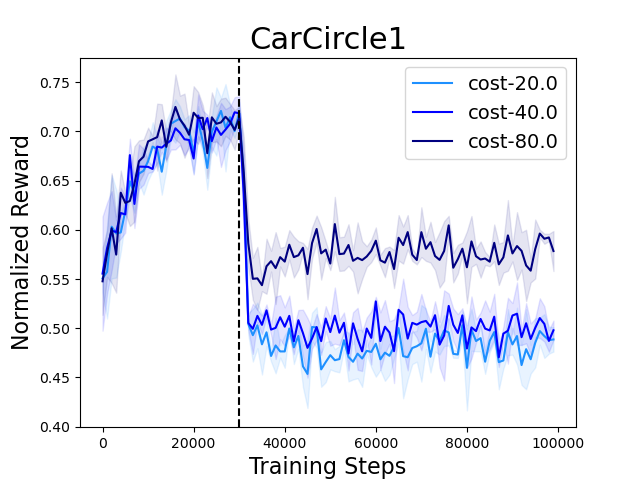}
    \end{subfigure}
    \begin{subfigure}[t]{0.245\textwidth}
        \centering
        \includegraphics[width=\textwidth]{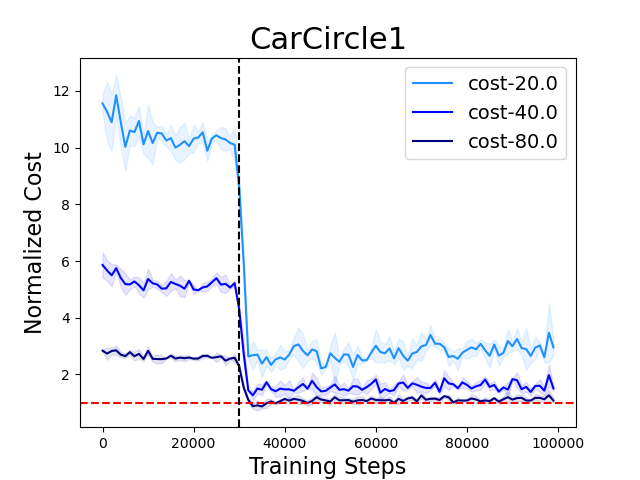}
    \end{subfigure}
    \smallskip
    \begin{subfigure}[t]{0.245\textwidth}
        \centering
        \includegraphics[width=\textwidth]{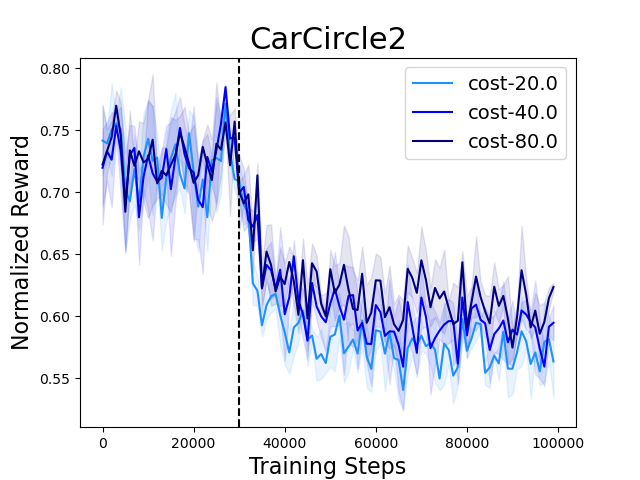}
    \end{subfigure}
    \begin{subfigure}[t]{0.245\textwidth}
        \centering
        \includegraphics[width=\textwidth]{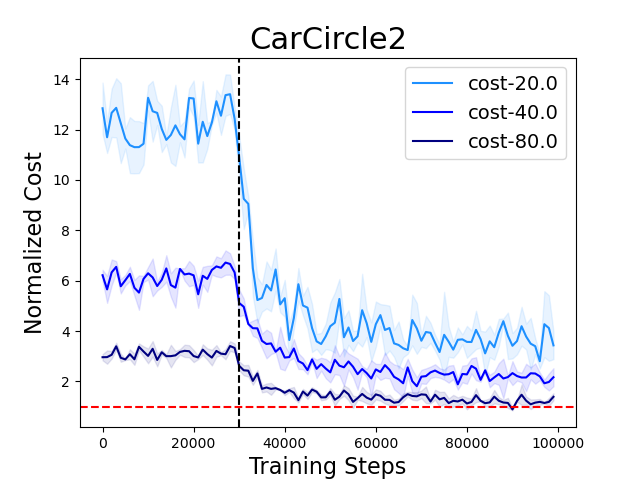}
    \end{subfigure}

    \begin{subfigure}[t]{0.245\textwidth}
        \centering
        \includegraphics[width=\textwidth]{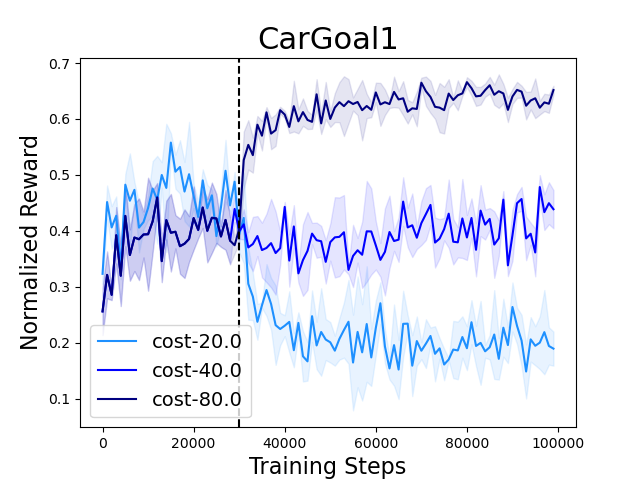}
    \end{subfigure}%
    \begin{subfigure}[t]{0.245\textwidth}
        \centering
        \includegraphics[width=\textwidth]{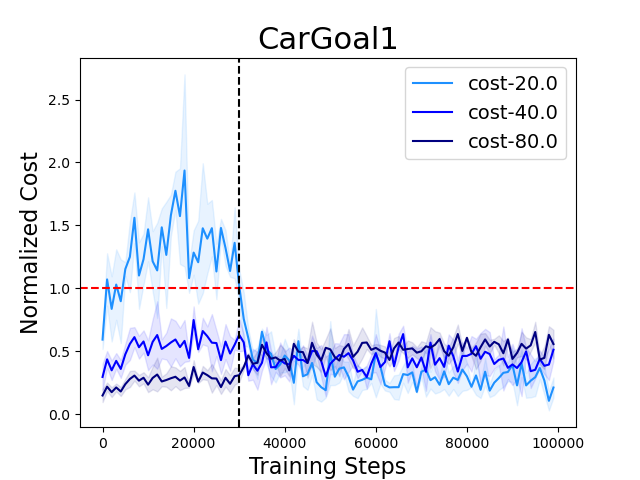}
    \end{subfigure}
    \smallskip
    \begin{subfigure}[t]{0.245\textwidth}
        \centering
        \includegraphics[width=\textwidth]{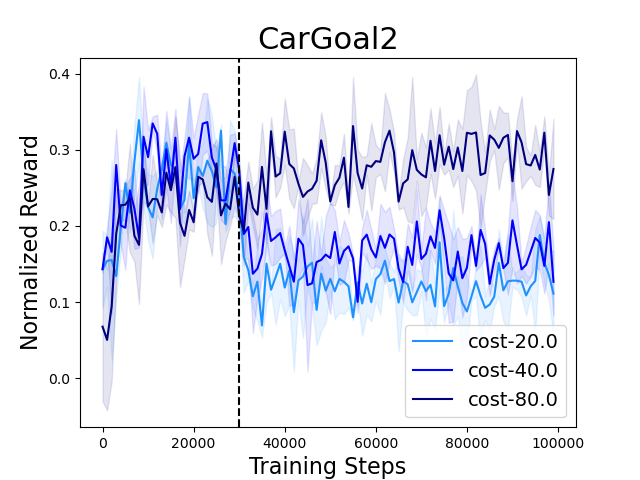}
    \end{subfigure}
    \begin{subfigure}[t]{0.245\textwidth}
        \centering
        \includegraphics[width=\textwidth]{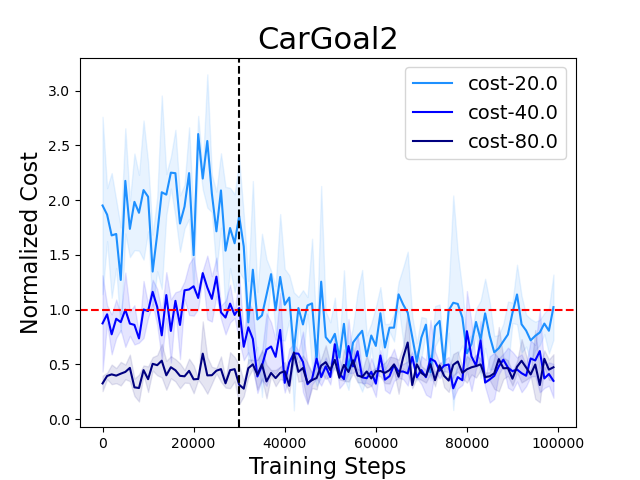}
    \end{subfigure}
    
    \begin{subfigure}[t]{0.245\textwidth}
        \centering
        \includegraphics[width=\textwidth]{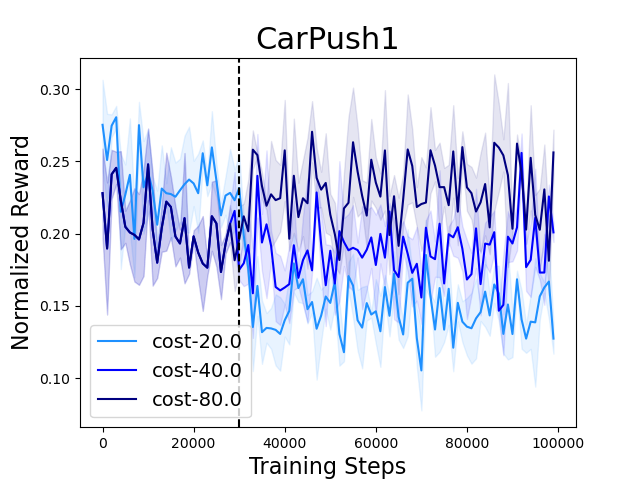}
    \end{subfigure}
    \begin{subfigure}[t]{0.245\textwidth}
        \centering
        \includegraphics[width=\textwidth]{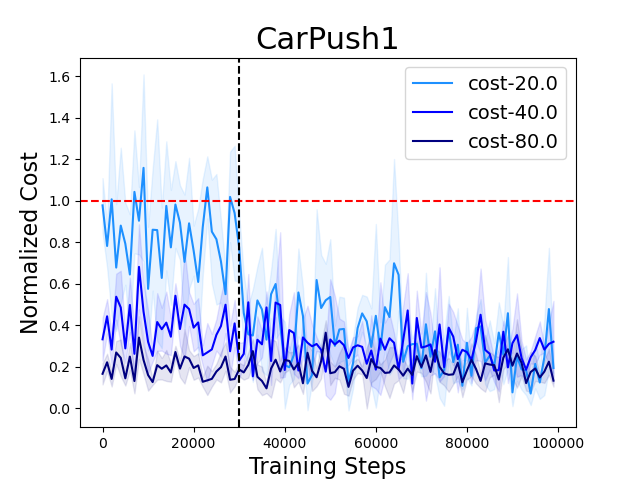}
    \end{subfigure}
    \smallskip
    \begin{subfigure}[t]{0.245\textwidth}
        \centering
        \includegraphics[width=\textwidth]{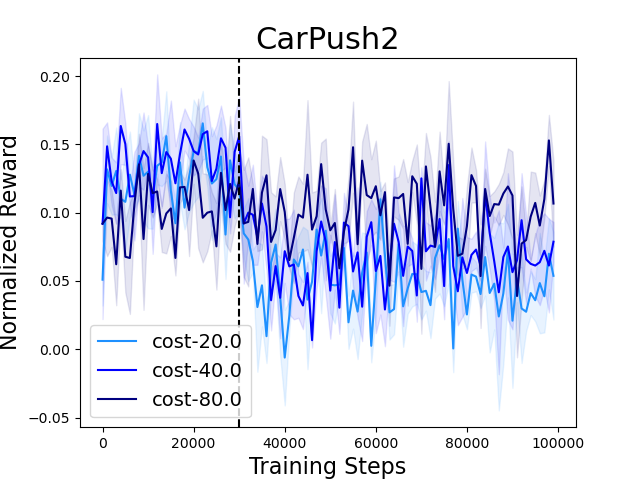}
    \end{subfigure}
    \begin{subfigure}[t]{0.245\textwidth}
        \centering
        \includegraphics[width=\textwidth]{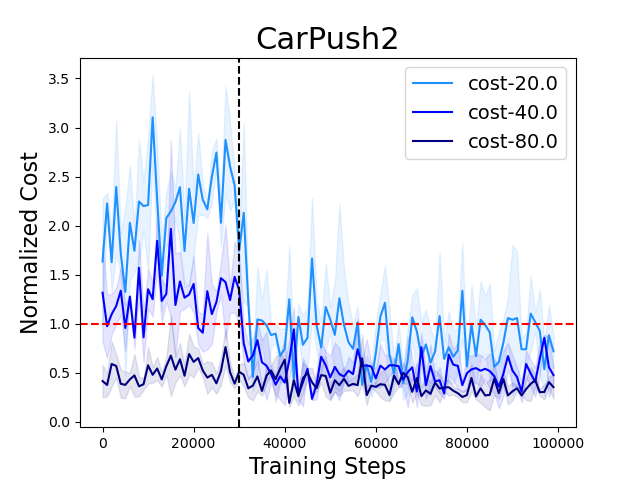}
    \end{subfigure}
    \caption{Learning curves for the 8 Car tasks in SafetyGym.}
    \label{fig:safetgym_curve1}
\end{figure*}

\begin{figure*}
    \centering
    \begin{subfigure}[t]{0.245\textwidth}
        \centering
        \includegraphics[width=\textwidth]{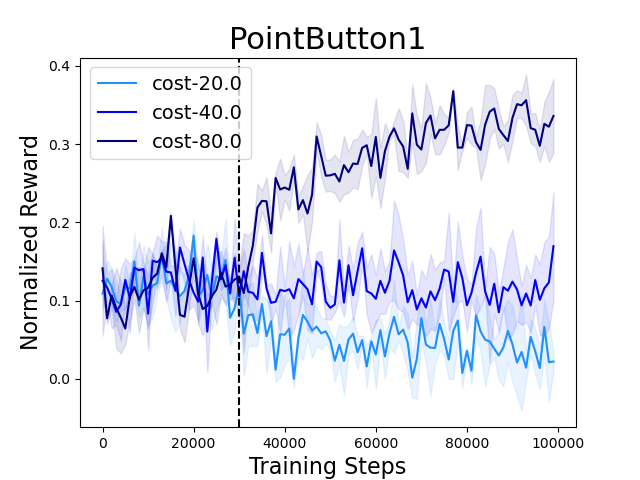}
    \end{subfigure}%
    \begin{subfigure}[t]{0.245\textwidth}
        \centering
        \includegraphics[width=\textwidth]{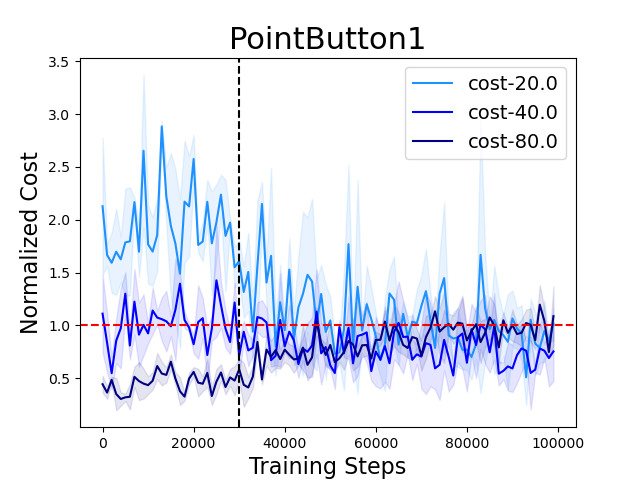}
    \end{subfigure}
    \smallskip
    \begin{subfigure}[t]{0.245\textwidth}
        \centering
        \includegraphics[width=\textwidth]{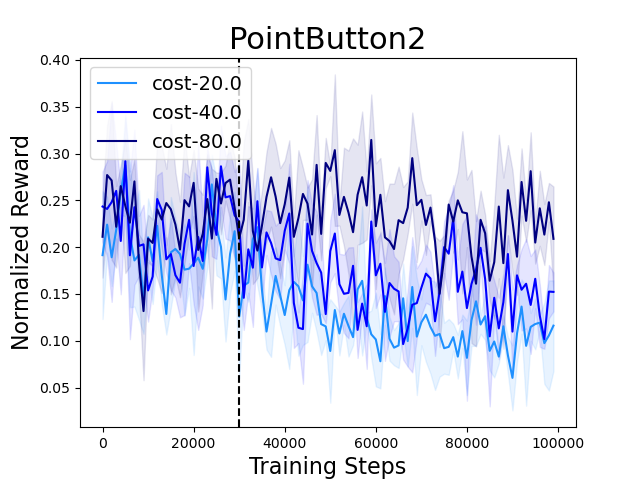}
    \end{subfigure}
    \begin{subfigure}[t]{0.245\textwidth}
        \centering
        \includegraphics[width=\textwidth]{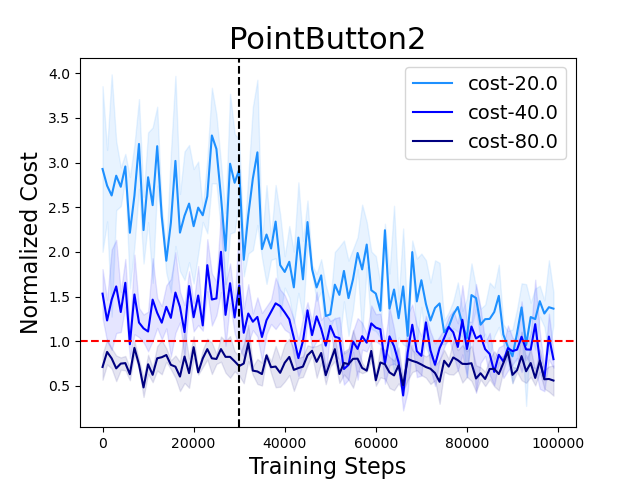}
    \end{subfigure}
    
    \begin{subfigure}[t]{0.245\textwidth}
        \centering
        \includegraphics[width=\textwidth]{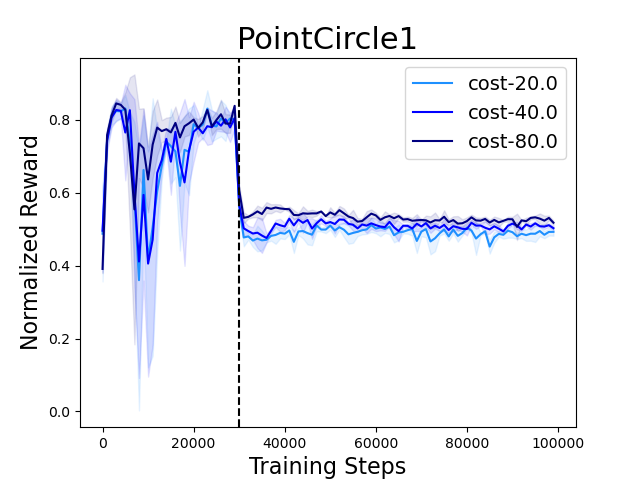}
    \end{subfigure}
    \begin{subfigure}[t]{0.245\textwidth}
        \centering
        \includegraphics[width=\textwidth]{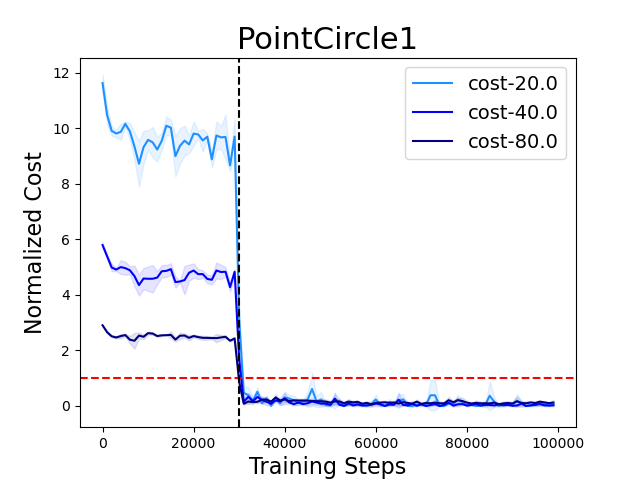}
    \end{subfigure}
    \smallskip
    \begin{subfigure}[t]{0.245\textwidth}
        \centering
        \includegraphics[width=\textwidth]{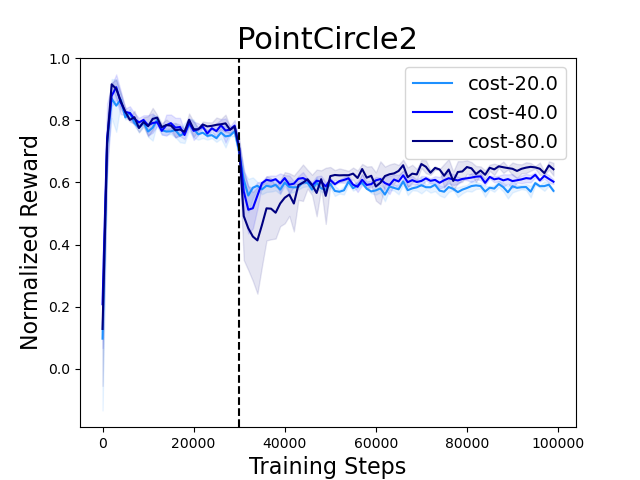}
    \end{subfigure}
    \begin{subfigure}[t]{0.245\textwidth}
        \centering
        \includegraphics[width=\textwidth]{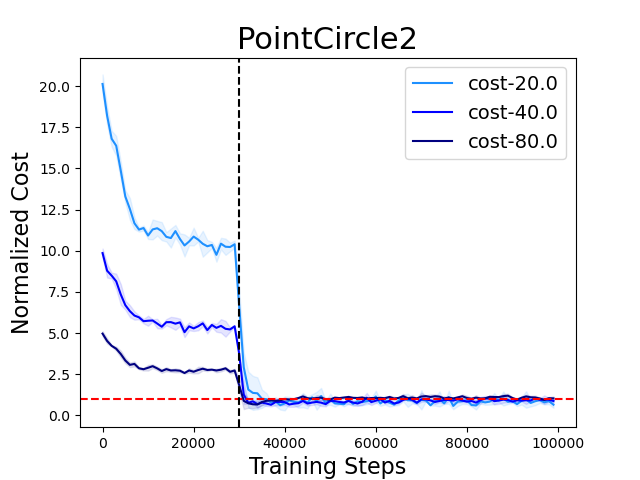}
    \end{subfigure}

    \begin{subfigure}[t]{0.245\textwidth}
        \centering
        \includegraphics[width=\textwidth]{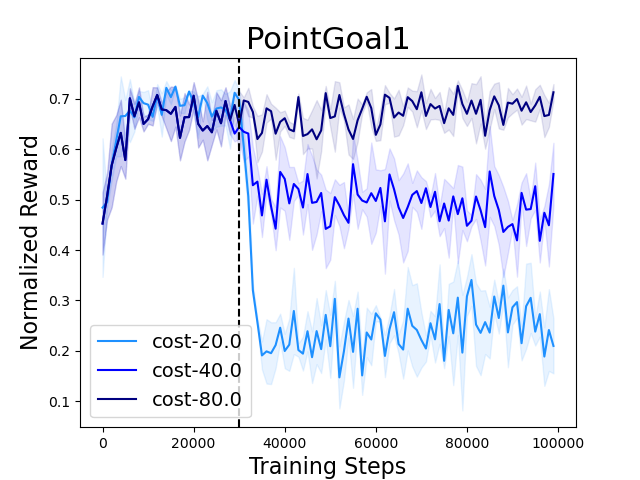}
    \end{subfigure}%
    \begin{subfigure}[t]{0.245\textwidth}
        \centering
        \includegraphics[width=\textwidth]{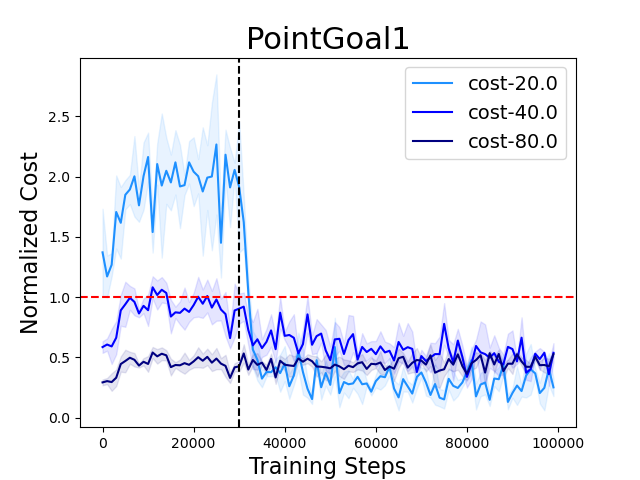}
    \end{subfigure}
    \smallskip
    \begin{subfigure}[t]{0.245\textwidth}
        \centering
        \includegraphics[width=\textwidth]{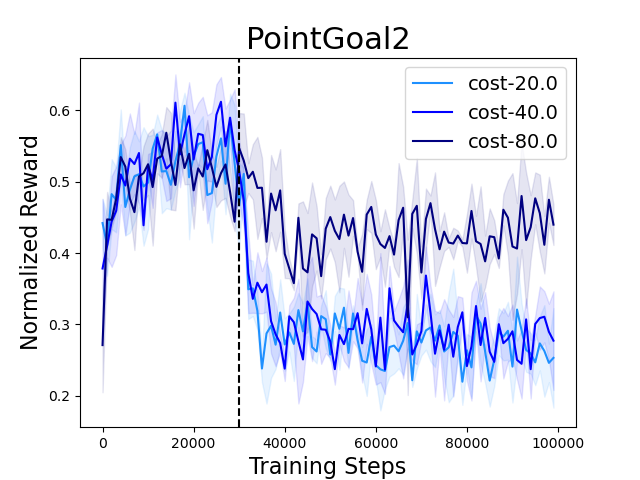}
    \end{subfigure}
    \begin{subfigure}[t]{0.245\textwidth}
        \centering
        \includegraphics[width=\textwidth]{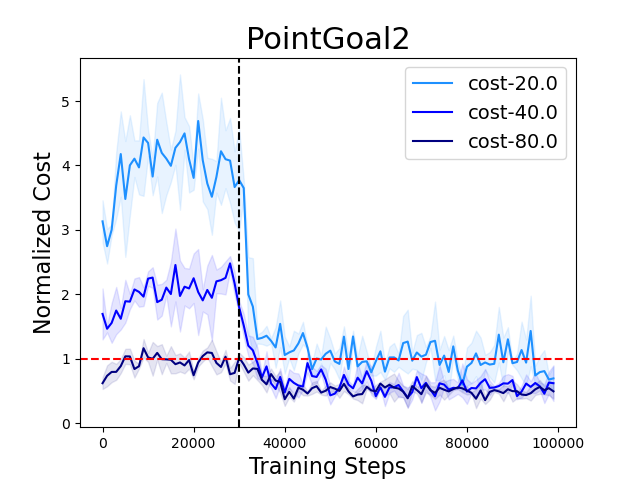}
    \end{subfigure}
    
    \begin{subfigure}[t]{0.245\textwidth}
        \centering
        \includegraphics[width=\textwidth]{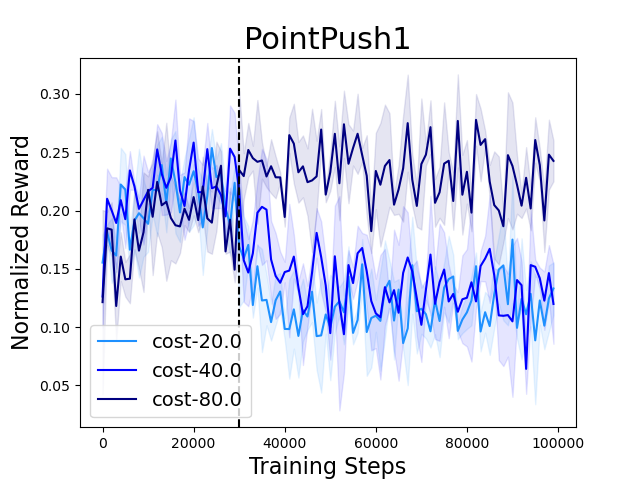}
    \end{subfigure}
    \begin{subfigure}[t]{0.245\textwidth}
        \centering
        \includegraphics[width=\textwidth]{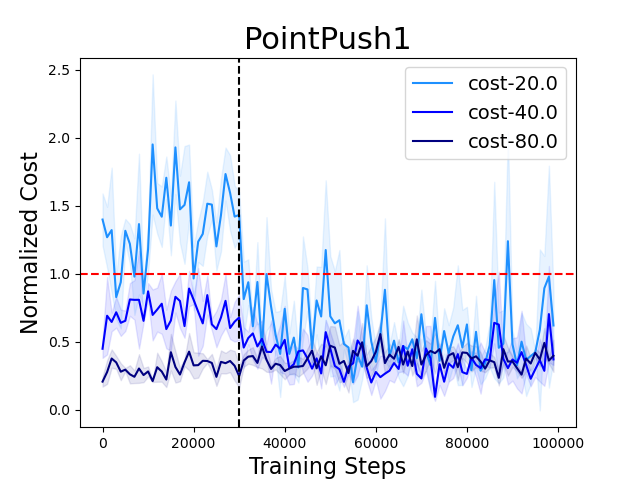}
    \end{subfigure}
    \smallskip
    \begin{subfigure}[t]{0.245\textwidth}
        \centering
        \includegraphics[width=\textwidth]{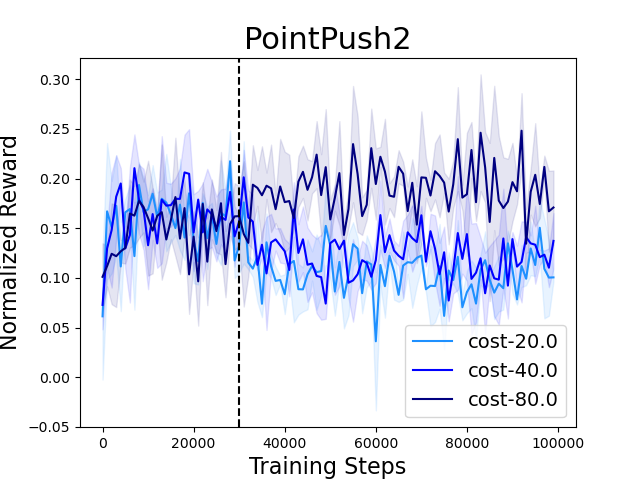}
    \end{subfigure}
    \begin{subfigure}[t]{0.245\textwidth}
        \centering
        \includegraphics[width=\textwidth]{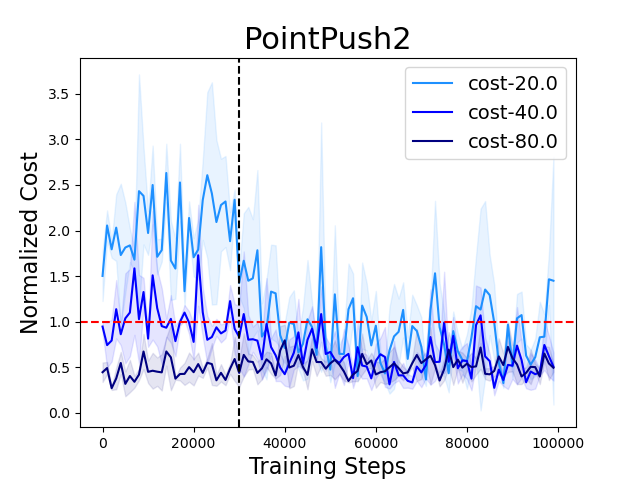}
    \end{subfigure}
    \caption{Learning curves for the 8 Point tasks in SafetyGym.}
    \label{fig:safetgym_curve2}
\end{figure*}

\begin{figure*}
    \centering
    \begin{subfigure}[t]{0.245\textwidth}
        \centering
        \includegraphics[width=\textwidth]{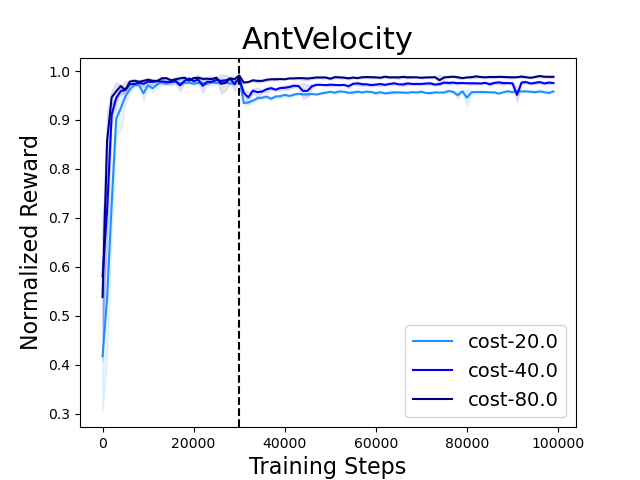}
    \end{subfigure}%
    \begin{subfigure}[t]{0.245\textwidth}
        \centering
        \includegraphics[width=\textwidth]{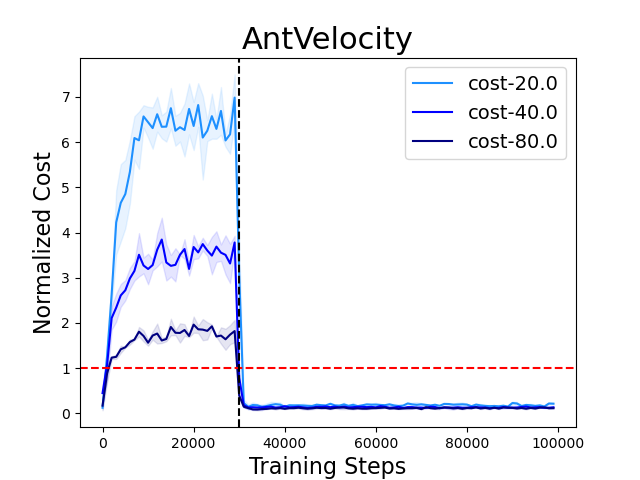}
    \end{subfigure}
    \smallskip
    \begin{subfigure}[t]{0.245\textwidth}
        \centering
        \includegraphics[width=\textwidth]{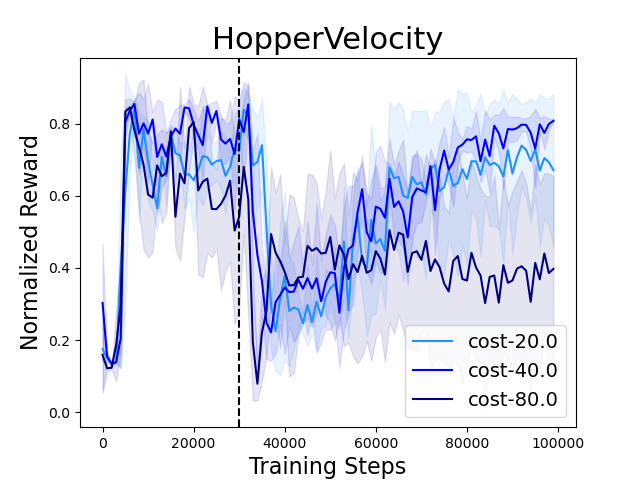}
    \end{subfigure}
    \begin{subfigure}[t]{0.245\textwidth}
        \centering
        \includegraphics[width=\textwidth]{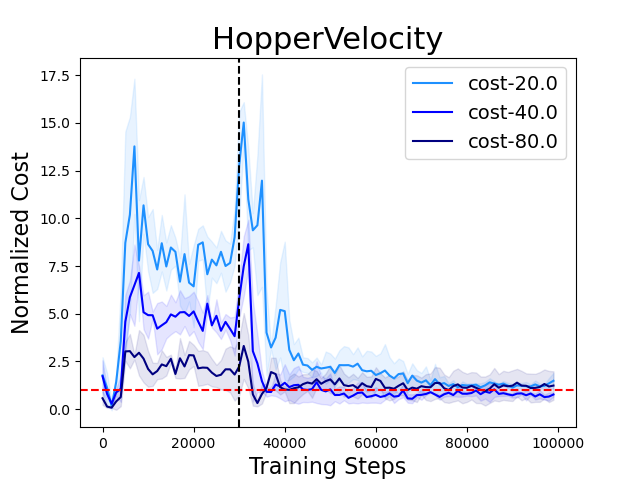}
    \end{subfigure}
    
    \begin{subfigure}[t]{0.245\textwidth}
        \centering
        \includegraphics[width=\textwidth]{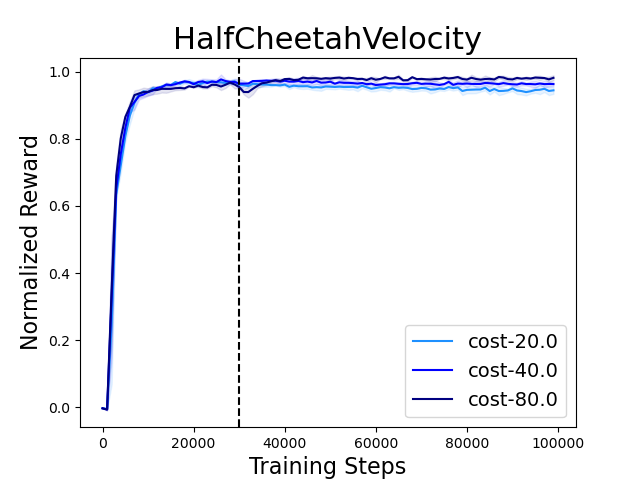}
    \end{subfigure}
    \begin{subfigure}[t]{0.245\textwidth}
        \centering
        \includegraphics[width=\textwidth]{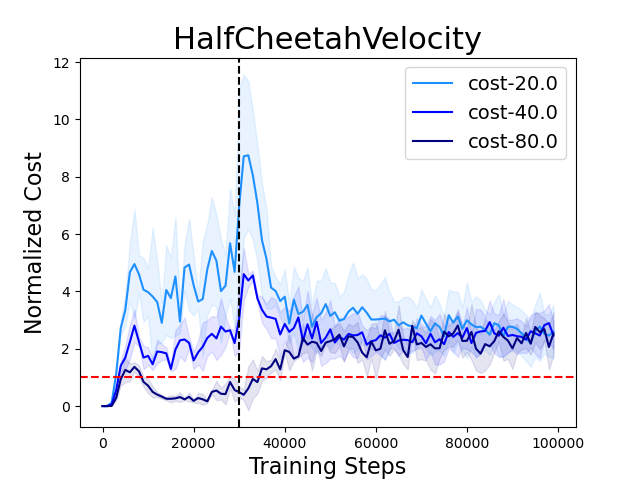}
    \end{subfigure}
    \smallskip
    \begin{subfigure}[t]{0.245\textwidth}
        \centering
        \includegraphics[width=\textwidth]{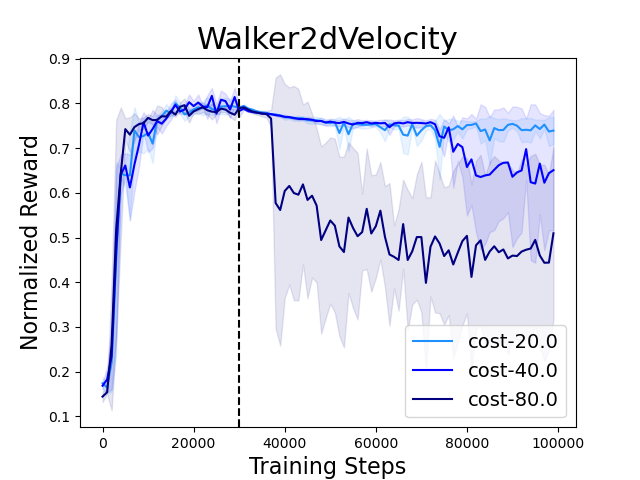}
    \end{subfigure}
    \begin{subfigure}[t]{0.245\textwidth}
        \centering
        \includegraphics[width=\textwidth]{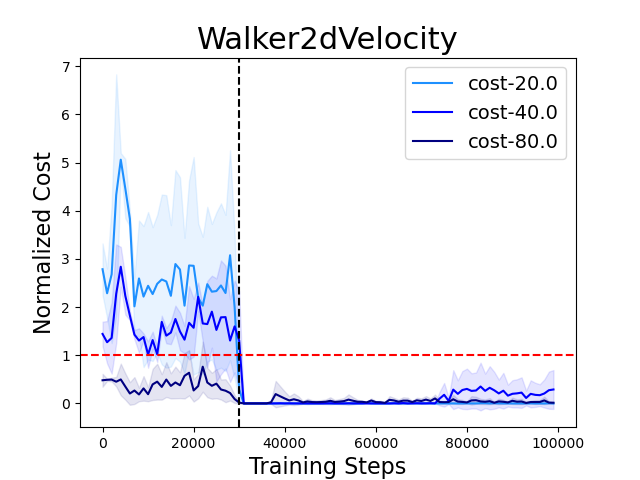}
    \end{subfigure}

    \begin{subfigure}[t]{0.245\textwidth}
        \centering
        \includegraphics[width=\textwidth]{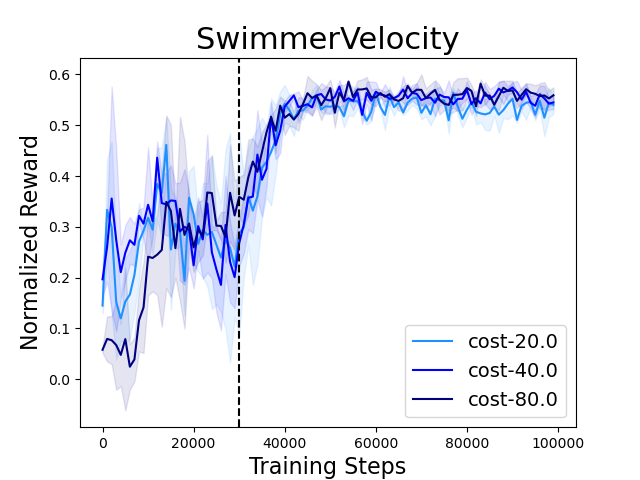}
    \end{subfigure}%
    \begin{subfigure}[t]{0.245\textwidth}
        \centering
        \includegraphics[width=\textwidth]{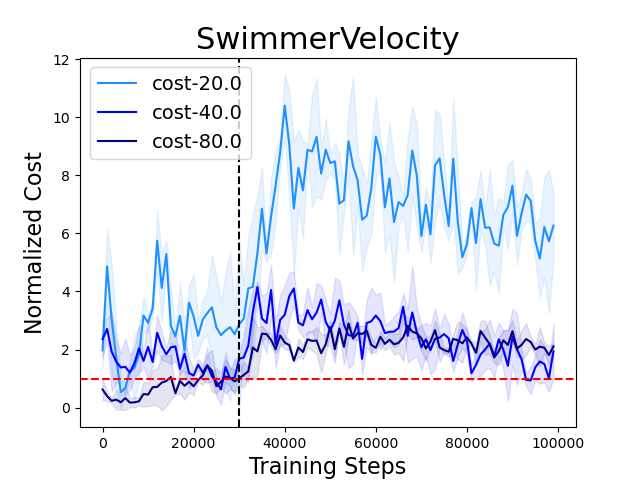}
    \end{subfigure}
    \caption{Learning curves for the 5 velocity constraint tasks in SafetyGym.}
    \label{fig:safetgym_curve3}
\end{figure*}

We train \texttt{TraC} with the parameters in Table \ref{tab:hyperparameters} for all 38 tasks from the DSRL benchmark \cite{liu2023datasets}. The learning curves are shown in Figure \ref{fig:bulletgym_curve}, \ref{fig:meta_curve}, \ref{fig:safetgym_curve1}, \ref{fig:safetgym_curve2}, and \ref{fig:safetgym_curve3}. In each figure, the dotted vertical line marks the point where $\pi_\text{ref}$ pretraining stops, while the dotted horizontal line indicates the cost threshold of 1.

\begin{table}
  \centering
  \begin{tabular}{ccccc}
    \toprule
    \multirow{2}[2]{*}{Task} & \multicolumn{2}{c}{$\pi_\text{ref}$ uniform} & \multicolumn{2}{c}{\texttt{TraC}} \\
    \cmidrule(lr){2-3}
    \cmidrule(lr){4-5}
      &   reward$\uparrow$    &   cost$\downarrow$ &   reward$\uparrow$    &   cost$\downarrow$     \\
    \midrule
    AntCircle   &   \cellcolor{blue!10}0.55    &   \cellcolor{red!10}4.73    &   0.47    &   \cellcolor{blue!10}0.98    \\
    AntRun   &   0.66    &   \cellcolor{blue!10}0.59    &   \cellcolor{blue!10}0.67    &   0.63    \\
    BallCircle  &  0.67    &   0.66    &   \cellcolor{blue!10}0.68    &   \cellcolor{blue!10}0.59      \\
    BallRun   &   0.27    &   0.49    &   0.27    &   \cellcolor{blue!10}0.47    \\
    CarCircle   &   0.61    &   0.81    &   \cellcolor{blue!10}0.64    &   \cellcolor{blue!10}0.76    \\
    CarRun  &  0.97    &   \cellcolor{blue!10}0.00    &   0.97    &   0.03      \\
    DroneCircle   &   0.6    &   \cellcolor{blue!10}0.61    &   0.6    &   0.67    \\
    DroneRun  &  0.55    &   0.01    &   0.55    &   0.01      \\
    \midrule
    Average  &  0.61    &   0.99    &   0.61    &   \cellcolor{blue!10}0.52      \\
    \bottomrule
  \end{tabular}
  \caption{Ablation study of $\pi_\text{ref}$ as a uniform distribution.}
  \label{tab:pi_ref}
\end{table}

\end{document}